\documentclass[11pt]{article}

\usepackage[letterpaper,margin=1in]{geometry}
\usepackage{times}
\usepackage[round,authoryear]{natbib}
\setcitestyle{citesep={;},aysep={,},yysep={;}}
\usepackage{microtype}
\usepackage[utf8]{inputenc}
\usepackage[T1]{fontenc}
\usepackage{amsmath,amssymb}
\usepackage{booktabs}
\usepackage{multirow}
\usepackage{graphicx}
\usepackage{xcolor}
\usepackage{float}
\usepackage{caption}
\usepackage{subcaption}
\usepackage{tikz}
\usetikzlibrary{arrows.meta,positioning,calc,fit,backgrounds}
\usepackage{pgfplots}
\pgfplotsset{compat=1.16}
\usepackage{hyperref}
\usepackage{url}

\definecolor{cA}{HTML}{2A78D6}
\definecolor{cB}{HTML}{EB6834}
\definecolor{cC}{HTML}{1BAF7A}
\definecolor{cD}{HTML}{EDA100}
\definecolor{inkmuted}{HTML}{6B6A63}
\definecolor{boxfill}{HTML}{F1F0EC}

\floatstyle{ruled}
\newfloat{algo}{t}{loa}
\floatname{algo}{Algorithm}

\newcommand{\dppl}{$\Delta$PPL}

\newcommand{\KL}{\operatorname{KL}}
\newcommand{\relu}{\operatorname{ReLU}}
\newcommand{\softmax}{\operatorname{softmax}}

\newcommand{\Dense}{\textsc{Dense}}
\newcommand{\Oracle}{\textsc{Oracle}}
\newcommand{\ShortI}{\textsc{Short-8K}}
\newcommand{\LongI}{\textsc{Long-32K}}
\newcommand{\LongLL}{\textsc{Long-64K}}
\newcommand{\Skip}[1]{\textsc{Skip-#1}}
\newcommand{\compress}{\textsc{+compress}}
\newcommand{\RopeEight}{\textsc{RoPE-8}}
\newcommand{\NoPE}{\textsc{NoPE}}
\newcommand{\DistBias}{\textsc{DistBias}}
\newcommand{\Codec}[1]{\textsc{Codec-#1}}

\newcommand{\rot}[1]{\Phi_{#1}}
\newcommand{\Src}{\mathcal{S}}
\newcommand{\Uniform}[1]{\textsc{Uniform-#1}}
\newcommand{\Gapfill}[1]{\textsc{GapFill-#1}}

\hypersetup{colorlinks=true,linkcolor=blue!60!black,citecolor=blue!60!black,urlcolor=blue!60!black}

\title{Position-Skipped Distillation of Sparse-Attention Indexers:\\ Train Short, Serve Long}

\author{Fu Dai \and Haodi Quan \and Kamal Taha}

\date{}

\begin{document}
\maketitle

\begin{abstract}
Language models that employ top-$k$ sparse attention rely on an \emph{indexer}: a lightweight scoring module in each layer that ranks cached tokens so attention is computed only over the $k$ highest-scoring candidates.
Indexers are typically distilled from the model's dense attention using sequences as long as the target serving context, which makes each training step incur the cost of dense attention at serving length.
We show that when an indexer is distilled only on short sequences, its failure to generalise to longer contexts is driven primarily by rotary positional encoding (RoPE), rather than by degraded content matching.

We make two contributions.
First, we propose \emph{position-skipped distillation}. We train the indexer on short sequences whose position IDs include a randomly placed gap; the frozen teacher model that supplies distillation targets receives the same position IDs, ensuring that the teacher never processes sequences longer than the short training length.
On DeepSeek-V2-Lite and Qwen3-30B-A3B, an indexer distilled on 8K-token sequences with position-skipped distillation matches an indexer distilled on 32K-token sequences to within about 0.5 perplexity up to 32K. On DeepSeek-V2-Lite, it similarly matches a 64K-distilled indexer at 64K while requiring roughly $6\times$ less training compute per token.
On Llama-3.1-8B, it is within this margin of an initial 32K run, but is 0.6 perplexity worse than a second 32K run trained with a larger batch; at a budget of 2{,}048 keys, it performs on par with simply compressing the positions of an indexer distilled on 8K sequences.
On retrieval tasks, performance is comparable on average, although one 8K-distilled indexer underperforms when multiple values must be retrieved for a single key.
At 64K, position compression of the 32K-distilled indexer matches or outperforms the position-skipped variant on two of three models (by up to about one perplexity point), indicating that the primary benefit of position-skipped distillation is reduced training cost rather than improved quality.
Beyond the training distances, the outcome becomes model-dependent: on Llama-3.1 at 128K, the first two layers must retain dense attention, after which the indexer increases perplexity by 3\%.

Second, we introduce a compact index representation.
Keys are quantised to three or four bits per coordinate without retraining and scored using a fused integer kernel; with four-bit keys, perplexity changes by at most 0.05 at a budget of 2{,}048 keys, and the kernel scans 1.6--2.2$\times$ faster than a half-precision scan.
In a current serving engine, the indexer accelerates decoding only for long contexts with multiple concurrent streams, and slows down single-stream decoding.
\end{abstract}

\section{Introduction}
\label{sec:intro}

Modern language models are increasingly deployed with contexts spanning hundreds of thousands of tokens, making dense attention over the full context the dominant cost of token generation.
Top-$k$ sparse attention largely eliminates this cost by restricting attention to a small subset of cached tokens.
In the design introduced by \citet{deepseekv32}, each layer includes a lightweight learned scoring module (the \emph{indexer}) that assigns a score to every cached token; the layer then attends only to the top $k$ tokens by score.
Several production models adopt this approach or closely related variants~\citep{minimax,deepseekv4,longcat,qwen38next}.

Indexers introduce two key costs.
First, training is typically performed by distillation from the model's dense attention using sequences as long as the intended serving context, so each training step pays for dense attention at serving length; \citet{deepseekv32} report 1{,}000 steps with 16 sequences of 128K tokens each.
Second, inference requires maintaining an index (one key vector per token per layer) and scoring it at every decoding step---an operation we refer to as a \emph{scan}.

Accordingly, we ask two questions: (i) can an indexer intended for contexts of length $L$ be distilled using sequences that are much shorter than $L$; and (ii) can the resulting index be made smaller and faster to scan without retraining?

\begin{figure}[t]
\centering
\includegraphics[width=0.93\textwidth]{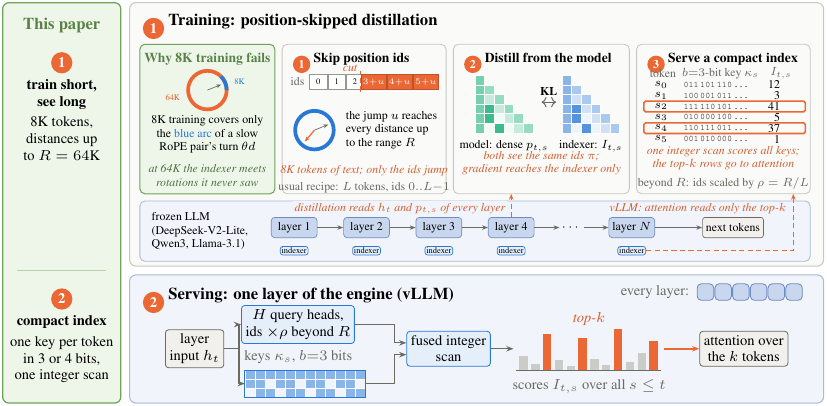}
\caption{Overview. Top, training. Left: an indexer distilled on 8K tokens sees only a small arc (blue) of a slow rotary pair, so at 64K it meets unseen rotations. (1)~Position-skipped distillation lets the ids of $T$ contiguous tokens jump by a random offset $u$ after a cut point, so every distance up to the range $R$ occurs. (2)~The frozen model computes dense attention under the same ids, and each layer's indexer matches its head-summed attention $p_{t,s}$ with a KL loss; only the indexer receives gradients. (3)~Each token keeps one $b$-bit key, an integer scan scores all keys, and attention reads the $k$ best (orange). Bottom, serving: in every layer $H$ query heads are scored against the $b$-bit keys by a fused integer scan; beyond $R$ the ids are scaled by $\rho=R/L$. Blue: our components.}
\label{fig:overview}
\end{figure}

A natural approach to reducing distillation cost is to train the indexer on shorter sequences; in practice, this naive strategy breaks down.
We evaluate an indexer using \dppl{} (the relative increase in perplexity compared with dense attention) and \emph{recall} (the fraction of the highest-attention tokens under dense attention that are also selected by the indexer).
On DeepSeek-V2-Lite, an indexer distilled on 8K-token sequences performs well at 8K; however, at 16K (with $k=512$), recall drops to 0.13 and \dppl{} exceeds 1000\%.
This failure is primarily positional: the indexer represents distance via rotary rotations of queries and keys, and beyond its training length it encounters rotations it has not observed. Rescaling only the position IDs at inference largely restores token selection.

To expose the indexer to long-range positional differences without running the full model on long sequences, we introduce \emph{position-skipped distillation} (Figure~\ref{fig:overview}, top), which applies the position-ID manipulation of \citet{pose} to the indexer only:
\begin{enumerate}
\item For $T$ contiguous tokens, the position IDs jump by a random offset after a random cut point. The text remains unchanged; only the coordinate system is stretched, so training covers all distances up to a range $R$.
\item The frozen teacher computes dense attention under the same position IDs and provides the distillation targets. Because both teacher and indexer use positions only through ID differences, the targets reflect long-range distances that are absent from the short sequence itself.
\item At inference time, distances beyond $R$ are mapped back into the trained range by scaling the indexer's position IDs, trading fine-grained resolution at long distances for coverage.
\end{enumerate}

To reduce the inference cost of the index, we quantise the stored keys of a distilled indexer to three or four bits per coordinate (without retraining) and score packed integers using a fused kernel.
We additionally investigate whether all layers require distinct indexers by allowing later layers to reuse the token selection produced by an earlier layer.

On DeepSeek-V2-Lite and Qwen3-30B-A3B, position-skipped distillation on 8K sequences achieves \dppl{} within about 0.5 points of an indexer distilled on 32K sequences up to 32K; on DeepSeek-V2-Lite it similarly matches a 64K-distilled indexer at 64K. In most comparisons, it is slightly worse, but by less than this margin.
On Llama-3.1-8B, it is 0.6 points worse than a second 32K run trained with a larger batch.
On a single H200 GPU, the cost per training token on DeepSeek-V2-Lite is 66\%, 83\%, and 91\% lower when using skipped IDs at 8K than when distilling at 32K, 64K, and 128K, respectively, and it does not require long-sequence memory or context parallelism.
Distilling at 32K and compressing position IDs performs better at 64K on two models (by up to about one \dppl{} point), reinforcing that skipping primarily improves training efficiency rather than peak quality.
We evaluate generalisation up to 160K and find that, beyond the trained range, performance degrades unless position IDs are compressed.
On Llama-3.1 at 128K, the first two layers must remain dense for both the indexer and \Oracle{}.
Finally, 4-bit keys incur no measurable perplexity increase at $k=2048$; when using indexers in only 8 of 27 layers, either placement reduces retrieval accuracy.

Our contributions are as follows.
\begin{itemize}
\item We show that an indexer distilled with standard position IDs fails to generalise beyond its training length, and that the dominant failure mode is positional (Section~\ref{sec:q2}).
\item We introduce position-skipped distillation, which matches distillation at 32K to within 0.5 perplexity up to 32K on two models and to within 1.0 perplexity on Llama-3.1, and matches distillation at 64K on DeepSeek-V2-Lite, while using 34\% and 17\% of the corresponding training cost per token (Sections~\ref{sec:psd} and~\ref{sec:q3}).
\item We present a compact index based on 3--4-bit keys and a fused integer scan that requires no retraining, and evaluate it in a serving engine (Sections~\ref{sec:index} and~\ref{sec:q5}).
\item We ablate the positional range on all three models; on DeepSeek-V2-Lite, we further study the effects of training tokens, positional mechanism, and sharing the indexer across fewer layers. On Llama-3.1, we trace the 128K failure mode to the first two layers (Sections~\ref{sec:placement}, \ref{sec:q4}, and~\ref{sec:q6}).
\end{itemize}

\section{Related work}
\label{sec:related}

\paragraph{Learned top-$k$ sparse attention and training the indexer.}
\citet{deepseekv32} introduced a sparse-attention design in which a learned \emph{indexer} scores all cached tokens and each layer attends only to the top $k$. In their recipe, the indexer is trained first via distillation on 128K-token sequences, with the base model frozen and dense attention providing the targets.
Related approaches learn selection at token or block granularity, either jointly with the language model~\citep{nsa,moba,minicpm4} or as part of later production systems~\citep{minimax,deepseekv4,longcat,qwen38next}.
\citet{seerattention,seerattentionr} distil a block-level gate from a frozen model's attention; our focus on distillation length is orthogonal to the choice of selection granularity.
In contrast, training-free selectors rely on heuristics such as page boundaries~\citep{quest}, vector search~\citep{retrievalattention}, or layerwise sharing of selections~\citep{tidaldecode}.
\citet{rtpurbo} and \citet{flashmemory} attach indexers to existing models and train them on contexts from 32K to 512K tokens, while \citet{adindex} distil a group-level indexer.
Beyond a brief remark in \citet{flashmemory} about using an indexer at twice its training length, we are not aware of systematic measurements of how far an indexer can be reliably extrapolated.

\paragraph{Position encodings and length extrapolation.}
The failure mode we analyse reflects a broader issue: positional encodings---including rotary embeddings (RoPE)~\citep{rope}, which the indexer uses---often generalise poorly to lengths not seen during training~\citep{nope,randpe}.
Prior work improves extrapolation by scaling position IDs into the trained range~\citep{pi}, refining scaling on a per-frequency basis~\citep{yarn}, or fine-tuning on short windows in which chunks receive random skips in their position IDs~\citep{pose}. \citet{shufflecontext} perturb rotary positions within a self-distillation objective to adapt models to longer contexts.
These methods extend the context of the language model itself, which can learn from next-token loss at any distance.
For indexers, however, the supervision signal---the model's dense attention---is only available at the distances present within the training sequence.
Our method combines the skipped IDs of \citet{pose} with the scaling of \citet{pi}, and additionally runs the frozen teacher under the same skipped IDs so that attention targets exist up to the desired range. We apply these ideas to a distilled ranking module rather than to full-model fine-tuning.
In a targeted (though not exhaustive) search, we did not find prior work that trains indexers using skipped position IDs.

\paragraph{Reducing index cost at inference.}
One line of work reduces the number of scans by reusing selections from nearby layers~\citep{indexcache}, filtering blocks before scoring tokens~\citep{hisa}, or sharing a scan across neighbouring queries~\citep{pivot,adindex}.
A second line reduces the cost of each scan: \citet{deepseekv32} execute the indexer in 8-bit floating point; \citet{deepseekv4} train the query--key path with 4-bit quantisation; and \citet{streamindex}, \citet{litetopk}, and \citet{gvr} avoid materialising the full score matrix or accelerate the top-$k$ step. Learned binary codes~\citep{hashattention,hata,spotlight} replace dot products in dense-attention models.
Our goal is to reduce both memory and decoding time for indexers whose keys must be read at every decoding step. We quantise keys after distillation, without retraining, to roughly one quarter of their original storage, while retaining the original indexer scoring function (Appendix~\ref{app:related}).

\section{Preliminaries and notation}
\label{sec:background}

This section summarises top-$k$ sparse attention with a learned indexer, together with the distillation objective used to train it. Table~\ref{tab:notation} lists the main symbols; we use them consistently throughout the paper.

\begin{table}[t]
\caption{Notation used throughout the paper and its appendix.}
\label{tab:notation}
\centering
\footnotesize
\setlength{\tabcolsep}{2.5pt}
\begin{tabular}{@{}ll@{\hspace{10pt}}ll@{}}
\toprule
Symbol & Meaning & Symbol & Meaning \\
\midrule
$L$ & context length at evaluation or serving & $h_t \in \mathbb{R}^{D}$ & input of a layer at token $t$ \\
$T$ & length of a training sequence & $a^{(g)}_{t,s}$ & dense attention of model head $g$ \\
$R$ & position range of skipped ids & $p_{t,s}$ & distillation target (head-summed) \\
$t,\ s$ & query token, key token ($s \le t$) & $q_{t,j} \in \mathbb{R}^{d}$ & query vector of indexer head $j$ \\
$\pi_t$ & position id of token $t$ (usually $t$) & $\kappa_s \in \mathbb{R}^{d}$ & indexer key of token $s$ (shared) \\
$c,\ u$ & cut point and offset of the skipped ids & $w_{t,j}$ & weight of indexer head $j$ for query $t$ \\
$\rho$ & scale of the indexer's ids, $\min(1, R/L)$ & $I_{t,s}$ & indexer score of key $s$ for query $t$ \\
$\rot{p}$ & rotary embedding at position $p$ & $S_t$ & the $k$ selected keys of query $t$ \\
$\theta,\ d_r$ & rotary base, number of rotated coordinates & $k$ & selection budget per query and layer \\
$H,\ d$ & indexer query heads, indexer head dimension & $b$ & bits per coordinate of a stored key \\
$N,\ \ell$ & number of layers, layer index & $\tilde\kappa_s,\ \alpha_s$ & integer key codes and their scale \\
\dppl{} & relative perplexity increase over \Dense{} & $\tilde q_{t,j},\ \beta_{t,j}$ & 8-bit query codes and their scale \\
\bottomrule
\end{tabular}
\end{table}

\paragraph{Top-$k$ sparse attention.}
Consider an attention layer with input $h_t \in \mathbb{R}^{D}$ at token $t$ and dense attention weights $a_{t,s}^{(g)}$ from model head $g$ over keys $s \le t$. Top-$k$ sparse attention restricts all heads to a shared set $S_t$ of $k$ keys, and renormalises the softmax over this set.
We form $S_t$ by selecting the $k$ keys with the largest values of a cheap score $I_{t,s}$ computed by an indexer.
The ideal score would be the layer's head-summed dense attention $\sum_g a_{t,s}^{(g)}$ (\Oracle{} selection), but this is unavailable at inference because it requires computing the very dense attention that sparsification seeks to avoid.

\paragraph{The indexer.}
We use the indexer of DeepSeek Sparse Attention~\citep{deepseekv32}, which has $H$ query heads and one key per token shared by all heads, as in multi-query attention:
\begin{equation}
\kappa_s = \rot{\pi_s}\,\mathrm{LN}(W_K h_s), \qquad
q_{t,j} = \rot{\pi_t}\,(W_Q h_t)_j, \qquad
w_{t,j} = \tfrac{1}{\sqrt{H}}\, (W_w h_t)_j ,
\label{eq:keys}
\end{equation}
where $W_K \in \mathbb{R}^{d \times D}$, $W_Q \in \mathbb{R}^{Hd \times D}$, and $W_w \in \mathbb{R}^{H \times D}$ are trainable projections, $(\cdot)_j$ extracts the slice corresponding to head $j$, and $\mathrm{LN}$ denotes layer normalisation.
Here $\pi_t$ is the position id of token $t$ (typically $\pi_t=t$), and $\rot{p}$ is the rotary embedding~\citep{rope} at position $p$, which rotates the first $d_r \le d$ coordinates in $d_r/2$ planes; plane $i$ is rotated by angle $p\,\theta^{-2i/d_r}$.
The indexer score is a weighted sum of rectified per-head similarities between queries and the shared key:
\begin{equation}
I_{t,s} = \sum_{j=1}^{H} w_{t,j}\, \relu\!\Big(\tfrac{1}{\sqrt{d}}\; q_{t,j}^{\top} \kappa_s\Big).
\label{eq:score}
\end{equation}
The indexer stores only $\kappa_s$ per token; scoring a single query against $L$ keys therefore costs $LHd$ multiply--adds (a scan).
Finally, since $\rot{\pi_t}^{\top}\rot{\pi_s}=\rot{\pi_s-\pi_t}$, the score depends on positions only through id differences; setting $d_r=0$ yields a position-agnostic indexer.

\paragraph{Distillation.}
For each query $t$, the distillation target is the normalised head-summed dense attention
$p_{t,s} = \sum_g a^{(g)}_{t,s} \big/ \sum_{s'}\sum_g a^{(g)}_{t,s'}$.
Over a training sequence of $T$ tokens, we minimise
\begin{equation}
\mathcal{L} = \frac{1}{T}\sum_{t=1}^{T} \KL\!\Big(p_{t,\cdot}\,\Big\|\,\softmax_{s \le t}\, I_{t,s}\Big).
\label{eq:kl}
\end{equation}
This matches the warm-up objective of \citet{deepseekv32}: the frozen language model acts as the teacher, its layer inputs are detached, and each layer's indexer is trained with its own loss. Consequently, each optimisation step requires one dense forward pass over the sequence and one scan per layer.

\section{Method}
\label{sec:method}
Our approach comprises three components: (i) position-skipped distillation, which trains an indexer on short sequences while preserving validity at the serving length (Section~\ref{sec:psd}); (ii) a compact index representation (Section~\ref{sec:index}); and (iii) a layer-wise placement strategy that applies the indexer only in selected layers (Section~\ref{sec:placement}).

\subsection{Position-skipped distillation}
\label{sec:psd}
\paragraph{The length problem.}
When an indexer is distilled using $T$-token sequences, it only observes rotary-relative positions $\rot{\pi_s-\pi_t}$ for distances up to roughly $T$. For larger distances, the long-wavelength rotary components take unseen angles, yielding novel combinations and degraded selection (Section~\ref{sec:q2}).
Training directly at the serving length $L \gg T$ resolves this issue, but requires a dense forward pass over $L$ tokens per optimisation step.
Instead, we keep sequences short and perturb their position ids. This follows positional skip-wise training (PoSE)~\citep{pose}, which extends a model's effective context by fine-tuning on short segments with shifted ids.

\paragraph{Position ids.}
Let $R \ge T$ denote the \emph{position range}, i.e., the maximum context length that the indexer is expected to support.
For each training sequence, we sample a cut point $c \sim \mathrm{Unif}([T/4,\,3T/4])$ and an offset $u \sim \mathrm{Unif}(\{0,\dots,R-T\})$, and define
\begin{equation}
\pi_i = \begin{cases} i, & i < c,\\ i + u, & i \ge c. \end{cases}
\label{eq:skip}
\end{equation}
The underlying text remains contiguous; only the position ids contain a gap. Distances within each chunk are unchanged, while cross-chunk distances increase by $u$. Across training, this exposes the indexer to all relative distances up to $R$.
\paragraph{Teacher and student share position ids.}
We use the frozen language model as the teacher, evaluated on the same short sequence but with the modified ids from Eq.~\ref{eq:skip} applied to its own rotary embedding. The distillation target in Eq.~\ref{eq:kl} is computed from this teacher forward pass, and the indexer is trained using the identical ids.
This shift is well defined because both dense attention and the indexer score depend on positions only through id differences. In particular, for a query in the second chunk, the teacher interprets the first chunk as if it were $u$ tokens farther in the past, and its attention therefore indicates how selection should vary with that distance.
No text is inserted into the gap. We assume the teacher model behaves sensibly for ids up to $R$, so $R$ must not exceed the model's maximum context length.

\paragraph{Cost.}
The procedure introduces no additional parameters or loss terms. The training length $T$ determines the per-step compute, while $R$ determines the range over which the distilled indexer is expected to generalise. Each step has the same cost as standard distillation at length $T$: attention scales as $O(T)$ per training token rather than $O(L)$, whereas the remaining components of the forward pass scale as usual, so the overall speedup is typically smaller than $L/T$ (Section~\ref{sec:q3}).
Algorithm~\ref{alg:psd} in Appendix~\ref{app:details} provides the full procedure.

\paragraph{Position compression beyond the range.}
When serving with context length $L > R$, we compress the indexer's position ids by $\rho = R/L$ at inference time so that the maximum relative distance is mapped back to $R$:
\begin{equation}
\kappa_s = \rot{\rho\pi_s}\,\mathrm{LN}(W_K h_s), \qquad q_{t,j} = \rot{\rho\pi_t}\,(W_Q h_t)_j, \qquad \rho = \min(1,\,R/L).
\label{eq:compress}
\end{equation}
This is position interpolation~\citep{pi} applied only to the indexer (without fine-tuning). In practice, it can be effective because the compression preserves the ordering of keys by distance. Setting $R=T$ yields the baseline \compress{}.

\subsection{Compact index}
\label{sec:index}
In 16-bit floating point, each key requires $2d$ bytes, and each decoding step must scan all keys. We reduce both memory footprint and scan cost without retraining.
Because decoding performs relatively few multiply--adds per byte read from memory, the scan is often bandwidth-bound; reducing the bytes per key therefore improves both storage and throughput.

\paragraph{Low-bit keys and 8-bit queries.}
We quantise each key to $b$ bits per coordinate using one per-token scale.
Let $\gamma_b = 2^{b-1}-1$ denote the maximum code magnitude. We store an integer code vector $\tilde\kappa_s$ together with a scale $\alpha_s$:
\begin{equation}
\alpha_s = \frac{\max_i |\kappa_{s,i}|}{\gamma_b}, \qquad
\tilde\kappa_{s,i} = \operatorname{round}\!\big(\kappa_{s,i}/\alpha_s\big) \in \{-\gamma_b,\dots,\gamma_b\},
\label{eq:quant}
\end{equation}
This representation uses $bd/8$ bytes for the codes plus two bytes for a 16-bit scale.
We quantise queries analogously to 8 bits (codes $\tilde q_{t,j}$ with scales $\beta_{t,j}$). We also quantise keys after the rotary transform so that stored keys can be scanned directly.

\paragraph{Fused integer scan.}
Because the rectifier is positively homogeneous, the quantisation scales can be factored out of Eq.~\ref{eq:score}:
\begin{equation}
I_{t,s} \;\approx\; \alpha_s \sum_{j=1}^{H} \big(w_{t,j}\,\beta_{t,j}/\sqrt{d}\big)\; \relu\!\big(\tilde q_{t,j}^{\top} \tilde\kappa_s\big).
\label{eq:intscore}
\end{equation}
The inner product is an exact integer dot product between an 8-bit vector and a $b$-bit vector.
We implement Eq.~\ref{eq:intscore} in a single GPU kernel with one thread per key: each thread unpacks the key codes once and accumulates the per-head dot products using four-way 8-bit integer instructions.
Compared with a matrix-multiplication-based implementation, this fused approach avoids materialising the $L \times H$ matrix of per-head products prior to the rectifier. The trade-off is reduced use of tensor cores, so the net benefit is empirical.

\subsection{Indexer placement across layers}
\label{sec:placement}
We can further reduce scanning and storage by executing the indexer only in a subset of layers $\Src$ that includes layer 0, following index reuse~\citep{indexcache}. Each remaining layer reuses the selection set $S_t$ from the closest preceding layer in $\Src$.
Index reuse is beneficial only when nearby layers select similar keys. We quantify this similarity using the Jaccard overlap $J_{\ell\ell'} = |S^{(\ell)}_t \cap S^{(\ell')}_t| / |S^{(\ell)}_t \cup S^{(\ell')}_t|$ between the selected sets of layers $\ell$ and $\ell'$ for the same query, averaged across queries.
We evaluate two strategies for choosing $n$ indexed layers: \Uniform{$n$} spaces them evenly throughout the network depth, whereas the non-uniform \Gapfill{$n$} includes layer 0 and layers identified as retrieval-critical by a causal analysis (Appendix~\ref{app:reuse}), and then iteratively adds the midpoint of the largest remaining gap.

\section{Experiments}
\label{sec:experiments}

Our experiments address six questions, one per subsection: (Q1) the intrinsic cost of top-$k$ selection; (Q2) when and why an indexer distilled on short sequences fails; (Q3) how closely position-skipped distillation matches distillation at the serving length; (Q4) which components of the training recipe are essential; (Q5) how far the index can be compressed and the resulting impact at serving time; and (Q6) whether running the indexer in fewer layers is sufficient.

\subsection{Setup}
\label{sec:setup}

\paragraph{Models.}
We evaluate three pretrained models. Our primary model is the public DeepSeek-V2-Lite checkpoint~\citep{deepseekv2} (27 layers; 16 attention heads). We also consider Qwen3-30B-A3B-Instruct-2507~\citep{qwen3}, a mixture-of-experts model with 48 layers (32 query heads; 4 key--value heads), and Llama-3.1-8B-Instruct~\citep{llama3}, a dense model with 32 layers (32 query heads; 8 key--value heads) and a native context length of 128K.
All models were originally trained with dense attention and rotary embeddings. We keep their weights frozen in 16-bit precision so that changes in perplexity reflect the selection mechanism rather than model adaptation.
Unless stated otherwise, we attach one indexer to each layer, with $H=16$, $d=64$, $d_r=32$, and $\theta = 10{,}000$, and restrict attention to the $k$ keys selected by the indexer.
Indexers are distilled using Eq.~\ref{eq:kl}. For DeepSeek-V2-Lite, the main run uses 400M training tokens; ablations, the matched 64K comparison, and all runs on Qwen3 and Llama use 100M tokens (Appendix~\ref{app:details}).

\paragraph{Data.}
For training, we use FineWeb-Edu, a subset of FineWeb~\citep{fineweb} (a 10B-token sample). Documents are segmented into contiguous blocks of the training length.
For evaluation, we use held-out sources: PG-19~\citep{pg19} for perplexity and our re-implementation of RULER~\citep{ruler} for retrieval-style tasks. No evaluation blocks overlap with the training data.

\paragraph{Compared indexers.}
The compared indexers differ only in the sequences used for distillation. \ShortI{} is the baseline trained on 8,192-token sequences with standard position ids $0,\dots,8191$. \LongI{} and \LongLL{} are trained on 32,768- and 65,536-token sequences, respectively, also with standard ids, and serve as distillation-at-serving-length baselines. \Skip{$R$} (ours) uses 8,192-token sequences with the skipped ids from Eq.~\ref{eq:skip}, exposing the indexer to relative distances up to a chosen range $R \in \{32\text{K}, 64\text{K}, 128\text{K}\}$ (default: \Skip{64K}). Note that the number in \Skip{$R$} denotes the range, not the sequence length.
The suffix \compress{} indicates inference-time id compression into the trained range (Eq.~\ref{eq:compress}).
We report two reference points: \Dense{} denotes the original dense-attention model, and \Oracle{} performs per-layer top-$k$ selection using the true head-summed dense attention.

\paragraph{Evaluation.}
\dppl{} denotes the relative perplexity increase over \Dense{} (in percent), measured on the final $L/16$ tokens of PG-19 blocks of length $L$ (where 8K denotes 8,192). Unless stated otherwise, we evaluate four blocks up to 32K and two blocks beyond; key comparisons are repeated on 16 blocks up to 64K and on eight blocks at 128K and 160K. We report bootstrap confidence intervals over blocks.
The PG-19 blocks are consecutive windows drawn from a small number of books (joined in file order). As a result, intervals that treat blocks as independent are optimistic (Appendix~\ref{app:details}).
Recall is defined as the fraction of \Oracle{}-selected keys that are also selected by the indexer.
The \emph{passcode probe} inserts the string ``The passcode is'' followed by a random six-character code at one of eight depths within a 32K block, and queries the model at the end of the block using the same prefix. A trial is counted as successful if every code token is the model's top prediction; we report the number of successes out of eight.
We additionally evaluate seven RULER ``needle'' tasks using teacher forcing, with 200 samples at 32K and (for four tasks) 50 samples at 64K.
Unless stated otherwise, we report results for $k\in\{512, 2048\}$, matching the selection budgets used in DeepSeek-V3.2.

\subsection{Q1, Q2: Cost of selection and failure of short training}
\label{sec:q1}
\label{sec:q2}

\paragraph{Q1.} On DeepSeek-V2-Lite with $k=2048$, \Oracle{} increases perplexity by 0.18\% at 8K and 1.03\% at 32K while retaining only 6\% of the context (Table~\ref{tab:oracle}; eight documents scored on their final 2,048 tokens). Thus, even perfect top-2048 selection incurs an inherent cost of roughly 1\%, which serves as a target for learned indexers.
\begin{table}[t]
\caption{Indexers distilled with 400M tokens on DeepSeek-V2-Lite: \dppl{} (\%) on PG-19 with recall in parentheses. The lower block compresses the indexer's ids into its trained range; ``in range'' means compression changes nothing ($\rho=1$). Four blocks up to 32K, two at 64K; bold is the best per row. The runs are matched in tokens, not steps: \ShortI{} takes 24,414 steps, \Skip{64K} 8,138 and \LongI{} 2,035 (Appendix~\ref{app:details}).}
\label{tab:main}
\centering
\small
\setlength{\tabcolsep}{4pt}
\begin{tabular}{llccc}
\toprule
 & & \multicolumn{3}{c}{\dppl{} (\%) (recall)} \\
\cmidrule(lr){3-5}
Length & $k$ & \ShortI{} & \LongI{} & \Skip{64K} (ours) \\
\midrule
8K  & 512  & $+1.76$ (.79) & $+1.88$ (.77) & $\mathbf{+1.51}$ (.78) \\
8K  & 2048 & $+0.35$ (.85) & $\mathbf{+0.30}$ (.83) & $+0.32$ (.83) \\
16K & 512  & $+1182$ (.13) & $\mathbf{+2.35}$ (.74) & $\mathbf{+2.35}$ (.73) \\
16K & 2048 & $+257$ (.17) & $+0.35$ (.78) & $\mathbf{+0.34}$ (.76) \\
32K & 512  & $+2106$ (.05) & $+5.46$ (.70) & $\mathbf{+5.17}$ (.70) \\
32K & 2048 & $+1098$ (.08) & $\mathbf{+1.19}$ (.73) & $+1.23$ (.73) \\
64K & 512  & $+2891$ (.01) & $+15.2$ (.34) & $\mathbf{+5.92}$ (.68) \\
64K & 2048 & $+2248$ (.02) & $+4.75$ (.34) & $\mathbf{+0.88}$ (.71) \\
\midrule
 & & \ShortI\compress{} & \LongI\compress{} & \Skip{64K}\compress{} \\
\midrule
16K & 512 / 2048 & $+2.39$ (.64) / $+0.59$ (.65) & in range & in range \\
32K & 512 / 2048 & $+6.08$ (.59) / $+1.71$ (.60) & in range & in range \\
64K & 512 / 2048 & $+8.57$ (.55) / $+1.83$ (.59) & $+6.95$ (.59) / $+1.69$ (.60) & in range \\
\bottomrule
\end{tabular}
\end{table}

\paragraph{Q2.} Table~\ref{tab:main} compares \ShortI{} (distilled on 8K sequences with standard ids) and \LongI{} (distilled on 32K sequences with standard ids) with \Skip{64K} (distilled on 8K sequences whose ids are skipped up to 64K).
\ShortI{} is close to \Oracle{} at 8K (0.35\% for $k=2048$), but fails after a single length doubling: at 16K and $k=512$, it reaches 1182\% \dppl{} with recall 0.13. Similarly, \LongI{} degrades beyond 32K, with recall dropping from 0.73 to 0.34 and \dppl{} increasing from 1.19\% to 4.75\%.
These results indicate that indexers trained with standard position ids typically generalise only up to their training length.

Two controls suggest that this failure is primarily positional. First, \compress{} modifies only the indexer's position ids at inference and largely restores performance: for example, at 32K and $k=2048$, \ShortI\compress{} achieves 1.71\% \dppl{} versus 1098\% for \ShortI{}, and recovers all eight passcodes in the probe where \ShortI{} recovers none (Table~\ref{tab:needle}).
Second, for an early \ShortI{} checkpoint, compressing ids by a factor of two (still permitting distances up to 16K) reduces \dppl{} at 32K only from 805\% to 565\%, whereas a factor of four reduces it to 0.70\%.
Overall, unseen rotary rotations at large distances are sufficient to disrupt the ranking. However, \ShortI\compress{} still trails \LongI{} at 32K by roughly 0.5 points (recall 0.60 vs. 0.73), motivating explicit exposure to long distances during distillation.

\subsection{Q3: How close do skipped ids come to training at the serving length?}
\label{sec:q3}

\paragraph{Perplexity up to 32K.}
Although \Skip{64K} is trained only on 8K sequences, it matches \LongI{} closely up to 32K: across 8K--32K, the \dppl{} differs by at most 0.37 points, and at 32K with $k=2048$ we observe 1.23\% for \Skip{64K} versus 1.19\% for \LongI{} with identical recall (Table~\ref{tab:main}).
On 16 evaluation blocks, the paired \dppl{} difference for $k=2048$ is at most 0.04 points, and all 95\% intervals lie between $-0.15$ and $+0.21$ points (Table~\ref{tab:pg19ci}), within our practical-equivalence margin of $\pm 0.5$ points.

At 64K, \Skip{64K} attains recall 0.71 and \dppl{} 0.88\% for $k=2048$, outperforming all compressed baselines, including \LongI\compress{} at 1.69\% (Figure~\ref{fig:length}). On 16 blocks, the gap to \LongI\compress{} is 0.37 points (95\% interval: 0.19 to 0.57; Table~\ref{tab:pg19ci}).

\begin{table}[t]
\caption{Matched runs at 100M training tokens on three models, $k=2048$: \dppl{} (\%) with recall in parentheses, four PG-19 blocks up to 32K and eight at 64K, with intervals in Table~\ref{tab:pairci}. The \compress{} row compresses the ids of \LongI{} into 32K.}
\label{tab:pair-main}
\centering
\small
\setlength{\tabcolsep}{2.2pt}
\resizebox{\linewidth}{!}{%
\begin{tabular}{lccccccc}
\toprule
 & \multicolumn{3}{c}{DeepSeek-V2-Lite} & \multicolumn{2}{c}{Qwen3-30B-A3B} & \multicolumn{2}{c}{Llama-3.1-8B} \\
\cmidrule(lr){2-4}\cmidrule(lr){5-6}\cmidrule(lr){7-8}
Length & \Skip{64K} & \LongI{} & \LongLL{} & \Skip{64K} & \LongI{} & \Skip{64K} & \LongI{} \\
\midrule
8K  & $+0.45$ (.81) & $+0.26$ (.82) & $+0.23$ (.82) & $-0.49$ (.78) & $-0.41$ (.78) & $-1.06$ (.80) & $-1.05$ (.81) \\
16K & $+0.36$ (.73) & $+0.36$ (.76) & $+0.42$ (.76) & $+0.33$ (.74) & $+0.02$ (.75) & $+0.47$ (.77) & $+0.33$ (.78) \\
32K & $+1.51$ (.68) & $+1.28$ (.71) & $+1.43$ (.72) & $+0.73$ (.66) & $+0.71$ (.67) & $+2.25$ (.71) & $+1.72$ (.72) \\
64K & $+1.11$ (.64) & $+5.80$ (.26) & $+1.10$ (.68) & $+1.73$ (.54) & $+1.76$ (.42) & $+2.21$ (.65) & $+4.26$ (.32) \\
64K, \compress{} & in range & $+1.38$ (.57) & in range & in range & $+0.79$ (.59) & in range & $+1.74$ (.64) \\
\bottomrule
\end{tabular}}
\end{table}

\paragraph{A matched 64K baseline and additional models.}
Table~\ref{tab:pair-main} repeats the comparison using full 100M-token runs and includes \LongLL{}, a matched baseline distilled directly at 64K.
On DeepSeek-V2-Lite, \Skip{64K} and \LongLL{} are essentially indistinguishable at 64K (paired difference $-0.01$ points; 95\% interval $-0.12$ to $0.09$).
On Qwen3 and Llama-3.1, the two methods differ by at most 0.53 points up to 32K (four evaluation blocks).
Across the in-range settings in Table~\ref{tab:pair-main}, \Skip{64K} never exceeds the long-context baselines in recall: it approaches distillation at the serving length, but remains slightly behind.
At 64K, \LongI\compress{} outperforms \Skip{64K} by 0.94 points on Qwen3 (interval 0.59 to 1.32) and by 0.47 points on Llama-3.1 (0.30 to 0.65), while underperforming by 0.26 points on DeepSeek-V2-Lite. Overall, position skipping reduces training cost, but does not uniformly match the quality of direct long-context distillation.

\paragraph{Repeat runs on Llama-3.1.}
On Llama-3.1, a second \Skip{64K} run changes \dppl{} by at most 0.07 points at 32K. In contrast, a second \LongI{} run with twice the batch size improves by 0.59 points at 32K for $k=2048$ (95\% interval 0.28 to 0.89), indicating that the observed gap depends on the baseline's batch size. At matched cost, \ShortI\compress{} is comparable for $k=2048$ but is approximately two points worse for $k=512$ (Appendix~\ref{app:seeds}).

\paragraph{Retrieval tasks.}
On seven RULER tasks at 32K with $k=2048$ (Table~\ref{tab:ruler}), the mean exact match is 0.599 for \Dense{}, 0.516 for \Skip{64K}, and 0.527 for \LongI{}; results are within two percentage points on five tasks.
\Skip{64K} performs better on the ``single needle in filler text'' task (0.995 vs.
0.940), but worse on the task containing four values for a single key (0.230 vs.
0.360). Both differences are confirmed by a paired sign test in a rerun ($p=0.02$ and $p=0.0015$, respectively). At 64K, all indexers fail on these tasks (Table~\ref{tab:ruler64}).
This deficit does not persist with 100M training tokens (Table~\ref{tab:ruler100}) or on Llama-3.1 (Appendix~\ref{app:third}); on Qwen3, the tasks are saturated.
On LongBench~v2~\citep{longbenchv2} at 32K, both indexers are within 0.014 accuracy of \Dense{} on all three models; however, DeepSeek-V2-Lite performs at chance level (Table~\ref{tab:longbench}).
\paragraph{A block-level baseline and training cost.}
Our re-implementation of the block-level indexer from \citet{flashmemory} (requiring twelve inferred implementation details) underperforms \Skip{64K} by 3.2 \dppl{} points on PG-19 at 32K (Appendix~\ref{app:flashmemory}), and fails on all four evaluated RULER tasks.
Measured per training token on a single H200, distillation at 32K, 64K, and 128K is 2.94$\times$, 5.92$\times$, and 11.69$\times$ more expensive than skipped-id training at 8K (Table~\ref{tab:cost}). End-to-end run times are reported in Table~\ref{tab:pair}.

\subsection{Q4: Which parts of the recipe matter?}
\label{sec:q4}
\paragraph{Position range and training tokens.}
On DeepSeek-V2-Lite, \Skip{32K}, \Skip{64K}, and \Skip{128K} agree within 0.44 points when evaluated within their respective ranges, indicating that training for a larger range does not harm short-context performance (Table~\ref{tab:range}).
Beyond the trained range, each method degrades as the evaluation distance increases; applying \compress{} largely restores performance (Tables~\ref{tab:range} and~\ref{tab:pg19ci128}). At 128K and 160K, the compressed indexers outperform the degraded \Dense{} baseline, so we do not rank them against \Dense{} in those settings (Table~\ref{tab:pg19ci160}).
At four times the trained range, \Skip{32K} incurs a 16.75\% loss on DeepSeek-V2-Lite and a 3.39\% loss on Qwen3, and fails on Llama-3.1; compression improves results for all three models (Appendices~\ref{app:second} and~\ref{app:llama128}).
On Llama-3.1 at 128K, all indexers (and \Oracle{}) fail unless layers 0 and 1 remain dense, consistent with DeepSeek-V4~\citep{deepseekv4}. With layers 0--1 dense, \Skip{64K} achieves 2.95\%, \Oracle{} achieves 5.37\%, and \Skip{32K}\compress{} achieves $-0.97\%$ (interval $-2.21$ to $+0.29$), suggesting that the failure originates in early layers (Appendix~\ref{app:llama128}). Reducing training from 400M to 100M tokens leaves \Skip{64K} within 0.4 points (Table~\ref{tab:pos}).

\paragraph{Position mechanism.}
We compare against three simpler alternatives distilled on 8K sequences with standard ids for 100M tokens. All underperform \Skip{64K} at 64K (Table~\ref{tab:pos}): \RopeEight{} rotates only $d_r = 8$ coordinates (10.5\% \dppl{} at $k=2048$), \NoPE{} sets $d_r = 0$ (with an enforced local window; Appendix~\ref{app:window}), and \DistBias{} augments \NoPE{} with a learned function of $t-s$.
Overall, skipped ids combined with full rotary rotation yields the strongest performance.

\subsection{Q5: A compact index and its effect in a serving engine}
\label{sec:q5}
\paragraph{Key formats.}
Relative to full-precision keys, top-2048 selection with \Skip{64K} retains 0.995 (32K) and 0.994 (64K) of the dense-attention mass when using four-bit keys (34 bytes per token vs.
128), and 0.977 (32K) and 0.972 (64K) when using three-bit keys (26 bytes) (Table~\ref{tab:formats}).
End-to-end on 16 blocks with $k=2048$, four-bit keys change \dppl{} by at most 0.05 points (intervals include zero), whereas three-bit keys increase \dppl{} by up to 0.15 points up to 32K and by 0.46 points at 64K (Table~\ref{tab:quante2e}). Thus, for $k=2048$, four-bit keys replace 16-bit keys without a measurable loss; for $k=512$, four-bit keys incur up to a 0.22-point penalty.
The fused kernel scans three-bit keys 1.6$\times$ to 2.2$\times$ faster than a 16-bit cuBLAS-based scan (Table~\ref{tab:kernel}).

\paragraph{Decoding in a serving engine.}
We integrate the indexer as an attention plugin in vLLM~0.32.0 and benchmark on an H200 (Appendix~\ref{app:serving}).
On DeepSeek-V2-Lite, the key--value cache is 1.15\,KB per token per layer. The strongest dense baseline is FlashMLA with an 8-bit cache. Using three-bit keys together with FlashMLA's sparse kernel, the indexer outperforms this baseline only at 64K with 16 concurrent streams (by 4.8\%), and is slower in the single-stream setting (Table~\ref{tab:servingfp8}).
On Qwen3 (2\,KB per token per layer), the indexer outperforms FlashAttention decoding by 19\% at 32K with 16 streams and by 13\% at 64K with 4 streams, but is 37\% slower at 8K with one stream (Table~\ref{tab:servingq3}). On Llama-3.1, under the configuration that preserves perplexity at 128K, it is 1.41$\times$ and 1.18$\times$ faster than BF16 and FP8 dense decoding, respectively, at 128K with 4 streams, but remains slower up to 64K (Table~\ref{tab:servingl8dep}). Overall, the approach pays off primarily for multiple concurrent long-context streams.

\subsection{Q6: Does the indexer need to run in every layer?}
\label{sec:q6}
Neighbouring layers only partially agree on which keys to select: at 32K, the Jaccard overlap ranges from 0.22 to 0.46 (Appendix~\ref{app:fewer}).
When we run the indexer in only 8 of 27 layers and reuse the nearest preceding selection in the remaining layers, a placement tuned on perplexity and the passcode probe outperforms a uniform placement on those tuning signals (2.86\% vs.
6.46\% \dppl{} for $k=2048$). However, neither placement generalises well to held-out RULER tasks, where both lose 7 to 29 points on three tasks.

\section{Conclusion}
\label{sec:conclusion}

Across the evaluated models, position-skipped distillation trained on 8K-token sequences achieves perplexity within 0.5 points of an indexer distilled at serving length on two models and within 1.0 point on Llama-3.1, while reducing training cost per token by at least $3\times$.
At 64K, simply compressing the positional representation of a 32K-distilled indexer matches or outperforms the position-skipped variant on two of three models, suggesting that the main advantage of position-skipped distillation is lower training cost rather than improved long-context quality.
We also find that 4-bit key quantisation preserves perplexity at $k=2048$, and that decoding speedups in a production serving engine primarily arise for long contexts under multi-stream concurrency, with limited or negative benefit for single-stream decoding.

\paragraph{Limitations.}
\label{sec:limits}
Our evaluation is limited to models originally trained with dense attention, and most experimental configurations were run only once.
At 128K context on Llama-3.1, the first two layers still require dense attention.
On DeepSeek-V2-Lite, LongBench~v2 performance is near chance (Appendix~\ref{app:limits}).

\bibliographystyle{plainnat}
\bibliography{refs}

\appendix
\section{Training and evaluation details}
\label{app:details}

Table~\ref{tab:hparams} summarizes the hyperparameters shared across distillation runs and highlights the settings that vary by configuration.
Runs differ in the number of optimizer steps because we hold the total number of training tokens fixed while changing the number of GPUs.
\ShortI{} uses 24{,}414 steps with 16K tokens per step; the 400M-token \Skip{$R$} runs use 8{,}138 steps with 49K tokens per step; and \LongI{} uses 2{,}035 steps with 197K tokens per step.
All 100M-token runs use 2{,}035 steps with 49K tokens per step.
Accordingly, the comparison between \Skip{64K} and \LongI{} in Section~\ref{sec:q3} is matched by training tokens rather than optimizer steps; we did not run a step-matched variant.

\begin{table}[h]
\caption{Distillation settings. ``Tokens per step'' is the number of training tokens that contribute to one optimiser step, over all GPUs.}
\label{tab:hparams}
\centering
\small
\begin{tabular}{lp{0.62\linewidth}}
\toprule
Setting & Value \\
\midrule
Frozen model & DeepSeek-V2-Lite (27 layers), Qwen3-30B-A3B (48 layers) or Llama-3.1-8B (32 layers), 16-bit floating point \\
Indexer per layer & $H=16$, $d=64$, $d_r=32$, $\theta=10{,}000$ \\
Objective & Eq.~\ref{eq:kl}, all query positions, each layer separately \\
Optimiser & AdamW, $(\beta_1,\beta_2)=(0.9, 0.95)$, no weight decay \\
Learning rate & $10^{-3}$, 100 warm-up steps, cosine decay \\
Gradient clipping & norm 1.0 \\
Training text & FineWeb-Edu (sample-10BT), contiguous blocks of the training length \\
Seeds & one run per configuration \\
\midrule
\ShortI{} & 8K sequences, 24,414 steps, 16K tokens per step, two A100 \\
\LongI{} & 32K sequences, 2,035 steps, 197K tokens per step, six H200 \\
\Skip{$R$}, 400M tokens & 8K sequences, 8,138 steps, 49K tokens per step, six H200 \\
100M-token ablations & 8K sequences, 2,035 steps, 49K tokens per step, one A100 \\
100M-token matched runs & 8K, 32K or 64K sequences, one H200 each (Qwen3 \LongI{}: one, then two; Llama: data-parallel, three H200 at 8K and two at 32K, with the same tokens per step) \\
\bottomrule
\end{tabular}
\end{table}

\paragraph{Position skipping in the frozen model.}
The frozen model uses rotary position embeddings in attention. During distillation, we replace the model's default position IDs with the skipped IDs from Eq.~\ref{eq:skip} when computing target distributions.
In the released configurations, DeepSeek-V2-Lite supports position IDs up to 163{,}840, while Qwen3-30B-A3B and Llama-3.1-8B support at least 131{,}072; these limits bound the usable range $R$.

\paragraph{Evaluation blocks.}
We tokenize the PG-19 test books and concatenate them in file order without separators. For each length $L$, we then take the first consecutive windows of $L$ tokens from this stream.
As a result, a block may cross a book boundary, and blocks at a given length are drawn primarily from the first portion of the test set.
Using the tokenizers of all three models, the 16 blocks at 8K come from 2 books; at 16K from 4 books; and at 32K from 9 books, with 1, 3, and 7--8 blocks spanning a boundary, respectively. The 8 blocks at 64K come from 9 books (6 spanning a boundary), and the 8 blocks at 128K come from 15 books (all spanning a boundary).
Because blocks from the same book are not independent, bootstrap intervals computed over blocks (treating them as independent samples) are narrower than intervals computed over books would be, especially up to 32K.
We evaluate the sparse model and \Dense{} on the same blocks, and compute \dppl{} from the mean token loss over scored positions across all blocks.
When a table uses a different number of blocks, we note this in the caption; unless blocks are identical, comparisons should be made only within a table.

\paragraph{Distillation step.}
Algorithm~\ref{alg:psd} gives one step of position-skipped distillation.

\begin{algo}[h]
\caption{One step of position-skipped distillation for a frozen model with $N$ attention layers.}
\label{alg:psd}
\small
\begin{tabular}{@{}rl@{}}
\multicolumn{2}{@{}l}{\textbf{Input:} tokens $x_{1:T}$; position range $R \ge T$; frozen model $f$; indexers $\{\phi_\ell\}_{\ell=1}^{N}$; query chunk size $B$.}\\
\multicolumn{2}{@{}l}{\textbf{State:} indexer parameters and optimiser state. \textbf{Output:} updated indexer parameters.}\\[2pt]
1: & Draw $c \sim \mathcal{U}[T/4,\,3T/4]$ and $u \sim \mathcal{U}\{0,\dots,R-T\}$; set $\pi_i = i$ for $i<c$ and $\pi_i = i+u$ for $i \ge c$.\\
2: & Run $f$ on $x_{1:T}$ with position ids $\pi$, without gradients. For every layer $\ell$, record the layer\\
   & input $h^{(\ell)}_{1:T}$ and, per chunk of $B$ queries, the dense attention weights $a^{(\ell,g)}_{t,s}$.\\
3: & \textbf{for} each layer $\ell$ and each query chunk $\mathcal{C}$ \textbf{do}\\
4: & \quad $p_{t,s} \leftarrow \sum_g a^{(\ell,g)}_{t,s} \big/ \sum_{s'}\sum_g a^{(\ell,g)}_{t,s'}$ for $t \in \mathcal{C}$, $s \le t$. \hfill (target)\\
5: & \quad $\kappa_s \leftarrow \rot{\pi_s}\mathrm{LN}(W_K h^{(\ell)}_s)$ for $s \le \max \mathcal{C}$; \;$q_{t,j}, w_{t,j}$ from Eq.~\ref{eq:keys} for $t \in \mathcal{C}$.\\
6: & \quad $I_{t,s} \leftarrow \sum_j w_{t,j}\,\relu(q_{t,j}^{\top}\kappa_s/\sqrt{d})$; mask $s > t$.\\
7: & \quad Accumulate the gradient of $\frac{1}{T}\sum_{t \in \mathcal{C}} \KL\big(p_{t,\cdot}\,\|\,\softmax_s I_{t,s}\big)$ with respect to $\phi_\ell$.\\
8: & \textbf{end for}\\
9: & Clip the gradient norm and take one optimiser step on all $\phi_\ell$.\\
\end{tabular}
\end{algo}

\paragraph{Distance bias.}
The \DistBias{} indexer in Section~\ref{sec:q4} sets $d_r=0$ and augments the score with the term $v(h_t)^{\top}\varphi(t-s)$. Here, $\varphi$ is a piecewise-linear basis of six functions of $\log_2(1+t-s)$, and $v(h_t)$ is an affine function of the query input.

\paragraph{Training cost.}
Table~\ref{tab:cost} reports the per-token training time quoted in Section~\ref{sec:q3}.
Computing teacher targets for only one quarter of the query rows reduces wall-clock time by just 13\%, because the forward pass of the frozen model dominates the step cost.
Unless stated otherwise, timings use a single A100 with identical chunking and data pipelines, and are measured over steps 10--30 of a short run.
The full 100M-token runs completed in approximately 13 hours each, implying a lower time per token than the microbenchmark. We did not determine the source of this discrepancy; therefore, Table~\ref{tab:cost} should be interpreted primarily in terms of ratios between rows rather than absolute throughput.
On Qwen3-30B-A3B, the measured ratios are 2.63 at 32K and 5.45 at 64K, slightly below those of DeepSeek-V2-Lite. This run coincided with other jobs on the same node, making the ratios more reliable than the absolute times. For Llama-3.1-8B, we did not perform controlled timing; because the two runs used different GPU counts, we do not compare their wall-clock times.

\paragraph{Cost at 128K.}
We did not distill any indexer to completion on 128K sequences. Instead, we time 20 steps of this recipe on DeepSeek-V2-Lite using a single H200, ordinary positions, one 131{,}072-token sequence per step, and query chunk size 1024.
The median step time over steps 6--20 is 343.0\,s, corresponding to 2{,}616.9\,\textmu s per token, i.e., 11.69$\times$ the cost of skipped IDs at 8K (Table~\ref{tab:cost}).
These measurements were taken on a shared node. Under the same conditions, a 64K timing run yields 1{,}327.5\,\textmu s per token, only 0.23\% above the idle-node value in Table~\ref{tab:cost}, suggesting that node load perturbs the ratio by less than 1\%.
Overall, time per token approximately doubles with each doubling of sequence length: 2.01$\times$ from 32K to 64K and 1.98$\times$ from 64K to 128K.
We do not time the 128K recipe on Qwen3. If its per-token cost follows the slope observed between 32K and 64K, it would reach roughly 6{,}150\,\textmu s per token (11.1$\times$ the cost at 8K); this is an extrapolation, not a measurement.
\Skip{128K} trains on 8K sequences and therefore has the same per-token cost as \Skip{64K}.
The warm-up phase of \citet{deepseekv32} uses about 2.1B tokens (1{,}000 steps of 16 sequences of 128K tokens), compared to 100M and 400M tokens in our runs. We retain their objective but train on 8K sequences, and we do not test whether larger models require additional training tokens.

\begin{table}[h]
\caption{Distillation cost per training token, measured in steady state. Upper block: one A100, identical settings. Middle block: DeepSeek-V2-Lite on one H200 while the rest of the node was idle, with the settings of the 100M-token runs (49,152 tokens per step at 8K, 65,536 at 32K and 64K; query chunk 2048, 1024 at 64K), median over steps 6 to 20 of 20; the 128K row, one sequence per step and query chunk 1024, is a 20-step timing run made later while other jobs used the node, under which a 64K run took 1,327.5\,\textmu s, and no 128K indexer was trained. Lower block: Qwen3-30B-A3B with the settings of its 100M-token runs (query chunk 1024 at every length; no 64K indexer was trained on this model), timed in the same way on one H200 while other jobs ran on the other cards of the node; step times are recorded in whole seconds.}
\label{tab:cost}
\centering
\small
\begin{tabular}{lcc}
\toprule
Distillation run & ms per token & Relative \\
\midrule
8K sequences, position-skipped (\Skip{64K}) & 0.797 & 1.00 \\
8K sequences, position-skipped, one quarter of the query rows & 0.696 & 0.87 \\
32K sequences, ordinary positions (\LongI{}) & 1.554 & 1.95 \\
\midrule
\multicolumn{3}{l}{DeepSeek-V2-Lite, one H200, idle node, settings of the 100M-token runs (\textmu s per token)} \\
8K sequences, position-skipped (\Skip{64K}) & 223.8 & 1.00 \\
32K sequences, ordinary positions (\LongI{}) & 658.2 & 2.94 \\
64K sequences, ordinary positions (\LongLL{}) & 1324.5 & 5.92 \\
128K sequences, ordinary positions, timing run only & 2616.9 & 11.69 \\
\midrule
\multicolumn{3}{l}{Qwen3-30B-A3B, one H200, other cards of the node busy (\textmu s per token)} \\
8K sequences, position-skipped (\Skip{64K}) & 554.7 & 1.00 \\
32K sequences, ordinary positions (\LongI{}) & 1456.7 & 2.63 \\
64K sequences, ordinary positions & 3022.3 & 5.45 \\
\bottomrule
\end{tabular}
\end{table}

\paragraph{Matched runs at 100M tokens.}
Table~\ref{tab:pair} extends the DeepSeek-V2-Lite part of Table~\ref{tab:pair-main} with $k=512$ and with 128K, which lies outside the range of all three indexers.
The runs are matched in training tokens and not in steps: \LongI{} and \LongLL{} take 1,526 steps of 65.5K tokens and \Skip{64K} 2,035 steps of 49.2K tokens.
The 64K column comes from the eight blocks of Table~\ref{tab:pairci}, which gives intervals and paired differences, and the other columns from four documents up to 32K and two at 128K.
For $k=512$ \Skip{64K} is 0.36 points better than \LongLL{} at 64K, with an interval from 0.00 to 0.76.
The \Skip{64K} run repeats the 100M-token run of Table~\ref{tab:pos} on another machine, and up to 32K its values differ from that run by at most 0.16 points.
The wall-clock times come from one run each while other jobs used the other cards of the node, so the text quotes the controlled measurement of Table~\ref{tab:cost} for the cost ratio.

\begin{table}[h]
\caption{Matched runs at 100M training tokens, one H200 per run: \dppl{} (\%) for $k=512$ / $k=2048$ and wall-clock time of the complete distillation run. \LongLL{} is distilled on 64K sequences with ordinary positions. Four PG-19 documents up to 32K, eight at 64K and two at 128K; 128K is outside the range of all three indexers, and the values there are indicative only.}
\label{tab:pair}
\centering
\small
\setlength{\tabcolsep}{3.5pt}
\resizebox{\linewidth}{!}{%
\begin{tabular}{lcccccc}
\toprule
Indexer & Hours & 8K & 16K & 32K & 64K & 128K \\
\midrule
\Skip{64K} & 6.3 & $+1.74$ / $+0.45$ & $+2.35$ / $+0.36$ & $+5.37$ / $+1.51$ & $+5.66$ / $+1.11$ & $+11.59$ / $-0.43$ \\
\Skip{64K}\compress{} & & & & & & $+1.76$ / $-3.57$ \\
\LongI{} & 18.3 & $+1.95$ / $+0.26$ & $+2.37$ / $+0.36$ & $+5.73$ / $+1.28$ & $+17.54$ / $+5.80$ & $+35.61$ / $+12.66$ \\
\LongI\compress{} & & & & & $+6.12$ / $+1.38$ & $+4.82$ / $-2.79$ \\
\LongLL{} & 36.9 & $+1.81$ / $+0.23$ & $+2.51$ / $+0.42$ & $+5.80$ / $+1.43$ & $+6.02$ / $+1.10$ & $+13.25$ / $+3.56$ \\
\LongLL\compress{} & & & & & & $+4.18$ / $-3.40$ \\
\bottomrule
\end{tabular}}
\end{table}

\paragraph{Matched runs with intervals.}
Table~\ref{tab:pairci} repeats the matched comparison at 32K and 64K with intervals and paired differences over blocks.
On DeepSeek-V2-Lite at 64K, \Skip{64K} and \LongLL{} are level for $k=2048$, with a difference of $-0.01$ points and an interval from $-0.12$ to $0.09$.
\LongI\compress{} is 0.26 points behind \Skip{64K} there, with an interval from 0.00 to 0.52, in the same direction as the 400M-token runs of Table~\ref{tab:pg19ci}.
At 32K the two indexers distilled with ordinary ids are 0.07 and 0.11 points better than \Skip{64K} for $k=2048$, with intervals that reach zero, and 0.33 and 0.18 points worse for $k=512$; at this budget the 400M-token runs show no difference.
On Qwen3 at 64K the order of compression and skipped ids is reversed: \LongI\compress{} is 0.94 points better than \Skip{64K} for $k=2048$, with an interval from 0.59 to 1.32 (Appendix~\ref{app:second}), and on Llama-3.1 by 0.47 points, with an interval from 0.30 to 0.65 (Appendix~\ref{app:third}).
Each indexer was trained once, so the intervals cover the choice of blocks and not the variation between training runs.

\begin{table}[h]
\caption{Matched runs at 100M training tokens with 95\% bootstrap intervals over blocks: \dppl{} (\%) and the paired difference to \Skip{64K} of the same model on the same blocks, in points (positive: \Skip{64K} is better). DeepSeek-V2-Lite: the 16 blocks of Table~\ref{tab:pg19ci} at 32K and the first eight of them at 64K. Qwen3-30B-A3B and Llama-3.1-8B: eight blocks at 64K each. \compress{} compresses the ids of \LongI{} into 32K.}
\label{tab:pairci}
\centering
\small
\setlength{\tabcolsep}{3pt}
\resizebox{\linewidth}{!}{%
\begin{tabular}{llcccc}
\toprule
 & & \multicolumn{2}{c}{\dppl{} (\%)} & \multicolumn{2}{c}{Difference to \Skip{64K}} \\
\cmidrule(lr){3-4}\cmidrule(lr){5-6}
Length & Indexer & $k=512$ & $k=2048$ & $k=512$ & $k=2048$ \\
\midrule
\multicolumn{6}{@{}l}{\emph{DeepSeek-V2-Lite}} \\
32K & \Skip{64K} & $5.08$ [$4.19$, $6.00$] & $1.36$ [$1.00$, $1.74$] & & \\
 & \LongI{} & $5.26$ [$4.29$, $6.28$] & $1.25$ [$0.86$, $1.67$] & $+0.18$ [$-0.01$, $+0.39$] & $-0.11$ [$-0.23$, $+0.01$] \\
 & \LongLL{} & $5.41$ [$4.42$, $6.51$] & $1.29$ [$0.93$, $1.67$] & $+0.33$ [$+0.08$, $+0.60$] & $-0.07$ [$-0.16$, $-0.00$] \\
64K & \Skip{64K} & $5.66$ [$4.03$, $7.43$] & $1.11$ [$0.56$, $1.68$] & & \\
 & \LongI{} & $17.54$ [$14.24$, $21.38$] & $5.80$ [$4.33$, $7.28$] & $+11.88$ [$+9.51$, $+14.72$] & $+4.69$ [$+3.62$, $+5.80$] \\
 & \LongI\compress{} & $6.12$ [$4.24$, $8.16$] & $1.38$ [$0.75$, $2.10$] & $+0.46$ [$-0.11$, $+1.12$] & $+0.26$ [$+0.00$, $+0.52$] \\
 & \LongLL{} & $6.02$ [$4.37$, $7.91$] & $1.10$ [$0.63$, $1.59$] & $+0.36$ [$-0.00$, $+0.76$] & $-0.01$ [$-0.12$, $+0.09$] \\
\midrule
\multicolumn{6}{@{}l}{\emph{Qwen3-30B-A3B}} \\
64K & \Skip{64K} & $6.52$ [$4.88$, $8.25$] & $1.73$ [$1.25$, $2.16$] & & \\
 & \LongI{} & $7.02$ [$5.16$, $9.10$] & $1.76$ [$1.12$, $2.53$] & $+0.50$ [$+0.11$, $+0.92$] & $+0.04$ [$-0.27$, $+0.43$] \\
 & \LongI\compress{} & $5.29$ [$3.64$, $7.05$] & $0.79$ [$0.30$, $1.36$] & $-1.23$ [$-1.59$, $-0.92$] & $-0.94$ [$-1.32$, $-0.59$] \\
\midrule
\multicolumn{6}{@{}l}{\emph{Llama-3.1-8B}} \\
64K & \Skip{64K} & $6.11$ [$4.54$, $7.88$] & $2.21$ [$1.16$, $3.49$] & & \\
 & \LongI{} & $11.77$ [$9.59$, $14.38$] & $4.26$ [$2.98$, $5.64$] & $+5.67$ [$+4.00$, $+7.58$] & $+2.04$ [$+1.22$, $+3.16$] \\
 & \LongI\compress{} & $6.40$ [$5.13$, $7.90$] & $1.74$ [$0.76$, $2.87$] & $+0.30$ [$-0.16$, $+0.76$] & $-0.47$ [$-0.65$, $-0.30$] \\
\bottomrule
\end{tabular}}
\end{table}

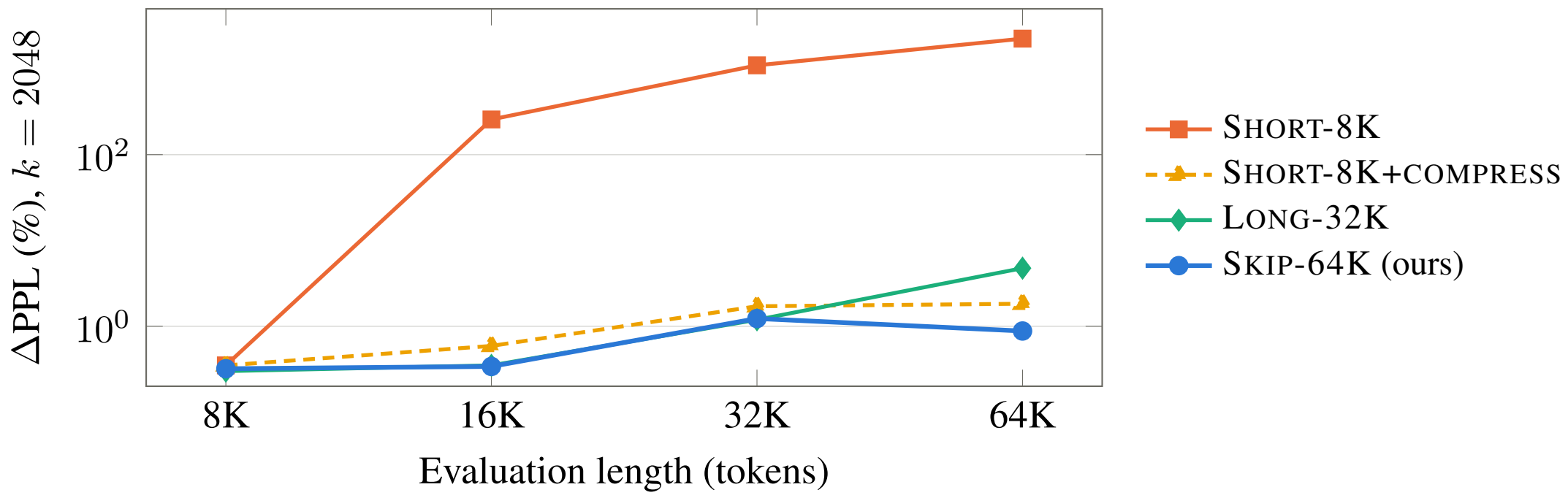
\begin{figure}[h]
\centering
\begin{tikzpicture}
\begin{axis}[
  width=0.62\linewidth, height=5.0cm,
  ymode=log, ymin=0.2, ymax=5000,
  symbolic x coords={8K,16K,32K,64K}, xtick=data,
  xlabel={Evaluation length (tokens)}, ylabel={\dppl{} (\%), $k=2048$},
  ymajorgrids, grid style={inkmuted!25, line width=0.3pt},
  axis line style={inkmuted}, tick style={inkmuted},
  tick label style={font=\small}, label style={font=\small},
  legend style={font=\footnotesize, draw=none, fill=none, at={(1.03,0.5)}, anchor=west, cells={anchor=west}},
  clip=false,
]
\addplot[cB, line width=1pt, mark=square*, mark size=1.8pt] coordinates {(8K,0.35) (16K,257) (32K,1098) (64K,2248)};
\addlegendentry{\ShortI{}}
\addplot[cD, line width=1pt, mark=triangle*, mark size=2.2pt, densely dashed] coordinates {(8K,0.35) (16K,0.59) (32K,1.71) (64K,1.83)};
\addlegendentry{\ShortI\compress{}}
\addplot[cC, line width=1pt, mark=diamond*, mark size=2.2pt] coordinates {(8K,0.30) (16K,0.35) (32K,1.19) (64K,4.75)};
\addlegendentry{\LongI{}}
\addplot[cA, line width=1.2pt, mark=*, mark size=1.9pt] coordinates {(8K,0.32) (16K,0.34) (32K,1.23) (64K,0.88)};
\addlegendentry{\Skip{64K} (ours)}
\end{axis}
\end{tikzpicture}
\caption{\dppl{} over \Dense{} at $k=2048$ as a function of the evaluation length, on a logarithmic axis; the values are those of Table~\ref{tab:main}. \ShortI{} fails beyond 8K and \LongI{} degrades beyond 32K. \Skip{64K}, distilled on 8K sequences, stays at or below 1.23\% up to 64K.}
\label{fig:length}
\end{figure}

\paragraph{Oracle selection and passcode probe.}
Table~\ref{tab:oracle} gives the \Oracle{} values with confidence intervals that Section~\ref{sec:q1} quotes.
Figure~\ref{fig:length} plots the $k=2048$ values of Table~\ref{tab:main} across lengths.
Table~\ref{tab:needle} gives the passcode retrieval at 32K for the indexers of Table~\ref{tab:main}.
In the probe, a sentence stating a random passcode of six letters and digits is inserted into PG-19 text at one of eight depths between 10\% and 80\% of a 32K context.
The context ends with the beginning of the same sentence, so that the model has to copy the passcode.

\begin{table}[h]
\caption{\dppl{} (\%) of \Oracle{} selection on PG-19 with a 95\% bootstrap confidence interval over eight documents. The last 2,048 tokens of each document are scored. The last column is the share of dense attention mass that falls on the selected positions at $k=2048$, averaged over layers.}
\label{tab:oracle}
\centering
\small
\setlength{\tabcolsep}{3.5pt}
\begin{tabular}{lcccc}
\toprule
Context & $k=512$ & $k=1024$ & $k=2048$ & Mass at $k=2048$ \\
\midrule
8K  & $+1.51$ [$+1.11$, $+1.88$] & $+0.61$ [$+0.31$, $+0.90$] & $+0.18$ [$-0.03$, $+0.37$] & 96.6\% \\
16K & $+4.41$ [$+3.27$, $+5.57$] & $+1.96$ [$+1.46$, $+2.42$] & $+0.70$ [$+0.43$, $+0.92$] & 91.4\% \\
32K & $+6.64$ [$+4.80$, $+8.38$] & $+2.86$ [$+1.86$, $+3.77$] & $+1.03$ [$+0.54$, $+1.52$] & 88.9\% \\
\bottomrule
\end{tabular}
\end{table}

\begin{table}[h]
\caption{Passcode retrieval at 32K: number of samples, out of eight, in which every passcode token is predicted correctly. \Dense{} retrieves 8/8.}
\label{tab:needle}
\centering
\small
\begin{tabular}{lcccc}
\toprule
$k$ & \ShortI{} & \ShortI\compress{} & \LongI{} & \Skip{64K} \\
\midrule
512  & 0/8 & 8/8 & 7/8 & 8/8 \\
2048 & 0/8 & 7/8 & 7/8 & 7/8 \\
\bottomrule
\end{tabular}
\end{table}

\section{Perplexity with confidence intervals and retrieval tasks}
\label{app:ruler}

This appendix gives the two evaluations with larger samples that Section~\ref{sec:q3} quotes.
Both use the 400M-token indexers of Table~\ref{tab:main}, except for the 100M-token \LongLL{} in Table~\ref{tab:ruler64} and the 100M-token indexers of Table~\ref{tab:ruler100}.

\paragraph{Perplexity on 16 blocks.}
Table~\ref{tab:pg19ci} repeats the comparison of Table~\ref{tab:main} on 16 blocks of PG-19 per length, with a 95\% bootstrap interval over the blocks.
The blocks differ from those of Table~\ref{tab:main}, so the absolute values differ, and differences of a few hundredths of a point change sign between the two tables.
The last column is the paired difference between \LongI{} and \Skip{64K} on the same blocks, where a positive value means that \Skip{64K} is better.
From 8K to 32K the difference for $k=2048$ is 0.02 to 0.04 points and its interval contains zero, and for $k=512$ \Skip{64K} is better by 0.16 to 0.28 points, with intervals that exclude zero at 8K and 32K.
At 64K \Skip{64K} is better than \LongI\compress{} by 0.37 points for $k=2048$, with an interval that excludes zero.

\begin{table}[h]
\caption{\dppl{} (\%) over \Dense{} on 16 blocks of PG-19 per length, with 95\% bootstrap intervals over blocks. At 64K the \LongI{} column is \LongI\compress{}; without compression \LongI{} has $+13.17$ and $+4.34$ there. \ShortI\compress{} is \ShortI{} without compression at 8K. The last column is the paired difference between the \LongI{} column and \Skip{64K}.}
\label{tab:pg19ci}
\centering
\small
\setlength{\tabcolsep}{3pt}
\begin{tabular}{llcccc}
\toprule
Length & $k$ & \Skip{64K} & \LongI{} & \ShortI\compress{} & \LongI{} $-$ \Skip{64K} \\
\midrule
8K  & 512  & $3.06$ [$2.27$, $3.88$] & $3.35$ [$2.54$, $4.15$] & $3.31$ [$2.55$, $4.07$] & $+0.28$ [$+0.06$, $+0.52$] \\
8K  & 2048 & $0.63$ [$0.41$, $0.86$] & $0.65$ [$0.41$, $0.87$] & $0.56$ [$0.36$, $0.74$] & $+0.02$ [$-0.15$, $+0.21$] \\
16K & 512  & $3.96$ [$3.00$, $4.92$] & $4.12$ [$3.04$, $5.27$] & $4.42$ [$3.35$, $5.52$] & $+0.16$ [$-0.11$, $+0.47$] \\
16K & 2048 & $0.86$ [$0.50$, $1.24$] & $0.89$ [$0.53$, $1.26$] & $1.09$ [$0.67$, $1.53$] & $+0.03$ [$-0.10$, $+0.16$] \\
32K & 512  & $4.96$ [$4.18$, $5.78$] & $5.19$ [$4.30$, $6.13$] & $6.12$ [$5.04$, $7.21$] & $+0.23$ [$+0.07$, $+0.39$] \\
32K & 2048 & $1.18$ [$0.89$, $1.48$] & $1.21$ [$0.88$, $1.56$] & $1.69$ [$1.30$, $2.06$] & $+0.04$ [$-0.09$, $+0.16$] \\
64K & 512  & $4.44$ [$3.45$, $5.47$] & $5.14$ [$3.91$, $6.46$] & $6.52$ [$4.93$, $8.28$] & $+0.70$ [$+0.39$, $+1.06$] \\
64K & 2048 & $0.73$ [$0.38$, $1.10$] & $1.10$ [$0.67$, $1.55$] & $1.49$ [$0.99$, $2.01$] & $+0.37$ [$+0.19$, $+0.57$] \\
\bottomrule
\end{tabular}
\end{table}

\paragraph{Perplexity at 128K.}
Table~\ref{tab:pg19ci128} gives the same evaluation at 128K on eight blocks for three 400M-token indexers: \Skip{64K}, \LongI{} and \Skip{128K}; \ShortI{} was not run at this length.
128K is outside the range of \Skip{64K} and \LongI{} and inside that of \Skip{128K}.
With their ordinary ids the first two fail: for $k=2048$ \Skip{64K} has 13.76\% with recall 0.29, and \LongI{} 4.08\% with recall 0.17.
Compressing the ids into the range brings \Skip{64K} to $-3.82$\% and \LongI{} to $-3.27$\%, with recall 0.57 and 0.56.
The two statements of Section~\ref{sec:q4}, failure outside the range and recovery by compression, therefore hold with intervals that exclude zero.

The three configurations whose ids lie inside a trained range, \Skip{64K}\compress{}, \LongI\compress{} and \Skip{128K}, all have a lower perplexity than \Dense{} for $k=2048$, on seven or eight of the eight blocks.
We read this as a weakness of the dense attention of this model at 128K and not as a merit of the indexers, and we do not count it as a gain.
For the same reason the differences among the three, at most 1.21 points, do not rank their selections.
\Skip{128K} agrees most with the dense top-$k$, with recall 0.63, while its \dppl{} is 0.76 points above that of \Skip{64K}\compress{} for $k=2048$ and level with it for $k=512$.
A range that covers 128K thus gives no lower perplexity at this length than compression into a range of 64K.

\begin{table}[h]
\caption{\dppl{} (\%) over \Dense{} at 128K on eight blocks of PG-19, 400M-token indexers, with 95\% bootstrap intervals over blocks. Recall is for $k=2048$. The last two columns are the paired difference to \Skip{64K}\compress{} on the same blocks, in points. \LongI\compress{} compresses into 32K, \Skip{64K}\compress{} into 64K; \Skip{128K} is inside its range.}
\label{tab:pg19ci128}
\centering
\small
\setlength{\tabcolsep}{3pt}
\resizebox{\linewidth}{!}{%
\begin{tabular}{lccccc}
\toprule
 & \multicolumn{2}{c}{\dppl{} (\%)} & & \multicolumn{2}{c}{Difference to \Skip{64K}\compress{}} \\
\cmidrule(lr){2-3}\cmidrule(lr){5-6}
Indexer & $k=512$ & $k=2048$ & Recall & $k=512$ & $k=2048$ \\
\midrule
\Skip{64K} & $27.07$ [$21.11$, $35.06$] & $13.76$ [$10.35$, $17.95$] & 0.29 & $+27.37$ [$+22.11$, $+34.31$] & $+17.58$ [$+14.52$, $+21.11$] \\
\LongI{} & $12.84$ [$8.95$, $17.47$] & $4.08$ [$1.29$, $7.55$] & 0.17 & $+13.15$ [$+10.00$, $+16.66$] & $+7.90$ [$+5.19$, $+11.07$] \\
\Skip{64K}\compress{} & $-0.30$ [$-2.10$, $+1.46$] & $-3.82$ [$-5.88$, $-1.99$] & 0.57 & & \\
\LongI\compress{} & $0.90$ [$-1.21$, $+3.09$] & $-3.27$ [$-5.31$, $-1.48$] & 0.56 & $+1.21$ [$+0.58$, $+1.97$] & $+0.55$ [$+0.19$, $+0.94$] \\
\Skip{128K} & $-0.14$ [$-1.66$, $+1.36$] & $-3.06$ [$-4.78$, $-1.50$] & 0.63 & $+0.17$ [$-0.20$, $+0.55$] & $+0.76$ [$+0.45$, $+1.07$] \\
\bottomrule
\end{tabular}}
\end{table}

\paragraph{Perplexity at 160K.}
DeepSeek-V2-Lite accepts position ids up to 163,840, so we repeat the evaluation at 160K, blocks of 163,840 tokens, on eight blocks with the last 10,240 tokens of each scored (Table~\ref{tab:pg19ci160}).
The compared configurations are \Skip{64K}\compress{} ($\rho=0.4$), \Skip{128K} with its own ids, 1.25 times beyond its range, \Skip{128K}\compress{} ($\rho=0.8$), \LongI\compress{} ($\rho=0.2$) and \Oracle{}.
As at 128K, every sparse configuration, \Oracle{} included, has a lower perplexity than \Dense{} on all eight blocks for $k=2048$.
The perplexity of \Dense{} also varies from 4.8 to 11.8 between blocks, so we read only the paired differences to \Skip{64K}\compress{} and do not count the negative \dppl{} as a gain.
\Oracle{}, which keeps the keys with the largest dense attention, is 2.90 points above \Skip{64K}\compress{} for $k=2048$; this suggests that the negative values reflect a weakness of dense attention at this length rather than a better selection.

For $k=2048$ the three compressed indexers lie within 0.39 points of each other.
\Skip{128K} with its own ids is 2.09 points behind \Skip{64K}\compress{} for $k=2048$ and 3.71 for $k=512$, with recall 0.457 against 0.555, and compressing its ids into 128K brings it to within 0.39 points.
An indexer beyond its range thus degrades at 160K as at 128K, and compression avoids this; compressing \LongI{} into 32K, five times shorter than the context, costs 0.24 points against compression into 64K for $k=2048$ and 0.96 for $k=512$.

\begin{table}[h]
\caption{\dppl{} (\%) over \Dense{} on DeepSeek-V2-Lite at 160K (163,840 tokens) on eight blocks of PG-19, 400M-token indexers, with 95\% bootstrap intervals over blocks. Recall is for $k=2048$. The last two columns are the paired difference to \Skip{64K}\compress{} on the same blocks, in points. $\rho$ is the compression factor of Eq.~\ref{eq:compress}; \Skip{128K} without compression uses its own ids, beyond its range. Every configuration is below \Dense{}, which is weak at this length, so only the paired differences are read.}
\label{tab:pg19ci160}
\centering
\small
\setlength{\tabcolsep}{3pt}
\resizebox{\linewidth}{!}{%
\begin{tabular}{lccccc}
\toprule
 & \multicolumn{2}{c}{\dppl{} (\%)} & & \multicolumn{2}{c}{Difference to \Skip{64K}\compress{}} \\
\cmidrule(lr){2-3}\cmidrule(lr){5-6}
Configuration & $k=512$ & $k=2048$ & Recall & $k=512$ & $k=2048$ \\
\midrule
\Skip{64K}\compress{}, $\rho=0.4$ & $-8.55$ [$-13.24$, $-4.91$] & $-11.73$ [$-16.69$, $-7.40$] & 0.555 & & \\
\Skip{128K} & $-4.84$ [$-9.02$, $-1.73$] & $-9.63$ [$-14.34$, $-5.68$] & 0.457 & $+3.71$ [$+2.76$, $+4.63$] & $+2.09$ [$+1.40$, $+2.86$] \\
\Skip{128K}\compress{}, $\rho=0.8$ & $-8.40$ [$-13.20$, $-4.63$] & $-11.34$ [$-16.27$, $-7.02$] & 0.544 & $+0.15$ [$-0.20$, $+0.49$] & $+0.39$ [$+0.18$, $+0.59$] \\
\LongI\compress{}, $\rho=0.2$ & $-7.59$ [$-11.93$, $-4.41$] & $-11.49$ [$-16.40$, $-7.23$] & 0.532 & $+0.96$ [$+0.42$, $+1.55$] & $+0.24$ [$+0.05$, $+0.45$] \\
\Oracle{} & $-5.95$ [$-10.03$, $-2.65$] & $-8.83$ [$-13.29$, $-5.08$] & --- & $+2.59$ [$+1.86$, $+3.38$] & $+2.90$ [$+2.20$, $+3.54$] \\
\bottomrule
\end{tabular}}
\end{table}

\paragraph{Retrieval tasks.}
We re-implemented seven needle-in-a-haystack tasks of RULER~\citep{ruler} from its published templates.
A needle is a sentence that assigns a value to a key, and the question asks for the values of a given key.
In the three single-needle tasks one needle is hidden in repeated filler sentences (single-1) or in book text, with a number (single-2) or a 32-character identifier (single-3) as the value.
Multi-key-1 adds three needles with other keys to book text, and in multi-key-2 the whole context consists of needles with other keys.
Multi-value hides four values of one key and asks for all of them, and multi-query hides four keys and asks for all four values.
We use text from PG-19 where RULER uses essays.
Our attention code does not support cached generation, so the scoring is teacher-forced: the gold answer follows the answer prefix of RULER, and a sample is an exact match when every token of the answer values is the most likely token.
The values are therefore not comparable with published RULER scores, which use free generation.
All rows of one task use the same 200 samples at 32K with $k=2048$.
The runs of Table~\ref{tab:ruler} hold one mean per task and no per-sample scores, so the standard errors there treat the rows as independent and are conservative for a paired comparison; the paired tests below come from a rerun with per-sample scores.

\begin{table}[h]
\caption{Exact-match rate on seven RULER tasks at 32K with $k=2048$, 200 samples per task. The indexers are those of Table~\ref{tab:main}. One standard error of a rate is at most 0.036. Bold marks the better of \Skip{64K} and \LongI{} where they differ by more than two unpaired standard errors.}
\label{tab:ruler}
\centering
\small
\setlength{\tabcolsep}{4.5pt}
\begin{tabular}{lccccc}
\toprule
Task & \Dense{} & \Skip{64K} & \LongI{} & \ShortI\compress{} & \ShortI{} \\
\midrule
single-1 (filler text)      & 0.995 & \textbf{0.995} & 0.940 & 0.935 & 0.000 \\
single-2 (book text)        & 0.990 & 0.955 & 0.975 & 0.985 & 0.000 \\
single-3 (identifier value) & 0.690 & 0.695 & 0.675 & 0.735 & 0.000 \\
multi-key-1 (four keys)     & 0.680 & 0.585 & 0.595 & 0.520 & 0.000 \\
multi-key-2 (all needles)   & 0.130 & 0.035 & 0.030 & 0.040 & 0.000 \\
multi-value (four values)   & 0.545 & 0.230 & \textbf{0.360} & 0.165 & 0.000 \\
multi-query (four queries)  & 0.160 & 0.120 & 0.115 & 0.095 & 0.000 \\
\midrule
Mean of seven tasks         & 0.599 & 0.516 & 0.527 & 0.496 & 0.000 \\
\bottomrule
\end{tabular}
\end{table}

Table~\ref{tab:ruler} reports exact-match rates.
\ShortI{} without compression fails to retrieve the needle on any of the seven tasks, consistent with its high perplexity in Table~\ref{tab:main}.
Averaged across tasks, \Skip{64K} and \LongI{} achieve mean exact-match rates of 0.516 and 0.527, respectively.
On single-1, \Skip{64K} outperforms \LongI{} by 5.5 points (3.1 standard errors), whereas on multi-value it underperforms by 13.0 points (2.9 standard errors). On the remaining five tasks, the difference is at most 2.0 points and below 1.2 standard errors.
Relative to \Dense{}, both indexers lose at most 5.5 points on the single-needle tasks and between 8.5 and 31.5 points on multi-key-1, multi-key-2, and multi-value.
The underlying model performs poorly on multi-key-2 and multi-query (\Dense{} achieves 0.130 and 0.160), limiting these tasks' ability to distinguish indexers.

\paragraph{Paired tests.}
We rerun single-1 and multi-value for \Dense{}, \Skip{64K}, and \LongI{} on an A100, logging per-sample scores on the same samples and using the same indexers.
On multi-value, the rerun yields exact-match rates of 0.540 (\Dense{}), 0.255 (\Skip{64K}), and 0.335 (\LongI{}). In 20 samples, only \LongI{} is correct, versus 4 samples where only \Skip{64K} is correct (exact two-sided sign test, $p=0.0015$).
On single-1, the rerun yields 0.995 (\Dense{}), 0.990 (\Skip{64K}), and 0.945 (\LongI{}), with 11 samples correct only under \Skip{64K} versus 2 only under \LongI{} ($p=0.02$).
Thus, both differences reported in Table~\ref{tab:ruler} persist under paired evaluation.
Because the rerun changes a small number of samples near the decision boundary, its rates differ from Table~\ref{tab:ruler} by up to 2.5 points; correspondingly, the multi-value gap is 8.0 points in the rerun versus 13.0 points in the table.

\paragraph{Indexers distilled with 100M tokens.}
Table~\ref{tab:ruler100} repeats four tasks for the three 100M-token indexers from Table~\ref{tab:pair-main}, using the same 200 samples and logging per-sample scores.
Performance is statistically indistinguishable: no paired sign test comparing \Skip{64K} to either alternative yields $p<0.14$.
On multi-value, \Skip{64K} reaches 0.335 versus 0.340 for \LongI{}, with 17 samples correct only under \Skip{64K} and 16 only under \LongI{}. Therefore, the multi-value deficit observed for the 400M-token \Skip{64K} does not appear at 100M tokens.
With only one run per configuration, we cannot determine whether the deficit reflects run-to-run variability or an effect of longer training.

\begin{table}[h]
\caption{Exact-match rate on four RULER tasks at 32K with $k=2048$, 200 samples per task, for the three 100M-token indexers of Table~\ref{tab:pair-main} on DeepSeek-V2-Lite. The last two columns count the samples right only with the named indexer / only with \Skip{64K}, with the $p$-value of an exact two-sided sign test.}
\label{tab:ruler100}
\centering
\small
\setlength{\tabcolsep}{4pt}
\resizebox{\linewidth}{!}{%
\begin{tabular}{lccccc}
\toprule
Task & \Skip{64K} & \LongLL{} & \LongI{} & \LongLL{} / \Skip{64K} & \LongI{} / \Skip{64K} \\
\midrule
single-1 (filler text)    & 0.940 & 0.970 & 0.920 & 12 / 6 ($p=0.24$) & 7 / 11 ($p=0.48$) \\
single-2 (book text)      & 0.955 & 0.975 & 0.975 & 5 / 1 ($p=0.22$) & 6 / 2 ($p=0.29$) \\
multi-key-1 (four keys)   & 0.570 & 0.600 & 0.580 & 11 / 5 ($p=0.21$) & 9 / 7 ($p=0.80$) \\
multi-value (four values) & 0.335 & 0.300 & 0.340 & 5 / 12 ($p=0.14$) & 17 / 16 ($p=1.00$) \\
\bottomrule
\end{tabular}}
\end{table}

\paragraph{Controls on the multi-needle loss.}
Table~\ref{tab:rulerctl} separates the loss of the selection budget from the loss of the indexer on multi-key-1.
\Oracle{} with $k=2048$ loses 6.5 points against \Dense{}, and \Skip{64K} loses a further 3.0 points.
With $k=4096$ \Skip{64K} is 3.5 points below \Dense{}, which is less than the standard error of this difference, 4.7 points.
On this task the budget of 2,048 positions is thus the larger part of the loss.
Multi-value, the task on which \Skip{64K} trails \LongI{}, behaves differently.
\Oracle{} with $k=2048$ reaches 0.430 there, which is 11.5 points below \Dense{}, and \Skip{64K} is a further 20.0 points below \Oracle{}, at 4.4 standard errors; \LongI{}, at 0.360, is 7.0 points below \Oracle{}.
With $k=4096$ \Skip{64K} reaches 0.445, which is the level of \Oracle{} with $k=2048$ and 10.0 points below \Dense{}.
On this task the budget of 2,048 thus explains about one third of the loss of \Skip{64K} and its selection the rest, and a doubled budget removes the part that is due to the selection.
\LongI{} and \Oracle{} were not run with $k=4096$, so the table does not show how much a better selection would gain at that budget.

\begin{table}[h]
\caption{Controls at 32K on 200 samples per task: \Oracle{} selection with $k=2048$ and the \Skip{64K} indexer with $k=4096$. Token accuracy is the share of answer-value tokens that are predicted correctly.}
\label{tab:rulerctl}
\centering
\small
\begin{tabular}{llccc}
\toprule
Task & Selection & $k$ & Exact match & Token accuracy \\
\midrule
multi-key-1 & \Dense{}     & ---  & 0.680 & 0.954 \\
multi-key-1 & \Oracle{}    & 2048 & 0.615 & 0.945 \\
multi-key-1 & \Skip{64K}   & 2048 & 0.585 & 0.939 \\
multi-key-1 & \Skip{64K}   & 4096 & 0.645 & 0.949 \\
\midrule
multi-value & \Dense{}     & ---  & 0.545 & 0.981 \\
multi-value & \Oracle{}    & 2048 & 0.430 & 0.972 \\
multi-value & \Skip{64K}   & 2048 & 0.230 & 0.959 \\
multi-value & \Skip{64K}   & 4096 & 0.445 & 0.972 \\
\bottomrule
\end{tabular}
\end{table}

\paragraph{Retrieval tasks at 64K.}
Table~\ref{tab:ruler64} repeats four of the tasks at 64K with 50 samples per task.
\LongI\compress{} is the 400M-token \LongI{} with its ids compressed into 32K, and \LongLL{} is the 100M-token indexer distilled at 64K, so the columns differ in training tokens.
On the two single-needle tasks every indexer is within two points of \Dense{}.
On multi-value all three indexers fall to between 0.06 and 0.08, against 0.26 for \Dense{}, so at this length the task shows a loss of every indexer rather than a deficit of \Skip{64K}.
On multi-key-1 \LongI\compress{} reaches only 0.18, against 0.38 for \Skip{64K} and \LongLL{}; against \LongLL{}, on the same samples, 11 samples are right only with \LongLL{} and one only with \LongI\compress{} (two-sided sign test, $p=0.006$). On single-2 the same holds, 0.74 against 0.92, with 11 samples right only with \LongLL{} and 2 only with \LongI\compress{} ($p=0.02$), so beyond its training length the compressed 32K indexer loses retrieval that the 64K-trained one keeps.
\Dense{} itself reaches only 0.34 and 0.26 on the two multi-needle tasks at this length, against 0.680 and 0.545 at 32K, so these rows say less about the indexers than those of Table~\ref{tab:ruler}.
The \Dense{} and \Skip{64K} rows have no per-sample scores, so comparisons with them are unpaired.

\begin{table}[h]
\caption{Exact-match rate on four RULER tasks at 64K with $k=2048$, 50 samples per task, the same samples in every column. \Skip{64K} and \LongI\compress{}: 400M training tokens; \LongLL{}: 100M. One standard error of a rate is at most 0.071.}
\label{tab:ruler64}
\centering
\small
\begin{tabular}{lcccc}
\toprule
Task & \Dense{} & \Skip{64K} & \LongI\compress{} & \LongLL{} \\
\midrule
single-1 (filler text)    & 0.98 & 0.96 & 0.98 & 1.00 \\
single-2 (book text)      & 0.92 & 0.90 & 0.74 & 0.92 \\
multi-key-1 (four keys)   & 0.34 & 0.38 & 0.18 & 0.38 \\
multi-value (four values) & 0.26 & 0.06 & 0.06 & 0.08 \\
\bottomrule
\end{tabular}
\end{table}

\section{Downstream questions: LongBench v2}
\label{app:longbench}

Because the RULER tasks and the passcode probe are synthetic, we additionally evaluate the indexers on LongBench~v2~\citep{longbenchv2}. LongBench~v2 contains 503 multiple-choice questions over long documents, with four answer options per question.
Each context is truncated in the middle to 32{,}768 tokens, following the benchmark's official truncation policy at this length. We score the model by selecting the option letter (A--D) with the highest next-token logit after the prompt, so no free-form text is generated.
For each model, we compare \Dense{} to \Skip{64K} and \LongI{} (400M-token indexers for DeepSeek-V2-Lite; 100M-token indexers for Qwen3 and Llama-3.1). Both indexers select $k=2048$ keys in every layer, and 32K falls within the supported range of both.
All configurations are evaluated on the same 503 questions, enabling per-question paired comparisons. We report two-sided paired sign tests (counting questions answered correctly by only one configuration) and a 95\% bootstrap interval for the accuracy difference.
Chance performance is 0.25.

Table~\ref{tab:longbench} reports accuracy.
On Llama-3.1 and Qwen3, \Dense{} achieves accuracies of 0.318 and 0.356, respectively; in both cases the confidence intervals exclude chance.
On these two models, both indexers remain within 0.014 of \Dense{}, and no paired difference is significant ($p\ge 0.35$). The largest observed gap is \LongI{} on Llama-3.1 ($-0.014$), corresponding to 24 questions lost and 17 gained.
On DeepSeek-V2-Lite, \Dense{} achieves 0.225, with a confidence interval of [0.189, 0.262] that includes chance. Consequently, LongBench~v2 provides little signal on this model, and the corresponding rows do not meaningfully evaluate selection quality.
With 503 questions, the confidence intervals for differences are approximately $\pm 0.025$, so the benchmark can rule out large degradations from indexing on models that perform above chance, but not differences of 1--2 points.

\begin{table}[h]
\caption{LongBench~v2 accuracy, 503 four-option questions truncated in the middle to 32K tokens, answer chosen by the highest next-token logit among A to D (chance 0.25), $k=2048$ in every layer. Columns by the benchmark's length category, with the number of questions. $\Delta$: paired difference to \Dense{} with a 95\% bootstrap interval over questions (10,000 resamples); $p$: two-sided sign test over the questions on which the two configurations disagree, with the counts answered correctly only by \Dense{} / only by the indexer. The interval of \Dense{} is a 95\% interval of its accuracy.}
\label{tab:longbench}
\centering
\small
\setlength{\tabcolsep}{3pt}
\resizebox{\linewidth}{!}{%
\begin{tabular}{llccccccc}
\toprule
Model & Configuration & All & Short (180) & Medium (215) & Long (108) & $\Delta$ [95\% CI] & Disagree & $p$ \\
\midrule
Llama-3.1-8B & \Dense{} & 0.318 [0.278, 0.360] & 0.378 & 0.293 & 0.269 & & & \\
 & \Skip{64K} & 0.316 & 0.406 & 0.279 & 0.241 & $-0.002$ [$-0.028$, $+0.024$] & 22 / 21 & 1.00 \\
 & \LongI{} & 0.304 & 0.400 & 0.256 & 0.241 & $-0.014$ [$-0.040$, $+0.010$] & 24 / 17 & 0.35 \\
\midrule
Qwen3-30B-A3B & \Dense{} & 0.356 [0.314, 0.400] & 0.411 & 0.326 & 0.324 & & & \\
 & \Skip{64K} & 0.360 & 0.422 & 0.326 & 0.324 & $+0.004$ [$-0.020$, $+0.028$] & 19 / 21 & 0.88 \\
 & \LongI{} & 0.360 & 0.411 & 0.326 & 0.343 & $+0.004$ [$-0.018$, $+0.026$] & 15 / 17 & 0.86 \\
\midrule
DeepSeek-V2-Lite & \Dense{} & 0.225 [0.189, 0.262] & 0.239 & 0.219 & 0.213 & & & \\
 & \Skip{64K} & 0.215 & 0.211 & 0.219 & 0.213 & $-0.010$ [$-0.038$, $+0.020$] & 29 / 24 & 0.58 \\
 & \LongI{} & 0.221 & 0.200 & 0.233 & 0.231 & $-0.004$ [$-0.034$, $+0.024$] & 28 / 26 & 0.89 \\
\bottomrule
\end{tabular}}
\end{table}

\section{Cross-layer overlap and index reuse}
\label{app:reuse}

This appendix gives the details of Section~\ref{sec:q6}.

\paragraph{Retrieval layers.}
The layers that \Gapfill{$n$} takes first were found by a causal analysis on the passcode probe at 32K with \Oracle{} selection and $k=2048$.
Removing the passcode span from the selected sets of layers 7, 10, 12, 19 and 21 lowers the token accuracy of the copy from 1.000 to 0.347 and retrieves none of the eight passcodes, whereas removing it from five other layers (1, 3, 5, 13 and 23) changes nothing.
Layer 12 alone accounts for a loss of 10 points of token accuracy, and layer 21 alone for none.
Because these layers were chosen with the passcode probe, the probe is in-sample for \Gapfill{$n$}, and only the RULER tasks test the placement on held-out data.
With eight layers, \Uniform{8} uses layers 0, 4, 7, 11, 15, 19, 22 and 26, and \Gapfill{8} uses layers 0, 3, 7, 10, 12, 15, 19 and 21.

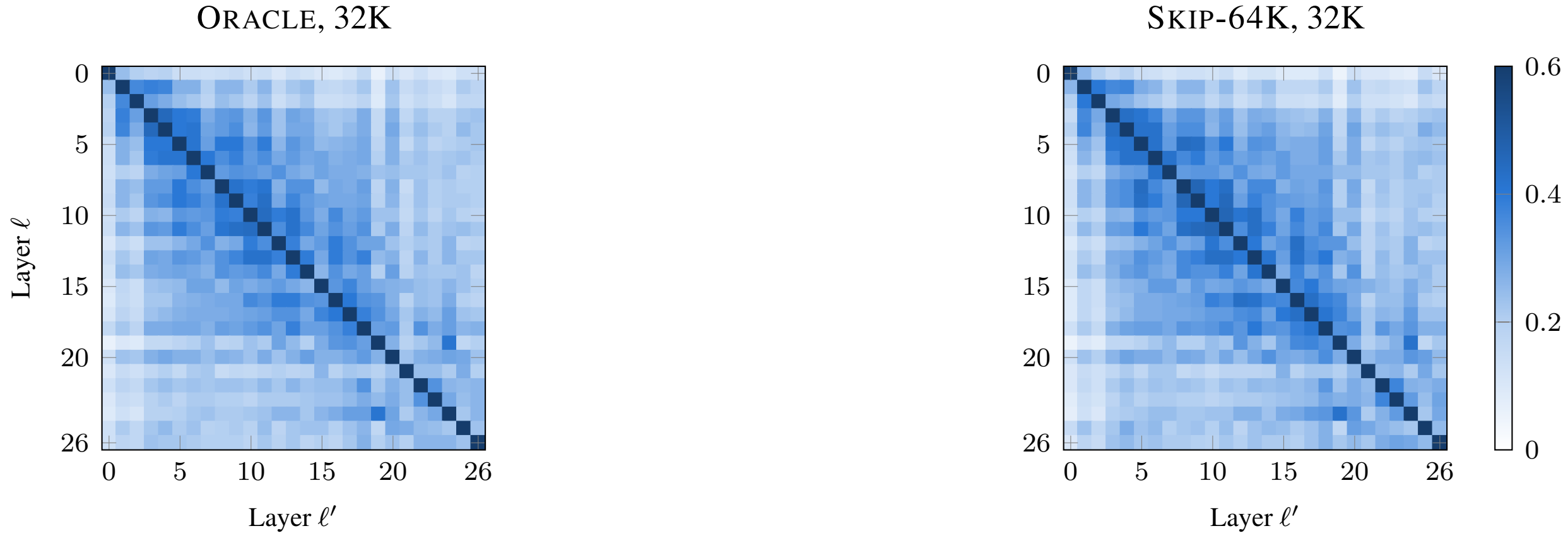
\begin{figure}[h]
\centering
\begin{tikzpicture}
\begin{axis}[width=5.6cm, height=5.6cm, enlargelimits=false, axis on top,
  title={\Oracle{}, 32K}, title style={font=\small},
  xlabel={Layer $\ell'$}, ylabel={Layer $\ell$},
  xtick={0,5,10,15,20,26}, ytick={0,5,10,15,20,26}, y dir=reverse,
  tick label style={font=\scriptsize}, label style={font=\scriptsize},
  colormap={wb}{color=(white) color=(cA!35) color=(cA) color=(cA!50!black)},
  point meta min=0, point meta max=0.6]
\addplot[matrix plot*, mesh/cols=27, point meta=explicit] table[meta index=2] {figs/jaccard_oracle.dat};
\end{axis}
\end{tikzpicture}\hfill
\begin{tikzpicture}
\begin{axis}[width=5.6cm, height=5.6cm, enlargelimits=false, axis on top,
  title={\Skip{64K}, 32K}, title style={font=\small},
  xlabel={Layer $\ell'$}, 
  xtick={0,5,10,15,20,26}, ytick={0,5,10,15,20,26}, y dir=reverse,
  tick label style={font=\scriptsize}, label style={font=\scriptsize},
  colormap={wb}{color=(white) color=(cA!35) color=(cA) color=(cA!50!black)},
  point meta min=0, point meta max=0.6, colorbar, colorbar style={width=0.18cm, tick label style={font=\scriptsize}}]
\addplot[matrix plot*, mesh/cols=27, point meta=explicit] table[meta index=2] {figs/jaccard_skip.dat};
\end{axis}
\end{tikzpicture}
\caption{Jaccard overlap $J_{\ell\ell'}$ of the top-2048 sets of all pairs of layers of DeepSeek-V2-Lite at the same query, at 32K, averaged over the last 1024 decoding steps of four PG-19 documents. Left: \Oracle{} selection. Right: \Skip{64K} at native positions. The diagonal is 1 and is clipped by the colour scale.}
\label{fig:jaccard}
\end{figure}

\paragraph{Overlap between all pairs of layers.}
Figure~\ref{fig:jaccard} gives the Jaccard overlap $J_{\ell\ell'}$ for all pairs of layers at 32K, from selection traces of four PG-19 documents with $k=2048$.
Both matrices have the same structure: overlap is highest between neighbouring layers in the lower half of the model and falls with the distance between layers, and the upper layers overlap little with any other layer.
The mean over all pairs of different layers is 0.25 for \Oracle{} and 0.26 for the indexer at 32K, and 0.39 for both at 8K.
The indexer is \Skip{64K} at native positions; the traces cover the last 1024 decoding steps of each document, and against \Oracle{} its selection has a recall of 0.83 at 8K and 0.73 at 32K.

\paragraph{Index reuse with the trained indexer.}
The table in Figure~\ref{fig:reuse} uses the 400M-token \Skip{64K} indexer in the indexer layers.
Its perplexity rows score the last 2,048 tokens of four PG-19 blocks of 32K, and they are compared only within that table; the passcode rows use the eight samples of the probe, and the RULER rows the 200 samples per task of Table~\ref{tab:ruler}.
With \Oracle{} selection in the indexer layers instead of the trained indexer, the order on perplexity is the same, with 2.3\% for \Gapfill{8} against 8.4\% for \Uniform{8} at $k=2048$.
The advantage of \Gapfill{8} is thus consistent on perplexity and on the in-sample probe, and it disappears on the held-out tasks.
Table~\ref{tab:reuseruler} gives all seven RULER tasks.
\Gapfill{8} is better than \Uniform{8} on multi-value and multi-query, by 7.5 and 5.0 points, and worse on the other five, by 2.0 to 5.0 points, and its mean over the seven tasks is 0.451 against 0.461.
On multi-key-1 and multi-value both placements lose 7 to 29 points against the indexer in all layers, and on single-1, where all layers reach the 0.995 of \Dense{}, they lose 8.0 and 10.0 points, about four unpaired standard errors.
On single-3, multi-key-2 and multi-query they differ from it by at most 6.0 points, which is less than two unpaired standard errors; \Dense{} itself reaches only 0.130 and 0.160 on the last two.
On multi-key-2 \Uniform{8} has the higher exact match of the two placements but a token accuracy of 0.38, against 0.81 for \Gapfill{8} and 0.76 with all layers, so most of its answers are far from correct.

\begin{table}[H]
\caption{Index reuse on RULER at 32K, $k=2048$, 200 samples per task (the samples of Table~\ref{tab:ruler}): exact-match rate with the 400M-token \Skip{64K} indexer in all 27 layers (All) or in 8 layers, the others reusing the nearest earlier selection. Token accuracy in parentheses where it is discussed in the text.}
\label{tab:reuseruler}
\centering
\small
\begin{tabular}{lcccc}
\toprule
Task & \Dense{} & All & \Uniform{8} & \Gapfill{8} \\
\midrule
single-1 (filler text)      & 0.995 & 0.995 & 0.915 & 0.895 \\
single-2 (book text)        & 0.990 & 0.955 & 0.980 & 0.930 \\
single-3 (identifier value) & 0.690 & 0.695 & 0.755 & 0.715 \\
multi-key-1 (four keys)     & 0.680 & 0.585 & 0.340 & 0.300 \\
multi-key-2 (all needles)   & 0.130 (0.85) & 0.035 (0.76) & 0.080 (0.38) & 0.035 (0.81) \\
multi-value (four values)   & 0.545 & 0.230 & 0.085 & 0.160 \\
multi-query (four queries)  & 0.160 & 0.120 & 0.070 & 0.120 \\
\midrule
Mean of the seven tasks & 0.599 & 0.516 & 0.461 & 0.451 \\
\bottomrule
\end{tabular}
\end{table}

\section{Second model: full results}
\label{app:second}

This appendix gives the full results on the second model, the public Qwen3-30B-A3B-Instruct-2507 checkpoint of the Qwen3 family~\citep{qwen3}, whose values for $k=2048$ Table~\ref{tab:pair-main} summarises.
It is a mixture-of-experts model with 48 attention layers, each with 32 query heads and 4 key and value heads, and it uses rotary position encoding; like DeepSeek-V2-Lite it was trained with dense attention, and we keep it frozen.
One indexer with the settings of Section~\ref{sec:setup} is attached to each of the 48 layers, and the positions it selects serve all heads of the layer.
\Skip{64K} is distilled on 8K sequences with range 64K and \LongI{} on 32K sequences with ordinary positions, each with 100M tokens and one run.
The \Skip{64K} run took 16.2 hours on one H200; the \LongI{} run was moved from one card to two during training, so its time is not comparable.

\paragraph{Perplexity.}
Table~\ref{tab:second} gives the \dppl{} of the two indexers with the protocol of Section~\ref{sec:setup}, on four blocks up to 32K, the eight blocks of Table~\ref{tab:pairci} at 64K and two blocks at 128K.
Up to 32K, the training length of \LongI{}, the two are level, as on the first model.
For $k=2048$ the \dppl{} at 32K is 0.73\% for \Skip{64K} and 0.71\% for \LongI{}, and from 8K to 32K the two differ by at most 0.31 points at either budget.
Both retrieve the passcode in eight of eight samples for $k=2048$, as \Dense{} does, and in seven of eight for $k=512$.
The equivalence of Section~\ref{sec:q3} is therefore reproduced on this model.

At 64K the result of the first model is not reproduced.
With its ordinary positions \LongI{} has a lower recall than \Skip{64K}, 0.42 against 0.54, but the same \dppl{} for $k=2048$, 1.76\% against 1.73\%, a paired difference of 0.04 points with an interval from $-0.27$ to $0.43$; for $k=512$ it is 0.50 points behind, with an interval from 0.11 to 0.92.
On DeepSeek-V2-Lite \LongI{} has 5.80\% at 64K on the same number of blocks (Table~\ref{tab:pairci}), so its loss beyond the training length is much smaller on this model.
With its positions compressed into its range, \LongI\compress{} has 0.79\% with recall 0.59, which is 0.94 points better than \Skip{64K}, with an interval from 0.59 to 1.32.
On DeepSeek-V2-Lite the order is the opposite, with \LongI\compress{} 0.26 points behind \Skip{64K} at equal tokens (interval 0.00 to 0.52).
The advantage of position skipping over compression at 64K, which Section~\ref{sec:q3} reports for the first model, is therefore reversed on this one; with one run per indexer, the intervals do not include the variation between training runs.
At 128K, outside both ranges, compression helps both indexers, from 3.72\% to 2.19\% for \Skip{64K} and from 4.08\% to 1.89\% for \LongI{}.

\begin{table}[h]
\caption{Second model, Qwen3-30B-A3B: \dppl{} (\%) over \Dense{} for $k=512$ / $k=2048$, indexers distilled with 100M tokens. Four PG-19 documents up to 32K, eight at 64K and two at 128K. The \compress{} rows compress the indexer positions into the trained range, 64K for \Skip{64K} and 32K for \LongI{}.}
\label{tab:second}
\centering
\small
\setlength{\tabcolsep}{3.5pt}
\resizebox{\linewidth}{!}{%
\begin{tabular}{lccccc}
\toprule
Indexer & 8K & 16K & 32K & 64K & 128K \\
\midrule
\Skip{64K} & $-0.15$ / $-0.49$ & $+1.08$ / $+0.33$ & $+3.15$ / $+0.73$ & $+6.52$ / $+1.73$ & $+9.88$ / $+3.72$ \\
\Skip{64K}\compress{} & & & & & $+7.63$ / $+2.19$ \\
\LongI{} & $+0.11$ / $-0.41$ & $+0.89$ / $+0.02$ & $+2.99$ / $+0.71$ & $+7.02$ / $+1.76$ & $+11.28$ / $+4.08$ \\
\LongI\compress{} & & & & $+5.29$ / $+0.79$ & $+7.63$ / $+1.89$ \\
\bottomrule
\end{tabular}}
\end{table}

\paragraph{A range of 32K.}
\Skip{32K} was distilled on this model as on Llama-3.1 (Appendix~\ref{app:llama128}), with 8K sequences, ids up to 32K and 100M tokens, and evaluated at 64K on the eight blocks of Table~\ref{tab:pairci} and at 128K on four blocks (Table~\ref{tab:q3range}).
At 64K, twice its range, it is level with \Skip{64K} for $k=2048$ with its own ids, a paired difference of $-0.03$ points ($-0.35$ to $+0.36$), and 0.87 points better with compressed ids (0.57 to 1.19), as \LongI\compress{} is on this model.
At 128K, four times its range, it does not fail: with its own ids it has 3.39\%, 0.49 points behind \Skip{64K} (0.14 to 0.94), and with compressed ids 1.37\%, 0.33 points better than \Skip{64K}\compress{} (0.15 to 0.60).
Keeping layers 0 and 1 dense changes these values by less than 0.8 points, to 2.66\% and 1.03\%, so unlike Llama-3.1 this model does not need dense first layers at 128K.
The cost of exceeding the range thus depends on the model: at four times its range \Skip{32K} with its own ids loses 16.75\% on DeepSeek-V2-Lite (two documents, Table~\ref{tab:range}) and 3.39\% on Qwen3, and on Llama-3.1 it fails unless the first two layers are dense; compression reduces the loss on all three.
The four-block values of \Skip{64K} at 128K come from a rerun on these blocks and differ from the two-block values of Table~\ref{tab:second}, 3.72\% and 2.19\% for $k=2048$.

\begin{table}[h]
\caption{Qwen3-30B-A3B beyond the range of \Skip{32K}: \dppl{} (\%) for $k=512$ / $k=2048$, indexers distilled on 8K sequences with 100M tokens. \compress{} compresses the ids into the range, 32K for \Skip{32K} and 64K for \Skip{64K}. 64K: the eight PG-19 blocks of Table~\ref{tab:pairci}, with 95\% bootstrap intervals over blocks. 128K: four blocks. -- is not run or inside the range.}
\label{tab:q3range}
\centering
\small
\setlength{\tabcolsep}{3pt}
\resizebox{\linewidth}{!}{%
\begin{tabular}{lccc}
\toprule
 & 64K, all sparse & 128K, all sparse & 128K, layers 0, 1 dense \\
\midrule
\Skip{32K} & $+7.09$ [$5.30$, $9.08$] / $+1.70$ [$1.11$, $2.40$] & $+9.80$ / $+3.39$ & $+9.36$ / $+2.66$ \\
\Skip{32K}\compress{} & $+5.53$ [$3.85$, $7.36$] / $+0.86$ [$0.41$, $1.39$] & $+5.96$ / $+1.37$ & $+5.95$ / $+1.03$ \\
\Skip{64K} & $+6.52$ [$4.88$, $8.25$] / $+1.73$ [$1.25$, $2.16$] & $+8.04$ / $+2.90$ & -- \\
\Skip{64K}\compress{} & -- & $+6.16$ / $+1.70$ & -- \\
\bottomrule
\end{tabular}}
\end{table}

\paragraph{Retrieval tasks.}
Table~\ref{tab:secondruler} gives four RULER tasks at 32K with $k=2048$ on the same 50 samples per task for all rows, scored as in Appendix~\ref{app:ruler}, and two of them again with 200 samples.
With 50 samples per task \Dense{} and \Skip{64K} answer all 200 questions of the four tasks, and \LongI{} misses one sample of the multi-value task.
With 200 samples every row answers 199 or 200 questions of each task.
The tasks are thus saturated on this model: they show that the 8K-trained indexer loses nothing that they can detect, and they cannot rank the two indexers.
In particular the loss of \Skip{64K} on the multi-value task of Table~\ref{tab:ruler} does not appear here, and the data do not tell whether this is because the model is stronger or because the task is too easy for it.

\begin{table}[h]
\caption{Second model, Qwen3-30B-A3B: samples with an exact match on RULER tasks at 32K with $k=2048$, out of 50 (upper block) or 200 (lower block). The indexers are those of Table~\ref{tab:second}.}
\label{tab:secondruler}
\centering
\small
\begin{tabular}{lccc}
\toprule
Task & \Dense{} & \Skip{64K} & \LongI{} \\
\midrule
single-1 (filler text)    & 50 & 50 & 50 \\
single-2 (book text)      & 50 & 50 & 50 \\
multi-key-1 (four keys)   & 50 & 50 & 50 \\
multi-value (four values) & 50 & 50 & 49 \\
\midrule
multi-key-1, 200 samples   & 199 & 200 & 200 \\
multi-value, 200 samples   & 200 & 200 & 199 \\
\bottomrule
\end{tabular}
\end{table}

\paragraph{Retrieval tasks that are not saturated.}
We therefore ran three further RULER tasks on which \Dense{} is not at the ceiling, on 200 samples per task at 32K with $k=2048$, the same samples for all rows.
Multi-query is defined in Appendix~\ref{app:ruler}.
In variable tracking, a value is assigned to one variable and passed along a chain of four further variables, \texttt{VAR B = VAR A}, hidden in repeated filler sentences, and the answer lists the five variable names.
In common-word extraction, the context is a numbered list of words in which ten words occur 30 times each and all others three times, and the answer lists the ten frequent words.
Each answer thus lists several items in an order fixed by the reference, so a single wrong token makes the exact match zero, and \Dense{} has an exact-match rate of 0.000, 0.030 and 0.000 on the three tasks.
We therefore compare the token accuracy, the fraction of answer-value tokens that are the most likely token, and the answer loss, the mean negative log-likelihood of these tokens, both teacher-forced as in Appendix~\ref{app:ruler}, with paired bootstrap intervals over the 200 samples.

Table~\ref{tab:secondruler2} gives the results.
In token accuracy each indexer differs from \Dense{} by at most 0.6 points on every task.
On variable tracking both indexers are slightly better than \Dense{} on both measures, and on multi-query both are 0.2 points lower in token accuracy, with an interval that only just excludes zero.
On common-word extraction, where the answer depends on word counts over the whole context, the answer loss is clearly higher than with \Dense{}, by 0.185 for \Skip{64K} and 0.187 for \LongI{}, while the token accuracy is 0.5 to 0.6 points lower, with intervals that include zero.
Between \Skip{64K} and \LongI{} every paired interval contains zero, for both measures and on all three tasks, with differences of at most 0.002 in token accuracy.
On this model the 8K-trained indexer thus matches the 32K-trained one on tasks that are not saturated, and both lose answer confidence on the counting task.

\begin{table}[h]
\caption{Second model, Qwen3-30B-A3B: three RULER tasks at 32K with $k=2048$ that are not saturated, 200 samples per task, the same samples in every column, scored by teacher forcing. Upper block: token accuracy over the answer values (higher is better); lower block: mean negative log-likelihood of the answer values (lower is better). The indexer columns give the paired difference to \Dense{} with a 95\% bootstrap interval over samples (20,000 resamples). The indexers are those of Table~\ref{tab:second}.}
\label{tab:secondruler2}
\centering
\small
\setlength{\tabcolsep}{4pt}
\begin{tabular}{lccc}
\toprule
Task & \Dense{} & \Skip{64K} $-$ \Dense{} & \LongI{} $-$ \Dense{} \\
\midrule
\multicolumn{4}{@{}l}{\emph{Token accuracy}} \\
variable tracking     & 0.922 & $+0.006$ [$+0.003$, $+0.009$] & $+0.004$ [$+0.002$, $+0.007$] \\
multi-query           & 0.955 & $-0.002$ [$-0.004$, $-0.000$] & $-0.002$ [$-0.004$, $-0.000$] \\
common-word extraction & 0.760 & $-0.006$ [$-0.012$, $+0.000$] & $-0.005$ [$-0.011$, $+0.001$] \\
\midrule
\multicolumn{4}{@{}l}{\emph{Answer loss}} \\
variable tracking     & 0.579 & $-0.037$ [$-0.044$, $-0.029$] & $-0.035$ [$-0.042$, $-0.028$] \\
multi-query           & 0.267 & $+0.005$ [$-0.001$, $+0.010$] & $+0.004$ [$-0.002$, $+0.010$] \\
common-word extraction & 1.036 & $+0.185$ [$+0.156$, $+0.215$] & $+0.187$ [$+0.156$, $+0.217$] \\
\bottomrule
\end{tabular}
\end{table}

\section{Third model: Llama-3.1-8B}
\label{app:third}

This appendix gives the results on the third model, the public Llama-3.1-8B-Instruct checkpoint~\citep{llama3}, whose values for $k=2048$ Table~\ref{tab:pair-main} summarises.
It is a dense model with 32 attention layers, each with 32 query heads and 8 key and value heads, and rotary position encoding with a native context of 128K tokens; like the other two models it was trained with dense attention, and we keep it frozen.
One indexer with the settings of Section~\ref{sec:setup} is attached to each of the 32 layers, and the positions it selects serve all heads of the layer.
\Skip{64K} and \LongI{} follow the recipe of the Qwen3 runs, each with 100M tokens and one run.
\Skip{128K} uses the same 8K sequences with a range of 128K, the native context of the model, also with 100M tokens and one run.

\paragraph{Perplexity.}
Table~\ref{tab:third} gives the \dppl{} of the indexers on four blocks up to 32K and two at 64K and 128K; the intervals at 64K on eight blocks are in Table~\ref{tab:pairci}.
At 8K all indexers have a lower perplexity than \Dense{} for $k=2048$, by about one percent, as on Qwen3, so differences at that length do not rank them.
Up to 32K \Skip{64K} and \LongI{} differ by at most 0.53 points; \LongI{} is better for $k=2048$ at 32K, 1.72\% against 2.25\%, and \Skip{64K} is better for $k=512$ at every length.
On the eight blocks of Table~\ref{tab:pairci} at 64K, \LongI{} with its ordinary ids fails as on DeepSeek-V2-Lite, with recall 0.32, and \Skip{64K} is 2.04 points better; with compressed ids \LongI\compress{} is 0.47 points better than \Skip{64K} for $k=2048$ (interval 0.30 to 0.65) and level for $k=512$.
Second runs of both recipes, on 16 blocks at 32K, are in Appendix~\ref{app:seeds}.
At 128K every configuration fails, \Skip{64K} with compressed ids included; Appendix~\ref{app:llama128} locates the cause.
\Skip{128K} has the lowest \dppl{} of the three indexers at 32K and 64K for both budgets, 1.29\% and 2.18\% for $k=2048$, but these values rest on four and two blocks and one run, and at 128K, inside its range, it fails as well, with 277.32\%.
All three indexers retrieve the passcode in eight of eight samples for $k=2048$, as \Dense{} does, and in seven of eight for $k=512$.

\begin{table}[h]
\caption{Third model, Llama-3.1-8B: \dppl{} (\%) over \Dense{} for $k=512$ / $k=2048$, indexers distilled with 100M tokens. Four PG-19 documents up to 32K and two at 64K and 128K. The \compress{} rows compress the indexer positions into the trained range, 64K for \Skip{64K} and 32K for \LongI{}; every length is inside the range of \Skip{128K}.}
\label{tab:third}
\centering
\small
\setlength{\tabcolsep}{3.5pt}
\resizebox{\linewidth}{!}{%
\begin{tabular}{lccccc}
\toprule
Indexer & 8K & 16K & 32K & 64K & 128K \\
\midrule
\Skip{64K} & $-0.40$ / $-1.06$ & $+0.67$ / $+0.47$ & $+3.08$ / $+2.25$ & $+6.82$ / $+3.29$ & $+111.28$ / $+60.95$ \\
\Skip{64K}\compress{} & & & & & $+77.56$ / $+107.52$ \\
\LongI{} & $-0.36$ / $-1.05$ & $+1.00$ / $+0.33$ & $+3.37$ / $+1.72$ & $+10.94$ / $+4.24$ & $+80.95$ / $+69.59$ \\
\LongI\compress{} & & & & $+6.92$ / $+2.44$ & $+54.47$ / $+69.50$ \\
\Skip{128K} & $-0.68$ / $-0.97$ & $+0.09$ / $+0.39$ & $+1.03$ / $+1.29$ & $+3.62$ / $+2.18$ & $+311.57$ / $+277.32$ \\
\bottomrule
\end{tabular}}
\end{table}

\paragraph{Retrieval tasks.}
Table~\ref{tab:thirdruler} gives four RULER tasks at 32K with $k=2048$ on 200 samples per task, scored as in Appendix~\ref{app:ruler}, with per-sample scores.
On single-1 and multi-key-1 both indexers are at the level of \Dense{}.
On multi-value both lose about five points against \Dense{}, with 10 and 11 samples right only with \Dense{} against one only with the indexer ($p=0.012$ and $p=0.006$), and they are level with each other, so the multi-value deficit of Table~\ref{tab:ruler} does not appear on this model.

On single-2 \Dense{} reaches only 0.485, below both indexers, and \LongI{} is better than \Skip{64K}, 0.820 against 0.755, with 13 samples right only with \LongI{} and none only with \Skip{64K} ($p=0.0002$).
This task is affected by an artefact of our teacher-forced scoring.
The question asks for all the special magic numbers of a key, and Llama-3.1 answers by listing the numbers that occur in the PG-19 text in the order of the context, with or without its chat template, while our metric scores only the first value of the gold answer.
In all 103 failures of \Dense{} the first digit token is wrong, and the logits of our dense attention path equal those of the reference implementation of the library.
The failures of \LongI{} are a subset of those of \Dense{}, and those of \Skip{64K} with one exception: the indexers drop most distractor numbers from the selection, which raises their score.
The string-match-all metric of RULER with free generation would credit such answers, so we read the single-2 row as a property of the metric and not as a retrieval failure, and do not use it to rank the indexers.

\begin{table}[h]
\caption{Third model, Llama-3.1-8B: exact-match rate on four RULER tasks at 32K with $k=2048$, 200 samples per task, the same samples in every column. Single-2 is affected by a scoring artefact (see text).}
\label{tab:thirdruler}
\centering
\small
\begin{tabular}{lccc}
\toprule
Task & \Dense{} & \Skip{64K} & \LongI{} \\
\midrule
single-1 (filler text)    & 1.000 & 1.000 & 1.000 \\
single-2 (book text)      & 0.485 & 0.755 & 0.820 \\
multi-key-1 (four keys)   & 0.960 & 0.965 & 0.970 \\
multi-value (four values) & 0.975 & 0.930 & 0.925 \\
\bottomrule
\end{tabular}
\end{table}

\paragraph{Serving.}
Decoding throughput in vLLM on this model is reported in Appendix~\ref{app:serving} (Table~\ref{tab:servingl8}); up to 64K the indexer is slower than dense decoding in every measured cell.

\section{Llama-3.1-8B at 128K}
\label{app:llama128}

This appendix examines why every indexer fails on Llama-3.1-8B at 128K (Table~\ref{tab:third}), using \Skip{64K}, $k=2048$ and the two PG-19 blocks of 128K tokens of that table.
We split the queries into four bins by position, 0--32K, 32--64K, 64--96K and 96--128K, and measure in each bin the recall of the indexer against the dense top-2048 when the model runs with dense attention, so that errors do not compound across layers, and the cosine between the mean layer input of the bin and that of the first bin.

\paragraph{The indexer degrades gradually and the selection budget is too small.}
Table~\ref{tab:llama128}, upper block, shows that the recall of \Skip{64K} falls from 0.797 in the first bin to 0.494 in the last, without a step at 64K, the end of its range; on Qwen3 the same indexer recipe falls similarly, from 0.760 to 0.465.
The layer input drifts more on Llama-3.1, to a cosine of 0.798 in the last bin against 0.924 on Qwen3, and the recall falls most in layers 1 to 6, from about 0.75 to between 0.13 and 0.24.
Yet the indexer is not the cause of the failure: \Oracle{}, which selects the 2,048 keys with the largest head-summed dense attention, has a \dppl{} of 144.56\%, worse than the 60.95\% of \Skip{64K}, while at 64K on the same model it costs 3.67\%.
In the first layer the top-2048 keys hold only 43\% of the head-summed attention mass in the last bin, so at this length the attention of the first layers is spread over many more keys than the budget.
A larger budget helps \Oracle{}, which reaches 63.23\% with 4,096 keys and 12.11\% with 8,192, and so does a separate selection of 2,048 keys for each of the eight key-value groups, at 57.39\%, while an attention sink and a window of 512 recent keys do not (151.65\%).

\paragraph{Dense first layers.}
The lower block of Table~\ref{tab:llama128} keeps the first layers dense and the other layers sparse.
With layer 0 dense, \Skip{64K} reaches 9.49\%, and with layers 0 and 1 dense, 2 of 32 layers, it reaches 2.95\%.
In both settings the distilled indexer is better than \Oracle{}, which reaches 31.44\% and 5.37\%.
We read this as a failure that starts in the spread attention of the first two layers and compounds through the depth, rather than a failure of the indexer or of its position range.
The result also suggests that the head-summed attention mass is not the best selection criterion once the deep layers receive inputs that earlier sparse layers have changed, where the indexer, distilled to the full distribution, does better.
Two blocks were evaluated, so these values are indicative.

\paragraph{A range of 128K.}
\Skip{128K} (Appendix~\ref{app:third}) was trained with ids up to 128K, so at this length it needs no compression.
On the same two blocks it does not remove the need for dense first layers (Table~\ref{tab:llama128}, last row).
With all layers sparse its \dppl{} is 277.32\%, worse than the 60.95\% of \Skip{64K}, and with layers 0 and 1 dense it reaches 4.17\%, against 2.95\% for \Skip{64K}.
Its recall in the last query bin, measured in the sparse model itself, is 0.319 with all layers sparse and 0.383 with layers 0 and 1 dense, close to the 0.307 and 0.403 of \Skip{64K}.
A range that covers 128K therefore neither removes the need for the two dense layers nor improves on \Skip{64K} once they are dense; this rests on one run and two blocks, without an interval.

\paragraph{A range of 32K.}
To measure how an indexer degrades beyond its range on this model, \Skip{32K} was distilled with the recipe of \Skip{64K}, 8K sequences and 100M tokens, with ids up to 32K, and evaluated at 64K and 128K, two and four times its range (Table~\ref{tab:l8range}).
At 64K, on the eight blocks of Table~\ref{tab:pairci}, it is 2.00 points behind \Skip{64K} for $k=2048$ (interval 1.38 to 2.77) with its own ids, and level with it when its ids are compressed into 32K, $-0.06$ points ($-0.17$ to $+0.06$); for $k=512$ the compressed indexer remains 0.72 points behind (0.29 to 1.17).
This is the pattern of DeepSeek-V2-Lite (Table~\ref{tab:range}): a moderate loss at twice the range that compression removes.
At 128K, on four blocks, every configuration fails with all layers sparse, as \Skip{64K} does, and \Skip{32K} with its own ids fails most, with 316.38\% for $k=2048$.
With layers 0 and 1 dense, \Skip{32K} reaches 2.72\% (1.57 to 3.81) with its own ids and $-0.97\%$ ($-2.21$ to $+0.29$) with compressed ids, level with \Dense{}; on the first two of these blocks \Skip{64K} reaches 2.95\% and 1.77\%.
A range of 32K, with compression, is therefore enough at 128K on this model once the first two layers are dense.
These 128K values rest on one run, and some are not monotone in $k$, which we attribute to the failure of the selection in that regime.
The first two blocks are those of Table~\ref{tab:llama128}; on them a rerun of \Skip{64K} reproduced the values of Table~\ref{tab:third} to within 0.5 points and its dense-layer value to within 0.04 points, and the four-block values of \Skip{64K} in Table~\ref{tab:l8range} come from this rerun.

\begin{table}[h]
\caption{Llama-3.1-8B beyond the range of \Skip{32K}: \dppl{} (\%) for $k=512$ / $k=2048$, indexers distilled on 8K sequences with 100M tokens. \compress{} compresses the ids into the range, 32K for \Skip{32K} and 64K for \Skip{64K}. 64K: the eight PG-19 blocks of Table~\ref{tab:pairci}, with 95\% bootstrap intervals over blocks. 128K: four blocks, the first two of which are those of Table~\ref{tab:llama128}; $^\dagger$ on those two blocks only. -- is not run or inside the range.}
\label{tab:l8range}
\centering
\small
\setlength{\tabcolsep}{3pt}
\resizebox{\linewidth}{!}{%
\begin{tabular}{lccc}
\toprule
 & 64K, all sparse & 128K, all sparse & 128K, layers 0, 1 dense \\
\midrule
\Skip{32K} & $+11.26$ [$9.25$, $13.58$] / $+4.22$ [$2.93$, $5.59$] & $+84.15$ / $+316.38$ & $+10.96$ / $+2.72$ \\
\Skip{32K}\compress{} & $+6.82$ [$5.46$, $8.46$] / $+2.15$ [$1.12$, $3.40$] & $+68.69$ / $+72.97$ & $+3.23$ / $-0.97$ \\
\Skip{64K} & $+6.11$ [$4.54$, $7.88$] / $+2.21$ [$1.16$, $3.49$] & $+94.71$ / $+53.24$ & -- / $+2.95^\dagger$ \\
\Skip{64K}\compress{} & -- & $+82.60$ / $+120.98$ & -- / $+1.77^\dagger$ \\
\bottomrule
\end{tabular}}
\end{table}

\begin{table}[h]
\caption{Llama-3.1-8B at 128K, $k=2048$, two PG-19 blocks. Upper block: recall of \Skip{64K} against the dense top-2048 with dense attention in the model, and the cosine of the mean layer input of each bin to that of the first bin, both averaged over layers, by query bin; Qwen3-30B-A3B for comparison. Lower block: \dppl{} (\%) with all layers sparse or the first layers dense; \Oracle{} selects by head-summed dense attention, shared by all heads of a layer; -- is not run.}
\label{tab:llama128}
\centering
\small
\setlength{\tabcolsep}{4pt}
\begin{tabular}{lcccc}
\toprule
Query bin & 0--32K & 32--64K & 64--96K & 96--128K \\
\midrule
Recall, Llama-3.1-8B & 0.797 & 0.716 & 0.614 & 0.494 \\
Recall, Qwen3-30B-A3B & 0.760 & 0.625 & 0.556 & 0.465 \\
Input cosine, Llama-3.1-8B & 1 & 0.977 & 0.919 & 0.798 \\
Input cosine, Qwen3-30B-A3B & 1 & 0.980 & 0.946 & 0.924 \\
\midrule
\dppl{} (\%) & All sparse & Layer 0 dense & Layers 0, 1 dense & \\
\midrule
\Skip{64K} & $+60.95$ & $+9.49$ & $+2.95$ & \\
\Oracle{} & $+144.56$ & $+31.44$ & $+5.37$ & \\
\Skip{128K} & $+277.32$ & -- & $+4.17$ & \\
\bottomrule
\end{tabular}
\end{table}

\section{Second training runs and a cost-matched baseline on Llama-3.1-8B}
\label{app:seeds}
Every other configuration in this paper was trained once, so we trained the two main recipes on Llama-3.1-8B a second time and added the cheapest alternative to skipped ids.
All runs use 100M tokens of the data of Section~\ref{sec:setup}; the evaluation uses 16 PG-19 blocks at 32K and the eight blocks of Table~\ref{tab:pairci} at 64K, with the protocol of Appendix~\ref{app:ruler}.
The dense negative log-likelihood of every block agrees exactly with the earlier evaluation, so the second runs can be paired with the first block by block.
\begin{itemize}
\item \Skip{64K}, run 2: the recipe of run 1, with another seed and data order (global batch of six 8K sequences, 2,035 steps).
\item \LongI{}, run 2: another seed and data order, and a global batch of four 32K sequences for 763 steps instead of two for 1,526, so the two runs differ in batch size as well as seed.
\item \ShortI{}: 8K sequences with ordinary ids and the recipe of \Skip{64K}, evaluated with its own ids and with its ids compressed to 8K (by 0.25 at 32K and 0.125 at 64K), at the same training cost as \Skip{64K}.
\item YaRN: run 1 of \LongI{} at 64K with the NTK-by-parts interpolation of \citet{yarn} ($\alpha=1$, $\beta=32$, factor 2) applied to the indexer's rotary positions at inference, instead of compressing its ids.
\end{itemize}
Table~\ref{tab:seeds} gives the \dppl{}.

\begin{table}[H]
\caption{Llama-3.1-8B, indexers distilled with 100M tokens: \dppl{} (\%) over \Dense{} with 95\% bootstrap intervals over blocks, 16 PG-19 blocks at 32K and eight at 64K. \compress{} compresses the ids into the trained length (32K for \LongI{}, 8K for \ShortI{}). Run 2 of \LongI{} used twice the batch and half the steps of run 1. The 64K values of run 1 are those of Table~\ref{tab:pairci}.}
\label{tab:seeds}
\centering
\small
\resizebox{\linewidth}{!}{%
\begin{tabular}{lcccc}
\toprule
 & \multicolumn{2}{c}{32K} & \multicolumn{2}{c}{64K} \\
\cmidrule(lr){2-3}\cmidrule(lr){4-5}
Indexer & $k=512$ & $k=2048$ & $k=512$ & $k=2048$ \\
\midrule
\Skip{64K}, run 1 & $3.36$ [$2.42$, $4.41$] & $1.63$ [$0.96$, $2.29$] & $6.11$ [$4.54$, $7.88$] & $2.21$ [$1.16$, $3.49$] \\
\Skip{64K}, run 2 & $3.29$ [$2.44$, $4.24$] & $1.58$ [$0.91$, $2.25$] & $5.55$ [$4.16$, $7.02$] & $2.07$ [$1.02$, $3.34$] \\
\LongI{}, run 1 & $3.85$ [$3.04$, $4.80$] & $1.42$ [$0.79$, $2.04$] & $11.77$ [$9.59$, $14.38$] & $4.26$ [$2.98$, $5.64$] \\
\LongI\compress{}, run 1 & in range & in range & $6.40$ [$5.13$, $7.90$] & $1.74$ [$0.76$, $2.87$] \\
\LongI{}, run 1, YaRN & in range & in range & $6.54$ [$5.07$, $8.12$] & $1.75$ [$0.75$, $2.88$] \\
\LongI{}, run 2 & $2.39$ [$1.60$, $3.40$] & $0.99$ [$0.47$, $1.50$] & $7.39$ [$5.37$, $10.20$] & $2.21$ [$1.11$, $3.51$] \\
\LongI\compress{}, run 2 & in range & in range & $3.24$ [$2.04$, $4.74$] & $0.98$ [$0.13$, $1.88$] \\
\ShortI{} & $2759$ [$2469$, $3126$] & $132.65$ [$111.42$, $159.66$] & $7311$ [$5881$, $8934$] & $1313$ [$1130$, $1528$] \\
\ShortI\compress{} & $5.25$ [$4.33$, $6.25$] & $1.70$ [$1.02$, $2.36$] & $8.78$ [$7.16$, $10.72$] & $2.45$ [$1.37$, $3.66$] \\
\bottomrule
\end{tabular}}
\end{table}

\paragraph{Variation between runs.}
With the same recipe, run 2 of \Skip{64K} differs from run 1 by $-0.07$ and $-0.06$ points at 32K for $k=512$ and $k=2048$, with intervals that contain zero, and by $-0.56$ ($-0.91$ to $-0.30$) and $-0.15$ ($-0.25$ to $-0.05$) at 64K.
Run 2 of \LongI{} is better than run 1 in every cell: at 32K by 1.46 and 0.43 points, and at 64K with compressed ids by 3.17 and 0.77 points (intervals $-4.13$ to $-2.27$ and $-1.20$ to $-0.35$).
These changes are three to twenty times the change between the two runs of \Skip{64K}, so they more likely come from the larger batch than from the seed; with one run per batch size we cannot separate the two.

\paragraph{Skipped ids against distillation at 32K.}
On the 16 blocks at 32K, run 1 of \LongI{} is 0.49 points worse than run 1 of \Skip{64K} for $k=512$ (interval 0.29 to 0.73) and 0.21 points better for $k=2048$ (0.06 to 0.36), inside the margin of $\pm 0.5$ points.
Run 2 of \LongI{} is better than run 2 of \Skip{64K} by 0.90 and 0.59 points at 32K (intervals 0.41 to 1.40 and 0.28 to 0.89) and, with compressed ids, by 2.31 and 1.09 points at 64K (1.39 to 3.32 and 0.52 to 1.65).
On this model the gap between the two recipes thus depends on the batch of the baseline, and with a batch of four sequences distillation at 32K is ahead by more than the margin.
The batch of \Skip{64K} was not varied.

\paragraph{Compressing an 8K indexer.}
\ShortI{} with its own ids fails at both lengths, and compressing its ids into 8K restores it, as on DeepSeek-V2-Lite (Table~\ref{tab:main}).
At the same training cost, \ShortI\compress{} is level with \Skip{64K} for $k=2048$ at 32K ($+0.07$, interval $-0.12$ to $+0.27$) and 0.24 points worse at 64K (0.07 to 0.41), but for $k=512$ it is 1.89 and 2.67 points worse, on every block.
On Llama-3.1 the distances seen in training therefore matter mainly at the small budget.

\paragraph{YaRN instead of compression.}
YaRN and compressed ids give the same \dppl{} for run 1 of \LongI{} at 64K: the paired difference is $+0.13$ points for $k=512$ (interval $-0.23$ to $+0.48$) and $0.00$ for $k=2048$ ($-0.08$ to $+0.08$), and both are far better than the indexer's own ids.
The choice of interpolation thus does not change the comparison between compressed and skipped ids.

\section{A block-level indexer distilled at long context}
\label{app:flashmemory}

\citet{flashmemory} distill a block-level indexer at long context.
We re-implemented its design for DeepSeek-V2-Lite: blocks of four tokens with keys compressed as in DeepSeek-V4, a selection refreshed every $\tau=64$ queries, a window of 128 recent tokens inside the budget, and a focal loss on labels derived from the dense attention; it is distilled on 64K sequences with ordinary positions and 100M tokens, the budget of \LongLL{}.
Twelve details that the paper does not state are our guesses, among them learned block keys, one indexer per layer instead of three shared ones, and a fixed budget of $k$ tokens instead of a score threshold, so the results describe our re-implementation and not the original method.

On the PG-19 blocks of Table~\ref{tab:pairci}, the re-implementation has a \dppl{} of 4.56\% (interval 3.35 to 5.85) at 32K and 4.92\% (2.89 to 7.30) at 64K for $k=2048$, with recall 0.41 and 0.40, and 11.79\% and 13.71\% for $k=512$.
It is 3.20 points (2.16 to 4.28) behind \Skip{64K} at 32K and 3.81 points (2.07 to 5.77) at 64K, in paired differences on the same blocks.
Refreshing the selection at every query instead of every 64 lowers the \dppl{} by about half a point, to 4.04\% and 4.47\%.

\paragraph{Retrieval tasks.}
Table~\ref{tab:flashruler} gives four RULER tasks at 32K with $k=2048$ on the 200 samples of Table~\ref{tab:ruler100}, next to the three 100M-token indexers of that table.
With the selection refreshed every 64 queries, as in the design, the re-implementation fails on all four tasks: it reaches 0.475 on single-1 and 0.005 on single-2, and no sample of multi-key-1 or multi-value is an exact match.
This is far below what its deficit of 3.2 points on PG-19 would suggest.
Refreshing the selection at every query recovers single-1 completely, with 0.995 against 0.940 for \Skip{64K}, and 12 samples right only with the re-implementation against one only with \Skip{64K} ($p=0.003$).
It recovers single-2 only in part, to 0.635 against 0.955, with 68 samples right only with \Skip{64K} against 4 ($p<10^{-15}$).
Sharing one selection across 64 queries is thus the main cause of the failure.
Our data do not tell whether the remaining gap on single-2 comes from the block granularity of the design or from a detail of our re-implementation.
A refresh at every query also removes the saving that the block design is meant to provide, and it was not run on multi-key-1 and multi-value.
These results describe our re-implementation, and we make no claim about the retrieval accuracy of the original method.

\begin{table}[h]
\caption{Block-level re-implementation of \citet{flashmemory} on DeepSeek-V2-Lite: exact-match rate with token accuracy in parentheses on four RULER tasks at 32K with $k=2048$, 200 samples per task, the same samples in every column. $\tau$ is the number of queries that share one selection; $\tau=64$ is the design, $\tau=1$ refreshes at every query. The other columns are the 100M-token indexers of Table~\ref{tab:ruler100}; -- is not run.}
\label{tab:flashruler}
\centering
\small
\setlength{\tabcolsep}{4pt}
\resizebox{\linewidth}{!}{%
\begin{tabular}{lccccc}
\toprule
Task & \Skip{64K} & \LongLL{} & \LongI{} & Block, $\tau=64$ & Block, $\tau=1$ \\
\midrule
single-1 (filler text)    & 0.940 (.991) & 0.970 (.996) & 0.920 (.989) & 0.475 (.757) & 0.995 (.999) \\
single-2 (book text)      & 0.955 (.994) & 0.975 (.996) & 0.975 (.996) & 0.005 (.068) & 0.635 (.926) \\
multi-key-1 (four keys)   & 0.570 (.936) & 0.600 (.941) & 0.580 (.939) & 0.000 (.044) & -- \\
multi-value (four values) & 0.335 (.966) & 0.300 (.965) & 0.340 (.967) & 0.000 (.674) & -- \\
\bottomrule
\end{tabular}}
\end{table}

\section{Ablations: position range and position mechanism}
\label{app:range}

Table~\ref{tab:pos} gives the position-mechanism ablation and the 100M-token \Skip{64K} run that Section~\ref{sec:q4} summarises.
The \DistBias{} indexer is defined in Appendix~\ref{app:details}.

\begin{table}[h]
\caption{Position-mechanism ablation on DeepSeek-V2-Lite. Each length column gives \dppl{} (\%) for $k=512$ / $k=2048$; the passcode column gives the passcodes retrieved at 32K out of eight for $k=512$, $k=2048$. All rows are distilled on 8K sequences; the first uses 400M tokens and the others 100M. KL is the final training loss (Eq.~\ref{eq:kl}). Four blocks at 8K to 32K, two at 64K.}
\label{tab:pos}
\centering
\small
\setlength{\tabcolsep}{2.6pt}
\begin{tabular}{lcccccc}
\toprule
 & & \multicolumn{4}{c}{\dppl{} (\%), $k=512$ / $k=2048$} & \\
\cmidrule(lr){3-6}
Indexer & KL & 8K & 16K & 32K & 64K & Passcode \\
\midrule
\Skip{64K}, 400M & 0.041 & $+1.51$ / $+0.32$ & $+2.35$ / $+0.34$ & $+5.17$ / $+1.23$ & $+5.92$ / $+0.88$ & 8, 7 \\
\Skip{64K}, 100M & 0.055 & $+1.66$ / $+0.29$ & $+2.47$ / $+0.45$ & $+5.42$ / $+1.53$ & $+6.17$ / $+1.27$ & 7, 7 \\
\RopeEight{} & 0.076 & $+4.98$ / $+0.57$ & $+8.60$ / $+2.15$ & $+20.3$ / $+5.53$ & $+57.0$ / $+10.5$ & 7, 7 \\
\NoPE{} & 0.316 & $+3794$ / $+18.5$ & $+10476$ / $+266$ & $+13962$ / $+3550$ & $+23838$ / $+7938$ & 0, 0 \\
\DistBias{} & 0.112 & $+2.26$ / $+0.35$ & $+2.56$ / $+0.81$ & $+6.53$ / $+2.27$ & $+7.99$ / $+2.19$ & 8, 6 \\
\bottomrule
\end{tabular}
\end{table}

Table~\ref{tab:range} gives the values of the position-range ablation of Section~\ref{sec:q4}.
The negative values at 128K for $k=2048$ mean that the sparse model has lower perplexity than \Dense{}.
They come from two documents at a length close to the limit of the model's own position encoding, and we treat them as indicative only.
On eight blocks \Skip{64K} has $+27.07$ / $+13.76$ at 128K and $-0.30$ / $-3.82$ with compression, and \Skip{128K} has $-0.14$ / $-3.06$ (Table~\ref{tab:pg19ci128}), so the failure outside the range and the recovery are confirmed, and the size of both differs from the two-document values below.

\begin{table}[h]
\caption{Position-range ablation: \dppl{} (\%) for $k=512$ / $k=2048$. All indexers are distilled on 8K sequences with 400M tokens. The \compress{} rows compress indexer positions into the trained range; a dash marks a length inside the range. Two documents at 64K and 128K. For reference, \LongI{} at 128K gives $+22.9$ / $+10.9$ and \LongI\compress{} gives $+4.76$ / $-2.83$.}
\label{tab:range}
\centering
\small
\begin{tabular}{lccc}
\toprule
Length & \Skip{32K} & \Skip{64K} & \Skip{128K} \\
\midrule
8K   & $+1.58$ / $+0.32$ & $+1.51$ / $+0.32$ & $+1.71$ / $+0.31$ \\
16K  & $+2.40$ / $+0.46$ & $+2.35$ / $+0.34$ & $+2.18$ / $+0.45$ \\
32K  & $+5.35$ / $+1.04$ & $+5.17$ / $+1.23$ & $+4.91$ / $+1.21$ \\
64K  & $+25.31$ / $+7.09$ & $+5.92$ / $+0.88$ & $+5.60$ / $+1.04$ \\
128K & $+56.52$ / $+16.75$ & $+44.59$ / $+21.18$ & $+1.54$ / $-3.32$ \\
\midrule
64K \compress{}  & $+6.69$ / $+1.55$ & --- & --- \\
128K \compress{} & $+5.73$ / $-2.87$ & $+1.95$ / $-4.04$ & --- \\
\bottomrule
\end{tabular}
\end{table}

\section{Post-hoc sparse key codec}
\label{app:codec}

When index memory matters more than scan time, the key can be compressed further by a sparse code fitted after distillation.
Let $e_s = \mathrm{LN}(W_K h_s)$ be the key before rotation.
A dictionary of $m$ unit-norm atoms $E \in \mathbb{R}^{m \times d}$, an encoder $(W_e, o_e)$ and an offset $o_0$ are fitted per layer so that
\begin{equation}
z_s = \operatorname{top}_n\!\big(W_e (e_s - o_0) + o_e\big), \qquad \hat e_s = E^{\top} \relu(z_s) + o_0
\label{eq:codec}
\end{equation}
minimises the squared reconstruction error of $e_s$, where $\operatorname{top}_n$ keeps the $n$ largest entries and sets the others to zero.
The index stores the $n$ pairs of atom id and weight.
At scan time the key is decoded, rotated and scored with the unchanged query side of the indexer.
The codec is trained with forward passes of the frozen model only, without attention targets.
It reduces memory but not scan time, since the decoded key is scored as before.
The sparse-coding recipe follows \citet{lexico}, who showed for the cache of keys and values that a sparse code over a dictionary of about four thousand atoms stays accurate at memory budgets where 2-bit quantisation fails.
We apply it to indexer keys, and Appendix~\ref{app:negative} compares it with a sparse code distilled end to end.
In a bounded search we did not find a post-hoc sparse code for the keys of a sparse-attention indexer.

\paragraph{Evaluation.}
We fit dictionaries of $m=4096$ atoms to the keys of the 100M-token \Skip{64K} indexer, using 25M tokens and 2.4 hours on one A100, and name the result \Codec{$n$} for $n$ stored pairs per key.
Table~\ref{tab:codec} compares the codec with the exact keys.
\Codec{8} is within 0.4 points of the exact keys at every length, for example 1.66\% against 1.27\% at 64K for $k=2048$.
With a 12-bit atom id and a 4-bit weight per pair it takes 17 bytes per token; Table~\ref{tab:codec} keeps the weights in floating point, and Table~\ref{tab:quante2e} in Appendix~\ref{app:cascade} repeats the comparison on 16 blocks with 4-bit weights, where the codec costs about 0.2 points at $k=2048$ and loses passcodes at $k=512$.
The codec therefore offers a further memory reduction over three-bit keys at a small loss for the larger budget, but it does not reduce the scan time, because each key is decoded before it is scored.

\begin{table}[h]
\caption{Post-hoc sparse key codec on the 100M-token \Skip{64K} indexer: \dppl{} (\%) for $k=512$ / $k=2048$, with recall against \Oracle{} in parentheses. \Codec{$n$} stores $n$ pairs of atom id and weight per key. Bytes assume 12-bit ids and 4-bit weights. Four documents at 8K, two at 32K and 64K.}
\label{tab:codec}
\centering
\small
\setlength{\tabcolsep}{3.5pt}
\begin{tabular}{lrccc}
\toprule
Keys & Bytes & 8K & 32K & 64K \\
\midrule
Exact, 16-bit float & 128 & $+1.66$ / $+0.29$ (.75/.81) & $+2.84$ / $+0.55$ (.69/.70) & $+6.17$ / $+1.27$ (.64/.66) \\
\Codec{32} & 65 & $+1.89$ / $+0.49$ (.74/.80) & $+2.67$ / $+0.46$ (.67/.69) & $+6.34$ / $+1.06$ (.63/.65) \\
\Codec{8} & 17 & $+1.62$ / $+0.31$ (.69/.76) & $+3.01$ / $+0.61$ (.63/.66) & $+6.25$ / $+1.66$ (.59/.62) \\
\bottomrule
\end{tabular}
\end{table}

\section{Key formats and a low-bit first stage with rescoring}
\label{app:cascade}

Quantising the query to 8 bits changes none of the three-bit and four-bit values of this appendix at three decimals.

Table~\ref{tab:formats} gives the attention mass captured by every key format that Section~\ref{sec:q5} discusses, with and without a second stage that rescores the best 4,096 candidates with full-precision keys.
A value of 1.000 means that the compressed index selects positions of equal total attention.
One-bit keys are not accurate enough as a first stage.

\begin{table}[h]
\caption{Key formats for the \Skip{64K} indexer: attention mass captured by the top-2048 selection made with compressed keys, relative to the selection made with full-precision keys. ``Direct'' uses the compressed keys alone. ``Rescored'' takes the best 4,096 candidates under the compressed keys and rescores them with full-precision keys. Bytes are per token and per layer and include a two-byte scale where one is used. Two PG-19 documents per length; every eighth of the last 512 query positions is evaluated.}
\label{tab:formats}
\centering
\small
\begin{tabular}{lrcccc}
\toprule
 & & \multicolumn{2}{c}{32K} & \multicolumn{2}{c}{64K} \\
\cmidrule(lr){3-4}\cmidrule(lr){5-6}
Key format & Bytes & Direct & Rescored & Direct & Rescored \\
\midrule
16-bit float (reference) & 128 & 1.000 & 1.000 & 1.000 & 1.000 \\
4 bits per coordinate & 34 & 0.995 & 1.000 & 0.994 & 1.000 \\
3 bits per coordinate & 26 & 0.977 & 0.995 & 0.972 & 0.994 \\
2 bits per coordinate & 18 & 0.728 & 0.814 & 0.756 & 0.844 \\
Random projection, 128 sign bits & 18 & 0.930 & 0.973 & 0.905 & 0.957 \\
Trained code, 128 sign bits & 18 & 0.917 & 0.966 & 0.879 & 0.940 \\
Random rotation, 64 sign bits & 10 & 0.865 & 0.930 & 0.812 & 0.891 \\
Trained code, 64 sign bits & 10 & 0.834 & 0.912 & 0.747 & 0.844 \\
\bottomrule
\end{tabular}
\end{table}

Table~\ref{tab:cascade} gives the dependence on the number of candidates $M$ that the first stage passes to the full-precision rescoring.
With sign bits and a random rotation, retaining 0.97 of the attention mass requires $M$ of about a quarter of the context, 8,192 candidates at 32K and 16,384 at 64K.
The rescoring then costs a quarter of a full scan, and the two-stage design saves little.
Four-bit keys need no second stage.

\begin{table}[h]
\caption{Two-stage selection with a training-free first stage on the \Skip{64K} indexer, $k=2048$. Each cell gives the fraction of the full-precision top-2048 inside the $M$ candidates and, after the slash, the attention mass of the final selection relative to the full-precision selection. Two PG-19 documents per length.}
\label{tab:cascade}
\centering
\small
\begin{tabular}{llcccc}
\toprule
Length & First-stage key & Direct mass & $M=4096$ & $M=8192$ & $M=16384$ \\
\midrule
32K & 64 sign bits & 0.806 & 0.559 / 0.900 & 0.778 / 0.962 & 0.948 / 0.991 \\
32K & 64 sign bits, random rotation & 0.865 & 0.576 / 0.930 & 0.790 / 0.974 & 0.952 / 0.996 \\
32K & 4 bits per coordinate & 0.995 & 0.993 / 1.000 & 1.000 / 1.000 & 1.000 / 1.000 \\
64K & 64 sign bits & 0.712 & 0.438 / 0.821 & 0.631 / 0.909 & 0.825 / 0.967 \\
64K & 64 sign bits, random rotation & 0.812 & 0.494 / 0.891 & 0.692 / 0.949 & 0.869 / 0.983 \\
64K & 4 bits per coordinate & 0.994 & 0.991 / 1.000 & 1.000 / 1.000 & 1.000 / 1.000 \\
\bottomrule
\end{tabular}
\end{table}

The trained one-bit codes of Table~\ref{tab:formats} were obtained by self-distillation.
A linear projection followed by a sign function, trained with a straight-through gradient for 400 steps on 8K sequences, was fitted so that its scores reproduce the ranking of the full-precision indexer.
The trained codes are worse than random projections of the same width at both lengths.
One possible reason is that they were trained at 8K and evaluated at 32K and 64K; we did not test this.

\paragraph{End-to-end effect of compact keys.}
The attention mass of Table~\ref{tab:formats} is measured on the attention of the dense model, and it does not show what the model loses when every layer selects from compressed keys.
Table~\ref{tab:quante2e} therefore evaluates the whole model with the selection made from four-bit keys, from three-bit keys and from \Codec{8} of Appendix~\ref{app:codec}.
Each compact store is run on the same PG-19 blocks as the 16-bit keys of the same indexer, and the table gives the paired difference in \dppl{} in percentage points.
The 16-bit rows of the upper block repeat the \Skip{64K} column of Table~\ref{tab:pg19ci} on an A100 instead of an H200, with 8 of its 16 blocks at 64K, and differ from it by up to 0.09 points from 8K to 32K.

Four-bit keys are equivalent to 16-bit keys at $k=2048$: the difference is at most 0.05 points, and its interval contains zero at every length.
At $k=512$ they cost 0.06 to 0.22 points.
Three-bit keys cost more than their attention mass of 0.97 suggests.
At $k=2048$ the difference is at most 0.15 points up to 32K, with intervals that contain zero, and it is 0.46 points at 64K (interval 0.25 to 0.65), where \dppl{} rises from 0.80\% to 1.26\%.
At $k=512$ it is 0.81 points at 32K and 1.48 points at 64K.
The recall against \Oracle{} at 64K and $k=2048$ is 0.70 with 16-bit keys, 0.66 with four bits and 0.55 with three bits, so the three-bit selection replaces many positions, and at the small budget the replacements are not of equal value.
Four bits are therefore the format that can replace 16-bit keys without a measured loss, and three bits save a further 8 bytes per token at a cost that grows with the length.

\Codec{8} costs 0.16 to 0.22 points at $k=2048$, and quantising its weights to four bits changes this by at most 0.07 points, so the 17 bytes per token are reached at that cost.
At $k=512$ its perplexity differences are at most 0.32 points, and the interval excludes zero in one of eight cells.
Passcode retrieval separates the formats more clearly, although on eight samples.
With three-bit and four-bit keys the model retrieves at least as many passcodes as with 16-bit keys.
With \Codec{8} at $k=512$ it retrieves 5 of 8, and 4 of 8 with 4-bit weights, against 7 of 8 with the exact keys of the same indexer; at $k=2048$ all three rows retrieve 7.
We therefore regard the codec as usable only at the larger budget.

\begin{table}[h]
\caption{End-to-end effect of compact key stores on DeepSeek-V2-Lite. The rows ``16-bit float'' give \dppl{} (\%) over \Dense{} for $k=512$ and $k=2048$. The rows below them give the paired difference to that row in percentage points on the same blocks, and an asterisk marks a difference whose 95\% bootstrap interval over blocks excludes zero, without correction for the number of comparisons. The upper block uses the 400M-token \Skip{64K} indexer, and the lower block the 100M-token \Skip{64K} indexer to which the codec was fitted. Integer keys are scored with an 8-bit query, and the two \Codec{8} rows store the weights in floating point and in 4 bits. 16 blocks of PG-19 per length, 8 at 64K. The last column is the number of passcodes out of 8 that are retrieved exactly at 32K, for $k=512$ / $k=2048$; \Dense{} retrieves 8.}
\label{tab:quante2e}
\centering
\small
\setlength{\tabcolsep}{2.6pt}
\begin{tabular}{lrccccccccc}
\toprule
 & & \multicolumn{2}{c}{8K} & \multicolumn{2}{c}{16K} & \multicolumn{2}{c}{32K} & \multicolumn{2}{c}{64K} & \\
\cmidrule(lr){3-4}\cmidrule(lr){5-6}\cmidrule(lr){7-8}\cmidrule(lr){9-10}
Keys & Bytes & 512 & 2048 & 512 & 2048 & 512 & 2048 & 512 & 2048 & Passcode \\
\midrule
16-bit float & 128 & $+3.10$ & $+0.66$ & $+3.91$ & $+0.83$ & $+5.01$ & $+1.27$ & $+5.06$ & $+0.80$ & 8 / 7 \\
4-bit keys & 34 & $+0.06$ & $+0.05$ & $+0.17^{*}$ & $0.00$ & $+0.21^{*}$ & $+0.03$ & $+0.22^{*}$ & $+0.03$ & 8 / 7 \\
3-bit keys & 26 & $+0.71^{*}$ & $+0.09$ & $+0.16$ & $0.00$ & $+0.81^{*}$ & $+0.15$ & $+1.48^{*}$ & $+0.46^{*}$ & 8 / 8 \\
\midrule
16-bit float & 128 & $+3.27$ & $+0.66$ & $+3.92$ & $+0.89$ & $+4.88$ & $+1.27$ & $+5.24$ & $+1.06$ & 7 / 7 \\
\Codec{8}, float & & $+0.12$ & $+0.16$ & $+0.01$ & $+0.22^{*}$ & $+0.21$ & $+0.22^{*}$ & $-0.28$ & $+0.19$ & 5 / 7 \\
\Codec{8}, 4-bit & 17 & $+0.32$ & $+0.23^{*}$ & $+0.07$ & $+0.20$ & $+0.28^{*}$ & $+0.24^{*}$ & $-0.23$ & $+0.16^{*}$ & 4 / 7 \\
\bottomrule
\end{tabular}
\end{table}

\section{Scan kernel}
\label{app:kernel}

Table~\ref{tab:kernel} gives the time of one scan, which scores all $L$ cached keys for $Q$ queries with Eq.~\ref{eq:score} and stops before the top-$k$ step, as Section~\ref{sec:q5} summarises.
The codes are stored as unsigned values with an offset of $2^{b-1}$, and the offset is removed with one precomputed sum per query head.
The fused kernel on three-bit keys takes 21.7\,$\mu$s for one query over 131K keys, against 34.5\,$\mu$s for a cuBLAS half-precision matrix multiplication followed by a reduction kernel, and 2.26\,ms against 4.95\,ms for 32 queries over 1M keys.
Unpacking the keys and reusing the matrix multiplication is slower than this baseline, as is the fused kernel with a floating-point query.
For a single query the selection step that follows the scan takes longer than the fused scan at both lengths, so in that regime the top-$k$ step and not the scan dominates the indexer's kernel time.

\begin{table}[h]
\caption{Compact index: bytes per token and layer, attention mass of the top-2048 selection at 64K relative to full-precision keys (Table~\ref{tab:formats}), and time of one scan in microseconds on one A100 for $L$ cached keys and $Q$ queries ($H=16$, $d=64$, median of 50 to 200 runs). All rows compute Eq.~\ref{eq:score} for every key, on synthetic keys, with outputs checked against a CPU reference; the last row is the selection step alone.}
\label{tab:kernel}
\centering
\small
\setlength{\tabcolsep}{3.5pt}
\begin{tabular}{lrcrrrr}
\toprule
 & & & \multicolumn{2}{c}{$L=131$K} & \multicolumn{2}{c}{$L=1$M} \\
\cmidrule(lr){4-5}\cmidrule(lr){6-7}
Scan & Bytes & Mass & $Q=1$ & $Q=32$ & $Q=1$ & $Q=32$ \\
\midrule
PyTorch, 16-bit matrix multiplication & 128 & 1.000 & 143 & 912 & 347 & 6705 \\
cuBLAS 16-bit matrix multiplication + reduce kernel & 128 & 1.000 & 34.5 & 660 & 180 & 4952 \\
4-bit keys, unpack, then the same matrix multiplication & 34 & 0.994 & 87.5 & 704 & 580 & 5325 \\
4-bit keys, fused kernel, 32-bit float query & 34 & 0.994 & 47.3 & 1223 & 284 & 8182 \\
4-bit keys, fused kernel, 8-bit query (ours) & 34 & 0.994 & \textbf{21.2} & \textbf{332} & \textbf{94.0} & \textbf{2253} \\
3-bit keys, fused kernel, 8-bit query (ours) & 26 & 0.972 & 21.7 & 346 & 95.7 & 2264 \\
\midrule
Top-$k$ over the scores, $k=2048$ & & & 127 & 191 & 288 & 966 \\
\bottomrule
\end{tabular}
\end{table}

\section{Step time of the whole model with the indexer}
\label{app:decode}

Table~\ref{tab:kernel} times the scan alone on synthetic keys.
This appendix measures what the indexer and the compact index change in the time of one step of the whole model.

\paragraph{Setup.}
We use DeepSeek-V2-Lite in 16-bit precision on one A100, one PG-19 document, the \Skip{64K} indexer and $k=2048$.
The first $L$ tokens of the document are processed with indexer-selected sparse attention, which leaves in every layer the cached key and value rows of the model and the indexer keys, both as 16-bit floats and as packed three-bit codes.
A step then runs the next $Q$ tokens through all 27 layers against these caches and returns their logits.
With $Q=1$ a step is one decoding step of one sequence.
With $Q>1$ it is a prefill chunk of $Q$ tokens, in which every query sees the keys up to its own position, and it has the same cost profile as $Q$ sequences decoded on a shared prompt.
We compare three configurations: \Dense{} attends to all cached rows and uses no indexer; the indexer with 16-bit keys uses the strongest floating-point scan of Table~\ref{tab:kernel}; the indexer with three-bit keys uses the fused integer kernel.
Both indexer configurations select with an exact radix select on the GPU and then gather the $k$ selected rows for attention.
For fewer than about 200K keys a kernel launch with one thread per key occupies only part of the GPU, so in this measurement the scan kernels additionally split the queries across thread blocks; the scores are bit-identical.
The implementation is a PyTorch reference and not a serving engine: everything outside the scan and the selection consists of unmodified library modules.
The 128K rows are timing points only, because the positions of the indexer were trained up to 64K.

\begin{table}[H]
\caption{Time of one step of DeepSeek-V2-Lite in milliseconds (27 layers, one A100, median of 20 repetitions) for $L$ cached tokens and $Q$ new tokens per step, $k=2048$. The two indexer rows differ only in the key format and the scan; each cell is the faster of two variants of the radix select. The ratio rows divide the row named first by the three-bit row, so values above 1 mean that three-bit keys are faster. $^\dagger$Timing only: the indexer was trained for positions up to 64K.}
\label{tab:decode}
\centering
\small
\begin{tabular}{llrrrrr}
\toprule
$L$ & Configuration & $Q=1$ & $Q=8$ & $Q=32$ & $Q=128$ & $Q=512$ \\
\midrule
32K & \Dense{} & 72.3 & 99.0 & 127.0 & 177.1 & 360.0 \\
 & Indexer, 16-bit keys & 83.6 & 113.9 & 141.7 & 168.6 & 283.5 \\
 & Indexer, 3-bit keys & 83.3 & 112.6 & 140.9 & 159.5 & 250.9 \\
 & \quad \Dense{} / 3-bit & 0.87 & 0.88 & 0.90 & 1.11 & 1.43 \\
 & \quad 16-bit / 3-bit & 1.00 & 1.01 & 1.01 & 1.06 & 1.13 \\
\midrule
64K & \Dense{} & 70.8 & 94.8 & 136.8 & 239.1 & 615.9 \\
 & Indexer, 16-bit keys & 85.7 & 110.2 & 140.5 & 189.5 & 361.5 \\
 & Indexer, 3-bit keys & 84.7 & 109.2 & 140.7 & 170.7 & 296.4 \\
 & \quad \Dense{} / 3-bit & 0.84 & 0.87 & 0.97 & 1.40 & 2.08 \\
 & \quad 16-bit / 3-bit & 1.01 & 1.01 & 1.00 & 1.11 & 1.22 \\
\midrule
128K$^\dagger$ & \Dense{} & 68.9 & 94.9 & 164.5 & 364.2 & 1109 \\
 & Indexer, 16-bit keys & 83.1 & 112.3 & 143.0 & 228.9 & 516.6 \\
 & Indexer, 3-bit keys & 82.3 & 108.0 & 137.4 & 193.2 & 383.9 \\
 & \quad \Dense{} / 3-bit & 0.84 & 0.88 & 1.20 & 1.89 & 2.89 \\
 & \quad 16-bit / 3-bit & 1.01 & 1.04 & 1.04 & 1.19 & 1.35 \\
\bottomrule
\end{tabular}
\end{table}

\paragraph{Many queries per step.}
The indexer shortens a step when the step carries many queries, and the three-bit index then adds to the gain.
At $Q=512$ the step with three-bit keys takes 250.9\,ms against 360.0\,ms for \Dense{} at 32K and 296.4\,ms against 615.9\,ms at 64K, which is 1.43 and 2.08 times faster (Table~\ref{tab:decode}).
Against the indexer with 16-bit keys it is 1.13 and 1.22 times faster.
Table~\ref{tab:decodestage} shows where the difference comes from: the scan is the largest stage of the indexer at this $Q$, and the fused kernel takes 70.7\,ms for it at 64K where the floating-point scan takes 135.7\,ms.
The ratio of 1.9 inside the model agrees with the isolated measurement of Table~\ref{tab:kernel}.

\begin{table}[H]
\caption{Stages of the step of Table~\ref{tab:decode} in milliseconds, summed over the 27 layers. The stages were timed in separate repetitions that wait for the GPU after every stage, so their sum is slightly above the step time. ``Query'' is the query projection of the indexer and ``attention'' includes gathering the selected rows. ``Other'' is the feed-forward blocks and the query, key and output projections of the model. $^\dagger$Timing only.}
\label{tab:decodestage}
\centering
\small
\begin{tabular}{lrlrrrrr}
\toprule
 & & & \multicolumn{3}{c}{Indexer} & & \\
\cmidrule(lr){4-6}
$L$ & $Q$ & Configuration & Query & Scan & Top-$k$ & Attention & Other \\
\midrule
32K & 512 & \Dense{} & -- & -- & -- & 246.5 & 127.3 \\
32K & 512 & 16-bit keys & 15.1 & 70.1 & 8.4 & 80.4 & 127.2 \\
32K & 512 & 3-bit keys & 14.8 & 37.6 & 8.6 & 80.4 & 125.0 \\
\midrule
64K & 1 & \Dense{} & -- & -- & -- & 12.7 & 63.1 \\
64K & 1 & 16-bit keys & 13.7 & 1.8 & 3.5 & 10.7 & 63.0 \\
64K & 1 & 3-bit keys & 13.1 & 1.0 & 3.5 & 10.5 & 62.2 \\
\midrule
64K & 512 & \Dense{} & -- & -- & -- & 496.7 & 136.2 \\
64K & 512 & 16-bit keys & 15.1 & 135.7 & 19.2 & 79.7 & 129.2 \\
64K & 512 & 3-bit keys & 15.2 & 70.7 & 19.3 & 79.9 & 128.5 \\
\midrule
128K$^\dagger$ & 512 & \Dense{} & -- & -- & -- & 987.8 & 138.7 \\
128K$^\dagger$ & 512 & 16-bit keys & 15.0 & 270.2 & 39.1 & 81.0 & 128.5 \\
128K$^\dagger$ & 512 & 3-bit keys & 15.0 & 138.6 & 39.3 & 81.2 & 126.7 \\
\bottomrule
\end{tabular}
\end{table}

\paragraph{One query per step.}
For a single query the step with the indexer is slower than \Dense{} at every length that we measured, 82 to 86\,ms against 69 to 72\,ms.
Dense attention of one query over 32K to 128K cached rows costs 0.4 to 0.8\,ms per layer on this model.
The indexer replaces it with attention on 2,048 gathered rows, which costs 0.4\,ms per layer, and adds 0.5\,ms per layer for its query projection, which is library overhead, and 0.1 to 0.3\,ms for the scan and the selection.
About 62\,ms of the step are spent in the feed-forward blocks and the projections, which are identical in the three configurations.
The three-bit index therefore changes the time of a single-query step by about 1\%, and its benefit in that regime is the memory of the index.
Inside a serving engine the outcome depends on the dense baseline: in vLLM~0.8.5 the indexer shortens single-stream decoding from 32K on, whereas in vLLM~0.32.0, whose dense decoding is much faster, a single stream is slower with the indexer at every length (Appendix~\ref{app:serving}).

\paragraph{Selection quality inside the model.}
The same run checks the three-bit selection on the keys and queries that occur in the model.
At $Q=512$ the positions selected with three-bit keys hold 0.967 of the attention mass held by the full-precision selection at 32K and 0.969 at 64K, against 0.977 and 0.972 in Table~\ref{tab:formats}, which was measured on other documents.
Only 0.67 of the two sets of positions coincide, so the positions that differ carry little attention.
This is a consistency check on 512 tokens of one document whose caches were built with full-precision selections; the end-to-end \dppl{} with three-bit keys is given in Table~\ref{tab:quante2e}.

\section{Decoding throughput inside a serving engine}
\label{app:serving}

Appendix~\ref{app:decode} times single steps in a PyTorch reference.
This appendix measures decoding throughput when the indexer runs inside vLLM, with DeepSeek-V2-Lite, the \Skip{64K} indexer in all 27 layers and $k=2048$.
The primary measurement uses vLLM~0.32.0 on an H200, against both dense decoding kernels of the engine, FlashMLA and Triton, and against FlashMLA with an 8-bit cache; the same engine on an A100 and an earlier measurement with vLLM~0.8.5 are kept as secondary results because they show how much the ratio depends on the dense baseline.
A port of the plugin to grouped-query attention is measured with Qwen3-30B-A3B and Llama-3.1-8B on the H200.

\paragraph{The plugin.}
The indexer is loaded as an attention plugin of the engine's MLA backend and changes only decoding; prefill is the engine's own.
For every new token the plugin computes its indexer key and stores it at the token's cache slot, as 16-bit floats or as packed three-bit codes.
In a decoding step it scores the keys of all running sequences in one kernel through the engine's block table, selects 2,048 keys per sequence with an exact radix select, and attends to the selected cache rows with a separate sparse attention kernel.
The fused version of this path issues about eight kernels per layer: one projection of the layer input for key and query, one kernel that normalises, rotates and stores the new keys, one that prepares the query, the scan, the select, and two kernels of attention on the selected rows.
For vLLM~0.32.0 the fused path was ported to decoding with CUDA graphs.
The key stores are allocated after the engine has sized its key-value cache, with one row per cache slot, so they lie outside the memory fraction that the engine is given: 128 bytes per token and layer with 16-bit keys and 26 bytes with three-bit keys.
With the same cache size, the indexer configurations therefore use slightly more memory than \Dense{}.

\paragraph{Protocol.}
A client sends 1, 4 or 16 streaming requests at once, each with a PG-19 prompt of the given length and 256 greedily decoded tokens, and we report the generated tokens per second of all streams while all of them decode, as the mean of two repetitions.
\Dense{} is the stock engine without the plugin.
On the A100, with 16 streams at 64K the 16 prefills take about 77\,s, so no interval exists in which all streams decode, and this cell cannot be measured. On the H200 every cell is therefore measured after an untimed round of the same prompts with one generated token and with prefix caching on, so that the timed prefills hit the cache and all streams decode together; with 16 streams at 64K the window then holds about 4,000 of the 4,096 generated tokens.

\paragraph{vLLM 0.32.0 on an H200.}
Table~\ref{tab:servingh200} compares the indexer with the engine's two dense decoding kernels on one H200: FlashMLA, the default on Hopper GPUs, and the Triton MLA kernel, which the plugin replaces.
The table also gives the plugin with the top-$k$ step of the A100 measurement, which reads the full width of the score buffer in one thread block per sequence, and with a top-$k$ step that reads only the sequence's own length with a cluster of thread blocks (``length-aware select'').
The length-aware select removes about 1.4\,ms per step for one stream (7.34 to 5.92\,ms per token with three-bit keys at 32K) and raises the throughput with 16 streams by 8 to 9\%; with 16-bit keys, a scan with the query held in shared memory raises it at 64K with 16 streams from 566 to 934 tokens per second.
With three-bit keys and the length-aware select, the indexer is faster than Triton dense decoding in every cell with 4 or 16 streams, by 2.3 times at 64K with 16 streams, and for one stream at 32K and 64K.
Against FlashMLA with its default 16-bit cache it is faster only at 64K with 16 streams, by 8.5\% (1118 against 1030 tokens per second); it is 11\% slower at 32K with 16 streams and 10\% slower for one stream at 64K (5.97 against 5.38\,ms per token). With an 8-bit cache FlashMLA is faster still, as the next paragraph shows.
Teacher-forced checks of one stream per length and configuration find no token outside 0.5 logits of the reference's first choice for either dense kernel and at most 0.4\% with the indexer, for both select variants and with prefix-cache hits; the indexer's outputs differ between repetitions, which we attribute to the order in which the select writes ids and hence to the summation order in 16-bit attention.

\begin{table}[H]
\caption{Decoding throughput in vLLM~0.32.0 on one H200 (generated tokens per second of all streams, mean of two repetitions, protocol with an untimed warm-up round and prefix caching; DeepSeek-V2-Lite, \Skip{64K} indexer, $k=2048$). FlashMLA and Triton: the engine's dense decoding kernels. Indexer: plugin on the Triton backend with the full-width select of the A100 measurement or the length-aware select.}
\label{tab:servingh200}
\centering
\small
\setlength{\tabcolsep}{3.5pt}
\begin{tabular}{lrrrrrrrrr}
\toprule
 & \multicolumn{3}{c}{8K} & \multicolumn{3}{c}{32K} & \multicolumn{3}{c}{64K} \\
\cmidrule(lr){2-4}\cmidrule(lr){5-7}\cmidrule(lr){8-10}
Streams & 1 & 4 & 16 & 1 & 4 & 16 & 1 & 4 & 16 \\
\midrule
Dense, FlashMLA & 227 & 626 & 1600 & 198 & 535 & 1250 & 191 & 482 & 1030 \\
Dense, Triton & 179 & 407 & 914 & 164 & 344 & 667 & 151 & 290 & 487 \\
Indexer, 16-bit keys & 128 & 365 & 928 & 130 & 341 & 722 & 127 & 299 & 566 \\
Indexer, 3-bit keys & 137 & 388 & 1018 & 137 & 388 & 1029 & 138 & 392 & 1029 \\
\quad + length-aware select, 16-bit & 169 & 448 & 1079 & 169 & 437 & 1010 & 167 & 419 & 934 \\
\quad + length-aware select, 3-bit & 170 & 450 & 1105 & 171 & 453 & 1111 & 171 & 457 & \textbf{1118} \\
\bottomrule
\end{tabular}
\end{table}

\paragraph{FlashMLA's sparse kernels and an 8-bit cache.}
The comparison above leaves two questions open: whether the indexer's own sparse attention kernel limits it, and how it compares with dense decoding from an 8-bit cache, which halves the bytes that dense decoding reads.
We therefore ran, in one further session on one H200, seven configurations with the same model, indexer, select and protocol (Table~\ref{tab:servingfp8}).
\emph{FlashMLA, 16-bit cache} and \emph{FlashMLA, 8-bit cache} are the engine's dense decoding with its default cache and with the cache stored as 8-bit floating point (FP8).
\emph{Own kernel} is the indexer configuration of Table~\ref{tab:servingh200} with three-bit keys and the length-aware select, run again.
In \emph{FlashMLA sparse prefill} the selected positions are passed to FlashMLA's sparse attention kernel for prefill, which reads the 16-bit rows and needs the 16 query heads padded to 64.
In \emph{FlashMLA sparse decode, FP8 copy} they are passed to FlashMLA's sparse decoding kernel, which reads only an FP8 layout; the plugin therefore keeps an FP8 copy of each layer's cache beside the engine's 16-bit cache, 656 bytes per token and layer (512 FP8 values, four scales and the 64 rotary values in 16 bits), written whenever the engine writes its cache.
This copy lies outside the engine's memory fraction, so these configurations ran with the engine's fraction lowered from 0.85 to 0.52; all 16 streams at 64K still decoded together.

Dense decoding with the 8-bit cache is the fastest configuration for one stream at every length from 8K to 64K, 3.90 and 4.10\,ms per token at 32K and 64K, and with 16 streams at 32K.
The best indexer configuration, three-bit keys with FlashMLA's sparse decoding kernel, is faster than it only at 64K with 16 streams, by 4.8\% in throughput (1320 against 1260 tokens per second) and 3.6\% in time per token; for one stream at 64K it takes 5.75 against 4.10\,ms.
Our own kernel is 11\% slower than the 8-bit dense baseline in that cell.
FlashMLA's sparse prefill kernel is slower than our kernel for one stream (6.84 against 5.97\,ms at 64K) and faster with 16 streams (1240 against 1120 tokens per second at 64K).
In this session dense decoding with the 16-bit cache reached 998 tokens per second at 64K with 16 streams, against 1030 in Table~\ref{tab:servingh200}, while our kernel repeated its value within 0.2\%; a margin of a few percent is thus of the order of the variation of this cell between sessions.
The teacher-forced check shows a quality difference between the two FP8 paths: at 64K, 14 of 256 tokens generated from the dense 8-bit cache lie outside 0.5 logits of the 16-bit reference's first choice, against 1 of 256 for the indexer reading 2,048 rows of its FP8 copy.
A plausible reason is that the indexer reads few rows in FP8 and selects them with its own keys, but the check covers one stream of 256 tokens per length, and we did not measure perplexity with an FP8 cache.
Against the strongest dense baseline the indexer thus gains about 5\% in one setting, long contexts with many streams, and it does so at a memory cost: with the copy, the indexer configuration holds about 1.6 times the cache memory per token of 16-bit dense decoding, while dense FP8 decoding stores a single 8-bit cache.
A backend that stores the cache once in FP8, such as the sparse MLA backend of vLLM, would avoid this cost; we did not port the plugin to it.

\begin{table}[H]
\caption{Dense FlashMLA with a 16-bit and an 8-bit (FP8) cache against the indexer with three sparse attention paths, in vLLM~0.32.0 on one H200 in one session (DeepSeek-V2-Lite, \Skip{64K}, $k=2048$, length-aware select, protocol of Table~\ref{tab:servingh200}). Keys: format of the indexer keys. ms: median time between two tokens; tok/s: generated tokens per second of all streams; both the mean of two repetitions. Off: generated tokens of one stream, out of 256, that lie outside 0.5 logits of the first choice of the 16-bit Hugging Face model, at 32K / 64K. The two FlashMLA sparse decode rows ran with the engine's memory fraction at 0.52 instead of 0.85 because of their FP8 copy of the cache. Bold: fastest configuration of each column.}
\label{tab:servingfp8}
\centering
\small
\setlength{\tabcolsep}{3pt}
\begin{tabular}{llrrrrrrrrc}
\toprule
 & & \multicolumn{4}{c}{32K} & \multicolumn{4}{c}{64K} & \\
\cmidrule(lr){3-6}\cmidrule(lr){7-10}
 & & \multicolumn{2}{c}{1 stream} & \multicolumn{2}{c}{16 streams} & \multicolumn{2}{c}{1 stream} & \multicolumn{2}{c}{16 streams} & \\
Configuration & Keys & ms & tok/s & ms & tok/s & ms & tok/s & ms & tok/s & Off \\
\midrule
FlashMLA, 16-bit cache & -- & 5.17 & 196 & 13.48 & 1270 & 5.43 & 189 & 17.59 & 998 & 0 / 0 \\
FlashMLA, 8-bit cache & -- & \textbf{3.90} & \textbf{260} & \textbf{11.42} & \textbf{1498} & \textbf{4.10} & \textbf{252} & 14.27 & 1260 & 5 / 14 \\
Own kernel & 3-bit & 5.93 & 170 & 15.35 & 1105 & 5.97 & 171 & 15.96 & 1120 & 0 / 0 \\
FlashMLA sparse prefill & 3-bit & 6.78 & 149 & 13.76 & 1242 & 6.84 & 149 & 14.45 & 1240 & 1 / 0 \\
FlashMLA sparse prefill & 16-bit & 6.80 & 149 & 14.96 & 1142 & 6.94 & 147 & 16.85 & 1056 & 1 / 1 \\
FlashMLA sparse decode, FP8 copy & 3-bit & 5.71 & 177 & 13.08 & 1310 & 5.75 & 178 & \textbf{13.75} & \textbf{1320} & 1 / 1 \\
FlashMLA sparse decode, FP8 copy & 16-bit & 5.74 & 176 & 14.23 & 1202 & 5.88 & 175 & 16.06 & 1108 & 1 / 4 \\
\bottomrule
\end{tabular}
\end{table}

\paragraph{A model with grouped-query attention.}
Qwen3-30B-A3B stores 4 key and 4 value heads of 128 dimensions in each of its 48 layers, 2\,KB per token and layer and 96\,KB per token, against 1.15\,KB per token and layer for DeepSeek-V2-Lite.
Its dense decoding therefore reads more memory per step, and the indexer can pay off earlier.
We ported the plugin to the engine's FlashAttention backend: the indexer scores and selects as before, with the length-aware select, and a split-$k$ sparse attention kernel reads the 2,048 selected rows of all key-value heads directly from the paged cache, so that one selection serves the 32 query heads of a layer.
The indexer is the 100M-token \Skip{64K} of Table~\ref{tab:second}.
The protocol is that of the H200 measurement above, and 16 streams at 64K do not fit next to the model's weights on one card.
Table~\ref{tab:servingq3} gives the throughput.
For one stream the indexer takes 7.6 to 7.9\,ms per token at every length, against 4.8\,ms at 8K and 6.4\,ms at 64K for dense decoding, so it is 37\% slower at 8K and 19\% slower at 64K.
With three-bit keys it is faster than dense decoding at 32K with 16 streams, by 19\% (882 against 739 tokens per second), and at 64K with 4 streams, by 13\% (362 against 320); with 16-bit keys the gains there are 6\% and 3\%.
On DeepSeek-V2-Lite the indexer beats FlashMLA only at 64K with 16 streams, which is consistent with a larger cache moving the point where the indexer pays off to fewer streams.
Teacher-forced checks of one stream per length find no generated token outside 0.5 logits of the reference's first choice, for \Dense{} and for both key formats.

\begin{table}[H]
\caption{Decoding throughput of Qwen3-30B-A3B in vLLM~0.32.0 on one H200 (generated tokens per second of all streams, mean of two repetitions, protocol of Table~\ref{tab:servingh200}). \Dense{}: the engine's FlashAttention backend. Indexer: \Skip{64K} with 100M training tokens, $k=2048$, length-aware select and split-$k$ sparse attention. The ratio in parentheses is to \Dense{}. 16 streams at 64K do not fit in memory.}
\label{tab:servingq3}
\centering
\small
\begin{tabular}{rrrrr}
\toprule
Length & Streams & \Dense{} & Indexer, 16-bit keys & Indexer, 3-bit keys \\
\midrule
8K & 1 & 209.1 & 131.1 (0.63) & 131.4 (0.63) \\
8K & 4 & 545.4 & 369.3 (0.68) & 368.3 (0.68) \\
8K & 16 & 1215.0 & 874.6 (0.72) & 902.9 (0.74) \\
32K & 1 & 183.6 & 130.7 (0.71) & 131.5 (0.72) \\
32K & 4 & 417.6 & 348.2 (0.83) & 363.0 (0.87) \\
32K & 16 & 738.7 & 783.8 (1.06) & 881.6 (\textbf{1.19}) \\
64K & 1 & 161.2 & 128.1 (0.79) & 131.2 (0.81) \\
64K & 4 & 320.0 & 329.9 (1.03) & 361.5 (\textbf{1.13}) \\
64K & 16 & --- & --- & --- \\
\bottomrule
\end{tabular}
\end{table}

\paragraph{A dense model with grouped-query attention.}
Llama-3.1-8B stores 8 key and 8 value heads of 128 dimensions in each of its 32 layers, 4\,KB per token and layer.
We measured it with the same grouped-query port and the 100M-token \Skip{64K} of Llama-3.1 in all 32 layers without compression of its ids, three-bit or 16-bit indexer keys and $k=2048$, under the protocol above; 16 streams were not measured at 64K and beyond, and at 128K the prompt has 130,816 tokens so that the 256 generated tokens fit in the native context.
Table~\ref{tab:servingl8} gives the throughput.
For one stream the indexer decodes about 92 tokens per second at every length, 10.9 to 11.1\,ms per token, while dense decoding falls from 171 tokens per second at 8K to 131 at 64K and 104 at 128K, 5.9 to 9.9\,ms per token.
The indexer path thus adds about 5\,ms per step at 8K, more than the 1.9\,ms by which dense decoding slows down from 8K to 64K, so its ratio to \Dense{} rises with the length, from 0.54 at 8K to 0.88 at 128K for one stream with three-bit keys, but stays below one.
With 4 streams at 64K three-bit keys reach 279 tokens per second against 298 for \Dense{}, a ratio of 0.94, the closest cell; with 16 streams the ratio is 0.36 at 8K and 0.81 at 32K.
With 4 streams the time per step of the indexer stays near 15\,ms while that of dense decoding grows from 13.8\,ms at 64K to 18.0\,ms at 96K and 21.8\,ms at 128K, so three-bit keys are 1.21 times faster than \Dense{} at 96K and 1.46 times at 128K, and 16-bit keys 1.11 and 1.31 times.
On this model the indexer thus pays off only with several streams beyond 64K; for one stream it is slower at every measured length.
The configuration measured here keeps all layers sparse and the ids uncompressed, and its perplexity fails at 128K (Tables~\ref{tab:third} and~\ref{tab:l8range}); the configuration that works there keeps layers 0 and 1 dense and compresses the ids, and Table~\ref{tab:servingl8dep} measures it.
Teacher-forced checks of one stream per length up to 64K, not repeated at 96K and 128K, find no generated token outside 0.5 logits of the reference's first choice, for \Dense{} and for both key formats; the reference's first choice is generated for 0.977 to 1.000 of the tokens.

\begin{table}[H]
\caption{Decoding throughput of Llama-3.1-8B in vLLM~0.32.0 on one H200 (generated tokens per second of all streams, mean of two repetitions, 256 generated tokens, protocol of Table~\ref{tab:servingh200}). \Dense{}: the stock engine with its FlashAttention backend. Indexer: grouped-query port of the plugin, 100M-token \Skip{64K} in all layers with uncompressed ids, $k=2048$; at 128K this configuration fails in perplexity (Table~\ref{tab:l8range}). The ratio in parentheses is to \Dense{}. 16 streams were not measured at 64K and beyond; 96K and 128K were measured in a later session, with a prompt of 130,816 tokens at 128K.}
\label{tab:servingl8}
\centering
\small
\begin{tabular}{rrrrr}
\toprule
Length & Streams & \Dense{} & Indexer, 16-bit keys & Indexer, 3-bit keys \\
\midrule
8K & 1 & 171.4 & 92.1 (0.54) & 92.2 (0.54) \\
8K & 4 & 602.6 & 276.0 (0.46) & 276.3 (0.46) \\
8K & 16 & 1667.8 & 598.7 (0.36) & 607.3 (0.36) \\
32K & 1 & 150.9 & 92.2 (0.61) & 92.5 (0.61) \\
32K & 4 & 421.4 & 271.4 (0.64) & 277.5 (0.66) \\
32K & 16 & 749.5 & 576.3 (0.77) & 609.6 (0.81) \\
64K & 1 & 131.4 & 91.6 (0.70) & 92.8 (0.71) \\
64K & 4 & 297.6 & 264.7 (0.89) & 278.9 (0.94) \\
64K & 16 & --- & --- & --- \\
96K & 1 & 116.8 & 90.0 (0.77) & 92.9 (0.80) \\
96K & 4 & 230.7 & 257.1 (\textbf{1.11}) & 278.4 (\textbf{1.21}) \\
128K & 1 & 104.5 & 89.2 (0.85) & 92.5 (0.88) \\
128K & 4 & 190.3 & 249.8 (\textbf{1.31}) & 276.9 (\textbf{1.46}) \\
\bottomrule
\end{tabular}
\end{table}

\paragraph{The deployable configuration of Llama-3.1-8B.}
The configuration whose perplexity holds at 128K differs from the one above in two switches: layers 0 and 1 attend densely, and the indexer's rotary positions are scaled by 0.5 (Table~\ref{tab:l8range}).
We measured it at 96K and 128K with one and four streams, three repetitions per cell, with the same 100M-token \Skip{64K}, $k=2048$ and both key formats, and repeated \Dense{} in the same session; for 128K with four streams we also measured \Dense{} with an FP8 key-value cache.
Table~\ref{tab:servingl8dep} gives the throughput.
The repeated \Dense{} cells lie within 0.3\% of those in Table~\ref{tab:servingl8}, and the two dense layers cost at most 2.7\% of the indexer's throughput.
With four streams three-bit keys are 1.19 times faster than \Dense{} at 96K and 1.41 times at 128K, and 1.18 times faster than \Dense{} with an FP8 cache at 128K; with one stream they remain slower, at 0.81 and 0.89.
The deployable configuration therefore keeps the gain of the all-sparse one, so the speed in Table~\ref{tab:servingl8} at 96K and 128K is not bought with the quality loss.
Teacher-forced checks at 128K of one stream in each repetition compare the generated tokens with a Hugging Face reference that applies the same two switches: with either key format the reference's first choice is generated for 0.988 of the 256 tokens and every other token lies within 0.5 logits of it; for \Dense{} the fraction is 0.980, with no token outside 0.5 logits.
A reference without the two switches agrees on only 0.969 of the tokens in an 8K check, which confirms that both switches take effect in the engine.

\begin{table}[H]
\caption{Decoding throughput of the deployable configuration of Llama-3.1-8B in vLLM~0.32.0 on one H200 (generated tokens per second of all streams, mean of three repetitions with the range across them, 256 generated tokens). Indexer: 100M-token \Skip{64K}, $k=2048$, layers 0 and 1 dense, indexer positions scaled by 0.5. \Dense{}: stock engine, BF16 cache, repeated in the same session. The ratio in parentheses is to \Dense{}. With four streams at 128K, \Dense{} with an FP8 key-value cache decodes 228.7 tokens per second (224.3 to 231.9).}
\label{tab:servingl8dep}
\centering
\small
\begin{tabular}{rrrrr}
\toprule
Length & Streams & \Dense{} & Indexer, 16-bit keys & Indexer, 3-bit keys \\
\midrule
96K & 1 & 116.6 [116.4, 117.0] & 91.3 [91.1, 91.5] (0.78) & 93.9 [93.7, 94.2] (0.81) \\
96K & 4 & 230.9 [230.2, 231.9] & 255.9 [255.5, 256.5] (\textbf{1.11}) & 274.0 [273.4, 275.1] (\textbf{1.19}) \\
128K & 1 & 104.5 [104.4, 104.6] & 90.5 [90.2, 90.8] (0.87) & 93.0 [92.7, 93.5] (0.89) \\
128K & 4 & 190.9 [190.5, 191.6] & 244.4 [244.2, 244.7] (\textbf{1.28}) & 269.4 [268.2, 271.6] (\textbf{1.41}) \\
\bottomrule
\end{tabular}
\end{table}

\paragraph{vLLM 0.32.0 on an A100.}
Table~\ref{tab:serving32full} gives the throughput on one A100.
The stock engine of this version decodes with full CUDA graphs and a split-KV Triton MLA kernel, and in the same cells it is 2.5 to 7.0 times faster than the stock engine of vLLM~0.8.5 for a single stream and 1.0 to 3.3 times faster with 4 or 16 streams (Table~\ref{tab:serving}).
Against this baseline the indexer is slower for a single stream at every length, at 0.67 to 0.87 of the stock throughput, and with 16-bit keys it is slower in every cell.
With three-bit keys it is faster at 32K with 4 and 16 streams, 1.02 and 1.28 times, and at 64K with 4 streams, 1.21 times.
For a single stream the time between two tokens is 7.5\,ms at 8K and 9.1\,ms at 64K for the stock engine, and 10.4\,ms at every length for the indexer with three-bit keys.
The flat time shows that the indexer removes the growth of decoding time with the context, but it adds a fixed cost of about 3\,ms per step, which consists of the full-width scan, the top-$k$ step in a single thread block per sequence and the separate attention kernel.
Dense decoding of this model is cheap because its multi-head latent attention stores only 1.15\,KB per token and layer, so even at 64K the dense read of one stream costs less than this fixed cost.
The indexer pays off only when the dense read grows with both the context and the number of streams, and three-bit keys are needed there because the scan of the stored keys then dominates.
On the H200 the length-aware select removes most of this fixed cost (Table~\ref{tab:servingh200}).
We checked the generated tokens of one stream per length and configuration by teacher forcing them through the Hugging Face implementation of the model: in the stock configuration every token is the reference's first choice or within 0.5 logits of it, and with the indexer at most 0.4\% of the tokens are not.

\begin{table}[H]
\caption{Decoding throughput in vLLM~0.32.0 on one A100 (generated tokens per second of all streams, mean of two repetitions; DeepSeek-V2-Lite, \Skip{64K} indexer, $k=2048$, fused path with CUDA graphs). The ratio in parentheses is to \Dense{}, the stock engine. The 64K cell with 16 streams cannot be measured with this protocol.}
\label{tab:serving32full}
\centering
\small
\begin{tabular}{rrrrr}
\toprule
Length & Streams & \Dense{} & Indexer, 16-bit keys & Indexer, 3-bit keys \\
\midrule
8K & 1 & 133.3 & 89.3 (0.67) & 96.4 (0.72) \\
8K & 4 & 266.6 & 230.8 (0.87) & 241.4 (0.91) \\
8K & 16 & 569.3 & 507.7 (0.89) & 556.7 (0.98) \\
32K & 1 & 123.7 & 89.3 (0.72) & 96.6 (0.78) \\
32K & 4 & 232.0 & 207.9 (0.90) & 237.7 (1.02) \\
32K & 16 & 411.0 & 384.8 (0.94) & 525.2 (1.28) \\
64K & 1 & 111.5 & 86.1 (0.77) & 96.7 (0.87) \\
64K & 4 & 194.8 & 181.4 (0.93) & 236.4 (1.21) \\
64K & 16 & -- & -- & -- \\
\bottomrule
\end{tabular}
\end{table}

\paragraph{vLLM 0.8.5, an older engine.}
Table~\ref{tab:serving} gives the same grid in vLLM~0.8.5 on one A100, where the stock engine decodes with a slower Triton kernel and without full CUDA graphs.
Without FlashAttention~3, as on this card, vLLM~0.8.5 passes a non-contiguous view of the prefill output to its kernel that merges attention over cached and new tokens, so every prefill chunk after the first is computed with wrong values and the model generates degenerate text, also in the stock configuration.
The plugin passes contiguous copies in every configuration, \Dense{} included.
We checked the generated tokens of one stream per length and configuration by teacher forcing them through the Hugging Face implementation: in both dense configurations every generated token is the reference's first choice or within 0.5 logits of it, and with the indexer at most 0.8\% of the tokens are not.
\emph{PyTorch dense} gathers all cached rows and attends to them with batched matrix products, and shows how much of a ratio is due to the stock kernel.

\begin{table}[H]
\caption{Decoding throughput in vLLM~0.8.5 on one A100 (generated tokens per second of all streams, mean of two repetitions; DeepSeek-V2-Lite, \Skip{64K} indexer, $k=2048$). \Dense{}: stock engine. PyTorch dense: all cached rows attended with batched matrix products. Indexer: fused path with 16-bit or three-bit keys; the ratio in parentheses is to \Dense{}. These ratios are not comparable with Table~\ref{tab:serving32full}, whose stock engine is up to 7.0 times faster.}
\label{tab:serving}
\centering
\small
\begin{tabular}{rrrrrr}
\toprule
Length & Streams & \Dense{} & PyTorch dense & Indexer, 16-bit keys & Indexer, 3-bit keys \\
\midrule
8K & 1 & 53.3 & 58.5 & 55.0 (1.03) & 54.5 (1.02) \\
8K & 4 & 195.7 & 155.4 & 203.0 (1.04) & 202.9 (1.04) \\
8K & 16 & 555.6 & 291.6 & 477.7 (0.86) & 522.5 (0.94) \\
32K & 1 & 27.9 & 56.8 & 54.4 (1.95) & 54.0 (1.94) \\
32K & 4 & 98.4 & 67.8 & 182.8 (1.86) & 203.3 (2.07) \\
32K & 16 & 320.8 & 100.1 & 377.0 (1.18) & 494.6 (1.54) \\
64K & 1 & 16.0 & 40.6 & 54.5 (3.42) & 53.8 (3.37) \\
64K & 4 & 59.3 & 38.6 & 153.5 (2.59) & 191.5 (3.23) \\
64K & 16 & 203.8 & 55.5 & 294.4 (1.44) & 452.1 (2.22) \\
\bottomrule
\end{tabular}
\end{table}

In this engine the indexer is 1.9 times faster than the stock engine for one stream at 32K and 3.4 times at 64K, and with 16 streams and three-bit keys 1.5 and 2.2 times.
These ratios mainly reflect the stock decoding kernel of this version, which is slow for one long sequence: PyTorch dense, which uses no indexer, is already 2.5 times faster than it at 64K with one stream, and against the faster dense configuration of each cell the indexer is 0.96 to 1.34 times as fast with one stream at 32K and 64K.
At 8K with 16 streams the indexer reaches only 0.86 (16-bit) and 0.94 (three-bit) of the stock engine.

\paragraph{The unfused path.}
A first version of the plugin issued about fifty small kernels per layer around the same scan and select.
Outside the engine its indexer work for 27 layers took 31\,ms per step, of which 6\,ms were kernel time, so the step was bound by issuing kernels; the fused path takes 4.5 to 6.8\,ms for one stream.
Inside vLLM~0.8.5 the unfused path decoded one stream at about 26 tokens per second at every length, half the rate of the fused one.
The speed of sparse decoding in a serving engine therefore depends on the overhead around the kernels at least as much as on the kernels.

\paragraph{Scope.}
The measurements use three models that were trained with dense attention, a distilled indexer and greedy decoding of 256 tokens; FlashMLA is the baseline only on the H200, because it does not run on an A100, and Qwen3-30B-A3B and Llama-3.1-8B were measured only on the H200.

\section{Negative results on sparse-code indexers}
\label{app:negative}

We also tested an indexer whose keys are sparse codes by construction and are distilled end to end, as an alternative to the post-hoc codec of Appendix~\ref{app:codec}.
We report it because its failure explains why the codec is applied after distillation and why the rotation is kept.

\paragraph{Design.}
Each key is the $n$ largest entries of a 4,096-dimensional activation that is computed from the layer input through a projection of rank 256, and it is stored as $n$ pairs of atom id and 4-bit weight.
The query assigns one signed weight to each atom.
The score is the sum of the query weights over the atoms of the key, plus the learned distance term of \DistBias{}; there is no rotation, so that a score can be computed by table lookup.
The indexer is trained with Eq.~\ref{eq:kl} on 8K sequences with ordinary positions, with $n$ cycled over $\{8, 16, 32\}$ so that one model can be evaluated at each size.

\paragraph{Result.}
Table~\ref{tab:sparse} compares this indexer with dense-key indexers at 100M tokens.
The sparse-code indexer with $n=8$ is within 0.6 points of \DistBias{} for $k=2048$ at every length, so the sparsity itself costs little.
It is behind \Skip{64K} by a factor of about two at 32K and 64K, as \DistBias{} is.
The post-hoc \Codec{8} at the same 17 bytes is better at every length, for example 1.66\% against 2.64\% at 64K for $k=2048$.
The limit of the distilled sparse code is therefore its position-free score and not its sparsity.

\begin{table}[h]
\caption{Sparse-code indexers distilled end to end, 100M tokens: \dppl{} (\%) for $k=512$ / $k=2048$. Four documents at 8K to 32K and two at 64K; the \Codec{8} row at 32K uses two documents and is not comparable with the other rows in that column.}
\label{tab:sparse}
\centering
\small
\setlength{\tabcolsep}{3.6pt}
\begin{tabular}{lrcccc}
\toprule
Indexer & Bytes & 8K & 16K & 32K & 64K \\
\midrule
\Skip{64K} & 128 & $+1.66$ / $+0.29$ & $+2.47$ / $+0.45$ & $+5.42$ / $+1.53$ & $+6.17$ / $+1.27$ \\
\DistBias{} & 128 & $+2.26$ / $+0.35$ & $+2.56$ / $+0.81$ & $+6.53$ / $+2.27$ & $+7.99$ / $+2.19$ \\
Distilled sparse code, $n=32$ & 65 & $+2.32$ / $+0.43$ & $+3.05$ / $+0.97$ & $+7.79$ / $+2.64$ & $+9.90$ / $+2.57$ \\
Distilled sparse code, $n=8$ & 17 & $+2.22$ / $+0.65$ & $+2.94$ / $+0.91$ & $+7.84$ / $+2.79$ & $+9.45$ / $+2.64$ \\
\Codec{8} on \Skip{64K} & 17 & $+1.62$ / $+0.31$ & --- & ($+3.01$ / $+0.61$) & $+6.25$ / $+1.66$ \\
\bottomrule
\end{tabular}
\end{table}

\paragraph{Two attempts to repair it.}
The first attempt added position skipping to the distance term, so that distances up to 64K occur in training.
In a 25M-token pilot this reduced the \dppl{} at 32K for $k=2048$ from 1.82\% to 1.41\%, against 0.75\% for \Skip{64K} at the same number of tokens, and at 64K from 3.37\% to 3.18\%, against 1.84\%.
A learned function of the distance is thus a weaker position mechanism than a rotation inside the dot product, with or without skipping.
The second attempt trained the code as an inverted index, with queries restricted to their 64 largest positive atoms and a regulariser on the expected number of index entries read per query.
The regulariser drove the codes to a state in which a query and a key almost never share an atom.
The training loss rose from 1.05 to 1.91 during the run, and the \dppl{} at 8K for $k=2048$ was 9.3\%.
We did not retune this run.
Both pilots use one seed, and the conclusion is limited to the tested designs.

\section{Sharing keys across layers}
\label{app:shared}

The index holds one key per token in each of the 27 layers.
We tested whether layers can share keys.
In the shared-key variant eight layers own keys, namely layers 0, 3, 4, 5, 6, 10, 19 and 26, and each of the other 19 layers keeps only its query projections and scores the keys of the nearest earlier layer that owns keys.
The variant is initialised from \Skip{64K} and distilled for 50M tokens.
Table~\ref{tab:shared} reports it at 32K.
With a full scan in every layer, sharing costs 0.27 points of \dppl{} at $k=2048$, from 0.36\% to 0.63\%, and reduces the key memory by a factor of 3.4.
Reusing the selected set of a key-owning layer without any scoring is much worse at 3.16\%.
Rescoring a candidate set of the key-owning layer lies between the two.
One seed and two documents support these values, so we report the variant as a promising extension and not as a result of the paper.

\begin{table}[h]
\caption{Shared keys at 32K with eight key-owning layers: \dppl{} (\%) on two PG-19 documents. ``Reuse'' copies the selected set of the nearest earlier key-owning layer. ``Rescore $k'$'' selects the top $k$ among the $k'$ best candidates of that layer.}
\label{tab:shared}
\centering
\small
\begin{tabular}{lcc}
\toprule
Configuration & $k=512$ & $k=2048$ \\
\midrule
Scan in every layer, own keys (27 key sets) & $+2.48$ & $+0.36$ \\
Scan in every layer, shared keys (8 key sets) & $+3.29$ & $+0.63$ \\
Reuse, no scoring & $+24.9$ & $+3.16$ \\
Rescore $k'=2k$, own keys & $+13.9$ & $+1.59$ \\
Rescore $k'=4k$, own keys & $+6.28$ & $+0.54$ \\
Rescore $k'=2k$, shared keys & $+12.4$ & $+1.42$ \\
Rescore $k'=4k$, shared keys & $+6.13$ & $+0.80$ \\
\bottomrule
\end{tabular}
\end{table}

\section{Forced local window}
\label{app:window}

Table~\ref{tab:window} gives the forced-window results at 32K that Section~\ref{sec:q4} summarises.
With a forced window of $k_w$ tokens, the $k_w$ most recent positions are always selected and the indexer chooses the remaining $k-k_w$.
A window of 1,024 tokens brings \NoPE{} to 0.82\% for $k=2048$, so the position-free indexer matches content correctly at four times its training length.
The window consumes half of the budget at $k=2048$, and at 64K \NoPE{} with $k_w=1024$ has a \dppl{} of 3.91\% against 1.27\% for \Skip{64K} on the same documents.
The window has a second limit at small budgets.
At $k=512$ it leaves little room for the indexer, and \NoPE{} with $k_w=256$ has 7.16\% against 2.84\% for \Skip{64K}.
\Skip{64K} does not need the window, since its values change by at most 0.23 points for $k_w \le 256$.

\begin{table}[h]
\caption{Forced local window at 32K: \dppl{} (\%) for $k=512$ / $k=2048$ when the $k_w$ most recent positions are always selected. All indexers use 100M tokens. $k_w=1024$ is evaluated for $k=2048$ only. This table uses two documents, so its first column differs from the four-document values of Table~\ref{tab:pos}.}
\label{tab:window}
\centering
\small
\begin{tabular}{lcccc}
\toprule
Indexer & No window & $k_w=64$ & $k_w=256$ & $k_w=1024$ \\
\midrule
\NoPE{} & $+14902$ / $+3483$ & $+8.66$ / $+5.06$ & $+7.16$ / $+1.98$ & $+0.82$ \\
\RopeEight{} & $+17.2$ / $+4.19$ & $+11.2$ / $+2.62$ & $+6.89$ / $+1.71$ & $+0.93$ \\
\Skip{64K} & $+2.84$ / $+0.55$ & $+2.61$ / $+0.60$ & $+2.83$ / $+0.65$ & $+0.44$ \\
\bottomrule
\end{tabular}
\end{table}

Table~\ref{tab:window8k} gives the same ablation at 8K, the training length.
A window of 256 tokens brings \NoPE{} to 0.40\% for $k=2048$, close to \Skip{64K} at 0.29\%.
At the training length the position-free indexer therefore lacks only local context, and a narrower window suffices than at 32K.

\begin{table}[h]
\caption{Forced local window at 8K: \dppl{} (\%) for $k=512$ / $k=2048$ on four PG-19 documents. All indexers use 100M tokens. $k_w=1024$ is evaluated for $k=2048$ only.}
\label{tab:window8k}
\centering
\small
\begin{tabular}{lcccc}
\toprule
Indexer & No window & $k_w=64$ & $k_w=256$ & $k_w=1024$ \\
\midrule
\NoPE{} & $+3794$ / $+18.5$ & $+4.17$ / $+0.67$ & $+2.90$ / $+0.40$ & $+0.57$ \\
\RopeEight{} & $+4.98$ / $+0.57$ & $+3.51$ / $+0.40$ & $+2.74$ / $+0.53$ & $+0.77$ \\
\Skip{64K} & $+1.66$ / $+0.29$ & $+1.67$ / $+0.28$ & $+1.99$ / $+0.29$ & $+0.39$ \\
\bottomrule
\end{tabular}
\end{table}

\section{Extended discussion of related work}
\label{app:related}
This appendix provides the full version of Section~\ref{sec:related}.

\paragraph{Learned top-$k$ sparse attention and training the indexer.}
\citet{deepseekv32} introduced a sparse-attention mechanism in which a learned indexer assigns a relevance score to each cached position, and attention is computed only over the top-$k$ candidates.
Their training procedure first optimises the indexer alone---keeping the language model frozen and using dense attention as the distillation target---for 1{,}000 steps with batches of 16 sequences of length 128K, and then trains the full model using sparse attention.
Subsequent work adopted learned selection at either token or block granularity~\citep{minimax,deepseekv4,longcat,qwen38next}.

Two additional lines of work study how to add an indexer to an already-trained model.
\citet{rtpurbo} convert a model trained with full attention into a sparse variant by introducing a 16-dimensional indexer and running a few hundred training steps on sequences of 32K--80K tokens.
\citet{flashmemory} train a block-level indexer for DeepSeek-V4 as a dual encoder (without loading the full model), using contexts of up to 512K tokens, and report generalisation to twice the training length.
\citet{adindex} distill a group-level indexer whose scan is shared across neighbouring queries.

Beyond the remark of \citet{flashmemory}, we did not find quantitative evaluations of how far token-level indexers remain effective when deployed at lengths substantially beyond those seen during training.
In our setting, we retain the indexer architecture and objective of \citet{deepseekv32} and vary only the sequences used for distillation, which are substantially shorter than the serving context.

\paragraph{Position encodings and length extrapolation.}
The failure mode we analyse is a specific instance of a broader limitation of positional encodings.
\citet{rope} introduced the rotary position embedding used by the indexer.
\citet{nope} compared position-encoding schemes for length generalisation and found that commonly used choices---including rotary embeddings---extrapolate poorly; \citet{randpe} further linked this degradation to positions that are never observed during training.

For pretrained language models, \citet{pi} proposed scaling position indices into the training range followed by brief fine-tuning, and \citet{yarn} refined this idea via frequency-dependent scaling.
The method most closely related to ours is positional skip-wise training (PoSE)~\citep{pose}.
PoSE fine-tunes a language model on a short window partitioned into chunks, applies a random skipping bias to the position ids of each chunk, and thereby extends a model trained with a 2K window to 128K.

A separate set of approaches obtains extrapolation through architectural changes, for example by mixing attention with and without rotary embeddings~\citep{ropetonope,swangpt} or by using hierarchical sparse attention whose selection extrapolates beyond the training length~\citep{leng,hils,recsa}.
These methods train or fine-tune the language model.
By contrast, we apply the PoSE-style manipulation to a different component: the language model remains frozen and already supports long contexts, while only the indexer is trained.
Crucially, the frozen model is provided with the same skipped position ids so that its dense attention defines a target at long-range distances that are absent in the unskipped short sequences.
This is necessary because, unlike a language model fine-tuned with PoSE (which receives a next-token loss at every distance), the indexer has no intrinsic training loss and can only learn long-range selection when the teacher signal is present.

Finally, position-id compression for the indexer beyond the trained range (Eq.~\ref{eq:compress}) completes the method; no additional fine-tuning is required because the indexer only ranks keys.
The scaling idea of \citet{pi} appears in our work as position compression applied to the indexer alone.
In a targeted (though not exhaustive) search, we did not find prior work that trains sparse-attention indexers using skipped position ids.

\paragraph{Inference-time cost of the index.}
The index consists of one key per token per layer and must be scanned at each decoding step.
Several works reduce the number of scans.
\citet{indexcache} allow most layers to reuse selections from a nearby layer, \citet{hisa} filter blocks prior to token scoring, and \citet{pivot,adindex} share a single scan across neighbouring queries.
Other work reduces the cost of the scan itself.
\citet{deepseekv32} execute the indexer in 8-bit floating point, and \citet{deepseekv4} train the query--key pathway with 4-bit quantisation.
\citet{streamindex,litetopk} avoid materialising the full score matrix prior to the top-$k$ operation, and \citet{gvr} accelerate the top-$k$ step.

A further line of research replaces dot-product similarity with binary codes in dense-attention models: \citet{hashattention,hata,spotlight} learn hash functions for queries and keys and select candidates by Hamming distance.
Our approach differs from the low-bit indexers of \citet{deepseekv32,deepseekv4} in that we quantise the indexer keys after distillation (without retraining) and compute scores directly on packed integers in a single kernel.
It also differs from hashing-based methods in that we retain the indexer and its scoring function, rather than training a separate retrieval mechanism.
In our experiments, one-bit codes for indexer keys discard too much attention mass.
Because we do not train learned hash functions on our indexer, this observation pertains to key quantisation in our setting and should not be interpreted as a general statement about learned hashing.

Finally, scan-reduction methods are complementary to our compact index and could likely be combined; we leave this evaluation to future work.

\section{Indexers in fewer layers}
\label{app:fewer}

This appendix gives the data of Section~\ref{sec:q6}.
\paragraph{Overlap between layers and placement.}
Neighbouring layers overlap only partly (Figure~\ref{fig:reuse}, left): at 32K the Jaccard overlap with the previous layer is 0.22 to 0.46, lowest in layers 21 to 26, and the indexer follows \Oracle{}, so this is a property of the model's attention.
With \Skip{64K} in 8 of 27 layers (Figure~\ref{fig:reuse}, right), \Gapfill{8} beats \Uniform{8} on perplexity (2.86\% against 6.46\% for $k=2048$) and on the passcode probe on which its layers were chosen, but not on the held-out RULER tasks, where it is worse on five of seven and has the lower mean, 0.451 against 0.461.
A non-uniform placement thus does not beat an even spread on unseen tasks, and on multi-key-1, multi-value and single-1 either placement loses 7 to 29 points against all layers.

\begin{figure}[h]
\centering
\begin{minipage}[t]{0.46\linewidth}
\vspace{0pt}
\centering
\begin{tikzpicture}
\begin{axis}[
  width=\linewidth, height=3.9cm,
  xmin=0.5, xmax=26.5, ymin=0.15, ymax=0.82,
  xlabel={Layer $\ell$}, ylabel={Overlap $J_{\ell,\ell-1}$},
  xtick={1,5,10,15,20,26}, ytick={0.2,0.3,0.4,0.5,0.6},
  ymajorgrids, grid style={inkmuted!25, line width=0.3pt},
  axis line style={inkmuted}, tick style={inkmuted},
  tick label style={font=\scriptsize}, label style={font=\scriptsize},
  legend style={font=\scriptsize, draw=none, fill=white, fill opacity=0.8, text opacity=1, at={(0.99,0.99)}, anchor=north east, cells={anchor=west}, row sep=-2pt},
  clip=false,
]
\addplot[cB, line width=0.9pt, mark=square*, mark size=1.1pt] coordinates {(1,0.395) (2,0.454) (3,0.367) (4,0.564) (5,0.546) (6,0.554) (7,0.526) (8,0.477) (9,0.581) (10,0.535) (11,0.603) (12,0.502) (13,0.518) (14,0.540) (15,0.413) (16,0.484) (17,0.531) (18,0.530) (19,0.461) (20,0.381) (21,0.388) (22,0.402) (23,0.513) (24,0.462) (25,0.396) (26,0.430)};
\addlegendentry{indexer, 8K context}
\addplot[cA, line width=0.9pt, mark=*, mark size=1.1pt] coordinates {(1,0.256) (2,0.399) (3,0.286) (4,0.429) (5,0.417) (6,0.418) (7,0.418) (8,0.318) (9,0.459) (10,0.401) (11,0.459) (12,0.353) (13,0.399) (14,0.405) (15,0.277) (16,0.356) (17,0.426) (18,0.408) (19,0.336) (20,0.284) (21,0.247) (22,0.254) (23,0.369) (24,0.316) (25,0.223) (26,0.248)};
\addlegendentry{indexer, 32K context}
\addplot[inkmuted, line width=0.9pt, densely dashed, mark=none] coordinates {(1,0.244) (2,0.364) (3,0.313) (4,0.448) (5,0.396) (6,0.409) (7,0.397) (8,0.318) (9,0.423) (10,0.377) (11,0.439) (12,0.346) (13,0.378) (14,0.387) (15,0.259) (16,0.331) (17,0.398) (18,0.379) (19,0.310) (20,0.267) (21,0.230) (22,0.243) (23,0.341) (24,0.291) (25,0.220) (26,0.233)};
\addlegendentry{\Oracle{}, 32K context}
\pgfplotsinvokeforeach{7,10,12,19,21}{\draw[cD, line width=1.2pt] (axis cs:#1,0.15) -- (axis cs:#1,0.18);}
\end{axis}
\end{tikzpicture}
\end{minipage}\hfill
\begin{minipage}[t]{0.52\linewidth}
\vspace{2pt}
\centering
\footnotesize
\setlength{\tabcolsep}{2pt}
\begin{tabular}{lcccc}
\toprule
 & \Dense{} & All & \Uniform{8} & \Gapfill{8} \\
\midrule
\multicolumn{5}{@{}l}{\emph{PG-19 \dppl{} (\%), 32K}} \\
$k=512$ & & $+5.12$ & $+26.5$ & $+9.07$ \\
$k=2048$ & & $+1.10$ & $+6.46$ & $+2.86$ \\
\multicolumn{5}{@{}l}{\emph{Passcode, retrieved of 8 (in-sample)}} \\
$k=512$ & 8 & 8 & 4 & 8 \\
$k=2048$ & 8 & 7 & 3 & 7 \\
\multicolumn{5}{@{}l}{\emph{RULER exact match, 32K, $k=2048$}} \\
multi-key-1 & 0.680 & 0.585 & 0.340 & 0.300 \\
multi-value & 0.545 & 0.230 & 0.085 & 0.160 \\
mean of 7 & 0.599 & 0.516 & 0.461 & 0.451 \\
\bottomrule
\end{tabular}
\end{minipage}
\caption{Left: Jaccard overlap of the top-2048 sets of each layer and the previous one on DeepSeek-V2-Lite (four PG-19 documents, last 1024 decoding steps, indexer \Skip{64K}). Orange ticks: the five retrieval layers. Right: \Skip{64K} in all 27 layers (All) or in 8, the others reusing the nearest earlier selection; the passcode probe is in-sample for \Gapfill{8}, the RULER tasks are held out (all seven in Table~\ref{tab:reuseruler}).}
\label{fig:reuse}
\end{figure}

\section{Limitations in full}
\label{app:limits}

This appendix provides additional discussion for Section~\ref{sec:limits}.

\paragraph{Models.}
Our primary results use DeepSeek-V2-Lite, with smaller-scale replications on Qwen3-30B-A3B and Llama-3.1-8B.
All three models are trained with dense attention. We do not fine-tune them for sparse attention, and we do not evaluate models trained natively with sparse attention.
For Llama-3.1, each indexer is trained once on 100M tokens. At 64K, \LongI\compress{} improves perplexity by 0.47 points relative to \Skip{64K}; at 32K, \LongI{} improves by 0.53 points (measured on four blocks). At 128K, all indexers (and \Oracle{}) fail unless the first two layers retain dense attention; this diagnosis is based on two blocks. \Skip{128K}, trained with a 128K range, also requires dense attention in the first two layers.
For Qwen3, each indexer is trained once on 100M tokens. At 64K (eight blocks), \LongI\compress{} improves perplexity by 0.94 points over \Skip{64K}, reversing the advantage of \Skip{64K} over compression observed on DeepSeek-V2-Lite. The needle tasks on Qwen3 are saturated (even with 200 samples); for the three unsaturated tasks, we therefore compare indexers only via token accuracy and answer loss.

\paragraph{Statistics.}
Each configuration is trained once, except \Skip{64K} and \LongI{} on Llama-3.1-8B, which are trained twice (Appendix~\ref{app:seeds}). The second \LongI{} run also changes the batch size, so it does not measure seed-to-seed variability for a fixed recipe.
Perplexity values in the main tables are computed on four PG-19 blocks up to 32K and on two blocks at 64K and 128K. The exception is the 100M-token indexers at 64K (Tables~\ref{tab:pair-main}, \ref{tab:pair} and~\ref{tab:second}), which use eight blocks. Confidence intervals are reported for oracle selection: 16 blocks from 8K to 64K for the three 400M-token indexers (Table~\ref{tab:pg19ci}); 16 blocks at 32K and eight at 64K for the 100M-token DeepSeek-V2-Lite indexers, and eight blocks at 64K for the other two models (Table~\ref{tab:pairci}); eight blocks at 128K for \Skip{64K}, \LongI{}, and \Skip{128K} (Table~\ref{tab:pg19ci128}); and eight blocks at 160K (Table~\ref{tab:pg19ci160}).
All intervals are computed over blocks, which come from a small number of books and may cross book boundaries (Appendix~\ref{app:details}). They therefore exclude between-book variation and between-run variation.
The matched 64K baseline, \LongLL{}, is trained once on 100M tokens; its training time per token is measured separately on an idle GPU (Table~\ref{tab:cost}).
We do not distill any indexer with ordinary positions at 128K; the cost for that recipe is estimated from a 20-step timing run on a shared node.
At 128K and 160K, the sparse model can achieve lower perplexity than \Dense{} (Tables~\ref{tab:pg19ci128} and~\ref{tab:pg19ci160}). As a result, \dppl{} at these lengths is a less direct measure of selection quality, and we treat the corresponding values as indicative.
The comparison between \Skip{64K} and \LongI{} is matched by training tokens, not by the number of optimizer steps.

\paragraph{Retrieval.}
We run RULER with 200 samples at 32K, and with 50 samples for four tasks at 64K. We use teacher-forced scoring rather than generation. For DeepSeek-V2-Lite at 32K with 400M-token indexers, only single-1 and multi-value have per-sample scores; these come from a rerun on a different GPU, with rates that differ by up to 2.5 points.
In a paired test at 32K, the 400M-token \Skip{64K} underperforms \LongI{} on multi-value. At 64K, every indexer degrades on this task, while at 100M tokens \Skip{64K} and \LongI{} are tied. Increasing the budget to 4{,}096 matches \Oracle{} with $k=2048$ (Table~\ref{tab:rulerctl}); we have not identified why selection at $k=2048$ is worse on this task.
On Llama-3.1, single-2 is affected by our evaluation protocol, which scores only the first value (Appendix~\ref{app:third}).
The passcode probe uses eight samples and only distinguishes failures from non-failures.
Our end-to-end evaluation of compact keys includes perplexity up to 64K and the passcode probe, but not the RULER tasks.
The only downstream task with natural questions is LongBench~v2 (Appendix~\ref{app:longbench}). Contexts are truncated to 32K, and we score answers by the most likely option letter rather than by generation. DeepSeek-V2-Lite performs at chance, so on that model LongBench~v2 does not meaningfully test selection. With 503 questions, it also cannot reliably resolve differences of 1--2 points. We do not evaluate generative downstream tasks or downstream tasks beyond 32K.

\paragraph{Length.}
The maximum evaluated length is 128K on Qwen3 and Llama-3.1, and 160K on DeepSeek-V2-Lite (the limit of its position IDs). At 160K, \Dense{} is weak and we report only paired differences (Table~\ref{tab:pg19ci160}).
To measure out-of-range degradation, we evaluate \Skip{32K} at 64K and 128K, \Skip{64K} at 128K, and \Skip{128K} at 160K on DeepSeek-V2-Lite; we also evaluate \Skip{64K} at 128K and \Skip{32K} at 64K and 128K on Qwen3 and Llama-3.1. Results at 128K are computed on four blocks; for Qwen3 and Llama-3.1, the range ablation covers only \Skip{32K} and \Skip{64K}.

\paragraph{Speed.}
We time the scan kernel both in isolation (synthetic keys) and within a PyTorch reference implementation of a single decoding step on one document (Appendix~\ref{app:decode}).
We measure decoding throughput inside a serving engine on DeepSeek-V2-Lite (whose multi-head latent attention makes dense decoding comparatively cheap), and on H200 also on Qwen3-30B-A3B and Llama-3.1-8B.
On DeepSeek-V2-Lite with vLLM~0.32.0 on H200, the indexer with three-bit keys outperforms the Triton dense kernel with four or more streams. However, relative to FlashMLA with an 8-bit cache (our strongest dense baseline), it wins only at 64K with 16 streams and by about 5\% (Table~\ref{tab:servingfp8}), which is comparable to between-session measurement variation; it is slower than FlashMLA for a single stream at every length. On A100 (where FlashMLA does not run), it outperforms the stock engine only at 32K and 64K with several streams; the larger gains in vLLM~0.8.5 reflect the slower dense kernel in that version.
On Qwen3-30B-A3B, whose grouped-query cache is larger per token, three-bit keys beat dense decoding at 32K with 16 streams and at 64K with four streams; the indexer is slower for a single stream at every length (Table~\ref{tab:servingq3}).
On Llama-3.1-8B (H200; up to 128K; at most four streams beyond 64K), the indexer is slower than dense decoding for a single stream at every length, and in every setting up to 64K. It is faster only with four streams at 96K and 128K, including in the configuration whose perplexity remains stable at 128K (Tables~\ref{tab:servingl8} and~\ref{tab:servingl8dep}). We verified generated tokens against the reference implementation up to 64K and, for that configuration, at 128K. The indexer's key stores lie outside the memory fraction allocated to the engine.
The fastest DeepSeek-V2-Lite configuration keeps an FP8 copy of the cache alongside the engine's 16-bit cache, because FlashMLA's sparse kernel reads only FP8 rows. This increases cache memory per token by about 1.6$\times$ relative to 16-bit dense decoding; a backend that stores the cache once in FP8 would avoid this overhead and increase the effective context capacity.
We evaluate the effect of FP8 caches on quality only via token agreement on a single stream of 256 tokens per length, not via perplexity.

\paragraph{Index reuse.}
We evaluate running the indexer in 8 of 27 layers on seven RULER tasks, perplexity at 32K, and the passcode probe. The \Gapfill{8} layers are chosen using the passcode probe, making that probe in-sample; neither the model nor the indexer is adapted for reuse.

\paragraph{Baselines.}
Our block-level baseline is a re-implementation of \citet{flashmemory} with twelve details inferred from the paper. When its selection is shared across 64 queries, it fails on RULER. Refreshing selection at every query recovers single-1 and partially recovers single-2, but we did not run it on multi-key-1 or multi-value (Appendix~\ref{app:flashmemory}).

\paragraph{Assumption.}
Our method assumes the frozen model behaves sensibly for position IDs up to the chosen range (and only within the model's supported context length). The training target also ignores the text omitted by skipped positions.

\end{document}